\documentclass[lettersize,journal]{IEEEtran}
\usepackage{amsmath,amsfonts}
\usepackage[lined,boxed,commentsnumbered]{algorithm2e}
\usepackage{array}
\usepackage[caption=false,font=normalsize,labelfont=sf,textfont=sf]{subfig}
\usepackage{textcomp}
\usepackage{stfloats}
\usepackage{url}
\usepackage[numbers,sort&compress]{natbib}
\usepackage{verbatim}
\usepackage{tikz,xcolor}
\usepackage{color, xcolor}
\usepackage{soul}

\soulregister\cite7
\soulregister\citep7
\soulregister\citet7
\soulregister\ref7
\soulregister\pageref7

\definecolor{lime}{HTML}{A6CE39}
\DeclareRobustCommand{\orcidicon}{
\begin{tikzpicture}
\draw[lime, fill=lime] (0,0)
circle[radius=0.16]
node[white]{{\fontfamily{qag}\selectfont \tiny \.{I}D}};
\end{tikzpicture}
\hspace{-2mm}
}
\foreach \x in {A, ..., Z}{%
\expandafter\xdef\csname orcid\x\endcsname{\noexpand\href{https://orcid.org/\csname orcidauthor\x\endcsname}{\noexpand\orcidicon}}
}

\usepackage{graphicx}
\usepackage{multirow}
\usepackage{amsmath}

\newtheorem{definition}{Definition}
\usepackage{hyperref}
\begin{document}

\title{Dynamic Causal Explanation Based Diffusion-Variational Graph Neural Network for Spatio-temporal Forecasting}

\author{Guojun Liang\hspace{-1.5mm}\orcidL{}, Prayag Tiwari\hspace{-1.5mm}\orcidP{}, S\l{}awomir Nowaczyk\hspace{-1.5mm}\orcidS{}, Stefan Byttner, Fernando Alonso-Fernandez\hspace{-1.5mm}\orcidF{}

\thanks{This work was partially carried out with support from Vinnova (Sweden’s innovation agency) through Vehicle Strategic Research and Innovation programme, FFI; F. A.-F. is partially funded by the Swedish Research Council (VR) and by Vinnova. (\textit{Corresponding authors: Prayag Tiwari, S\l{}awomir Nowaczyk})}
\thanks{G. Liang, P. Tiwari, S. Nowaczyk, S. Byttner, and F.A. Fernandez are with CAISR, School of Information Technology, Halmstad University, Sweden (Email: guojun.liang@hh.se, prayag.tiwari@hh.se, slawomir.nowaczyk@hh.se, stefan.byttner@hh.se, feralo@hh.se)}
}




\maketitle

\begin{abstract}
Graph neural networks (GNNs), especially dynamic GNNs, have become a research hotspot in spatio-temporal forecasting problems. While many dynamic graph construction methods have been developed, relatively few of them explore the causal relationship between neighbour nodes. Thus, the resulting models lack strong explainability for the causal relationship between the neighbour nodes of the dynamically generated graphs, which can easily lead to a risk in subsequent decisions. Moreover, few of them consider the uncertainty and noise of dynamic graphs based on the time series datasets, which are ubiquitous in real-world graph structure networks. In this paper, we propose a novel Dynamic Diffusion-Variational Graph Neural Network (DVGNN) for spatio-temporal forecasting. For dynamic graph construction, an unsupervised generative model is devised. Two layers of graph convolutional network (GCN) are applied to calculate the posterior distribution of the latent node embeddings in the encoder stage. Then, a diffusion model is used to infer the dynamic link probability and reconstruct causal graphs in the decoder stage adaptively. The new loss function is derived theoretically, and  the reparameterization trick is adopted in estimating the probability distribution of the dynamic graphs by Evidence Lower Bound (ELBO) during the backpropagation period. After obtaining the generated graphs, dynamic GCN and temporal attention are applied to predict future states. Experiments are conducted on four real-world datasets of different graph structures in different domains. The results demonstrate that the proposed DVGNN model outperforms state-of-the-art approaches and achieves outstanding Root Mean Squared Error (RMSE) result while exhibiting higher robustness. Also, by F1-score and probability distribution analysis, we demonstrate that DVGNN better reflects the causal relationship and uncertainty of dynamic graphs. The website of the code is \href{https://github.com/gorgen2020/DVGNN}{\textit{https://github.com/gorgen2020/DVGNN}}.
\end{abstract}


\section{Introduction}
\IEEEPARstart{S}{patio-temporal} forecasting is a core issue in many applications, such as neuroscience, transportation, healthcare, etc. With the rapid growth of time series datasets generated by different domains, spatio-temporal forecasting is also becoming central to many fields of scientific research. However, owing to the fast-changing and complex correlation of spatio-temporal data, it is often challenging to discover the causal relationship and achieve high-accuracy prediction. Moreover, many collected data have complex graph structures and are noisy, making the problem more difficult. Traditional time series forecasting models, such as  Auto-Regressive Integrated Moving Average (ARIMA) \cite{zhang2022dynamic} and Vector AutoRegression (VAR) \cite{chandra2009predictions}, Hidden Markov Model (HMM) \cite{yu2003short} and Support Vector Regression (SVR) \cite{Smola2004} are of great interpretability and low computational complexity. However, they ignore the spatial features and require manual feature engineering, which is not suitable for big data application scenarios.

 \begin{figure}[]   
 \centering   
 \subfloat[]{   
 \includegraphics[width=\columnwidth]{diagrama.pdf}     \label{diagrama}    
 }        

 \subfloat[]{          \includegraphics[width=\columnwidth]{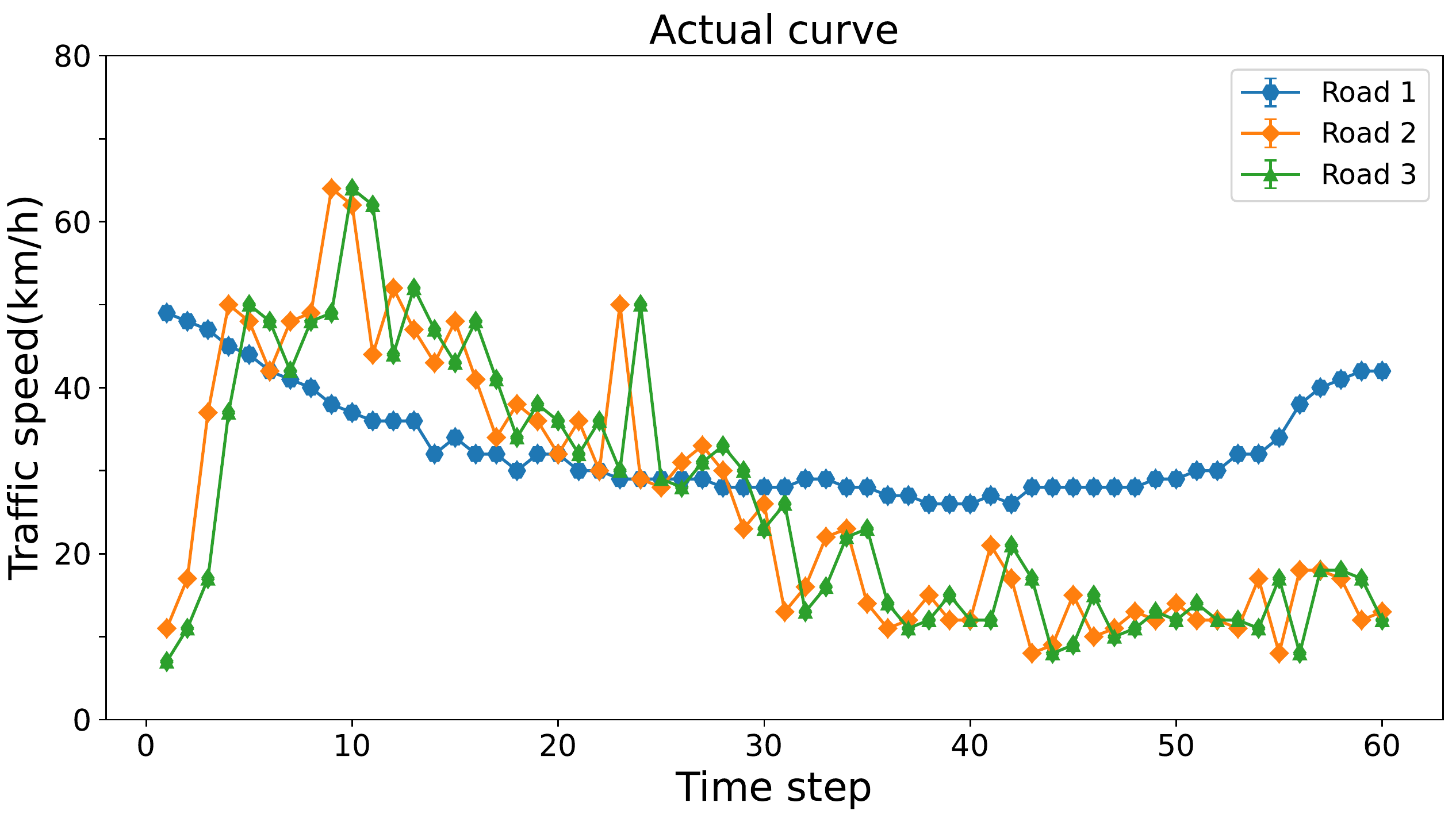}     \label{diagramb}  
 }       
 
 \subfloat[]{          \includegraphics[width=\columnwidth]{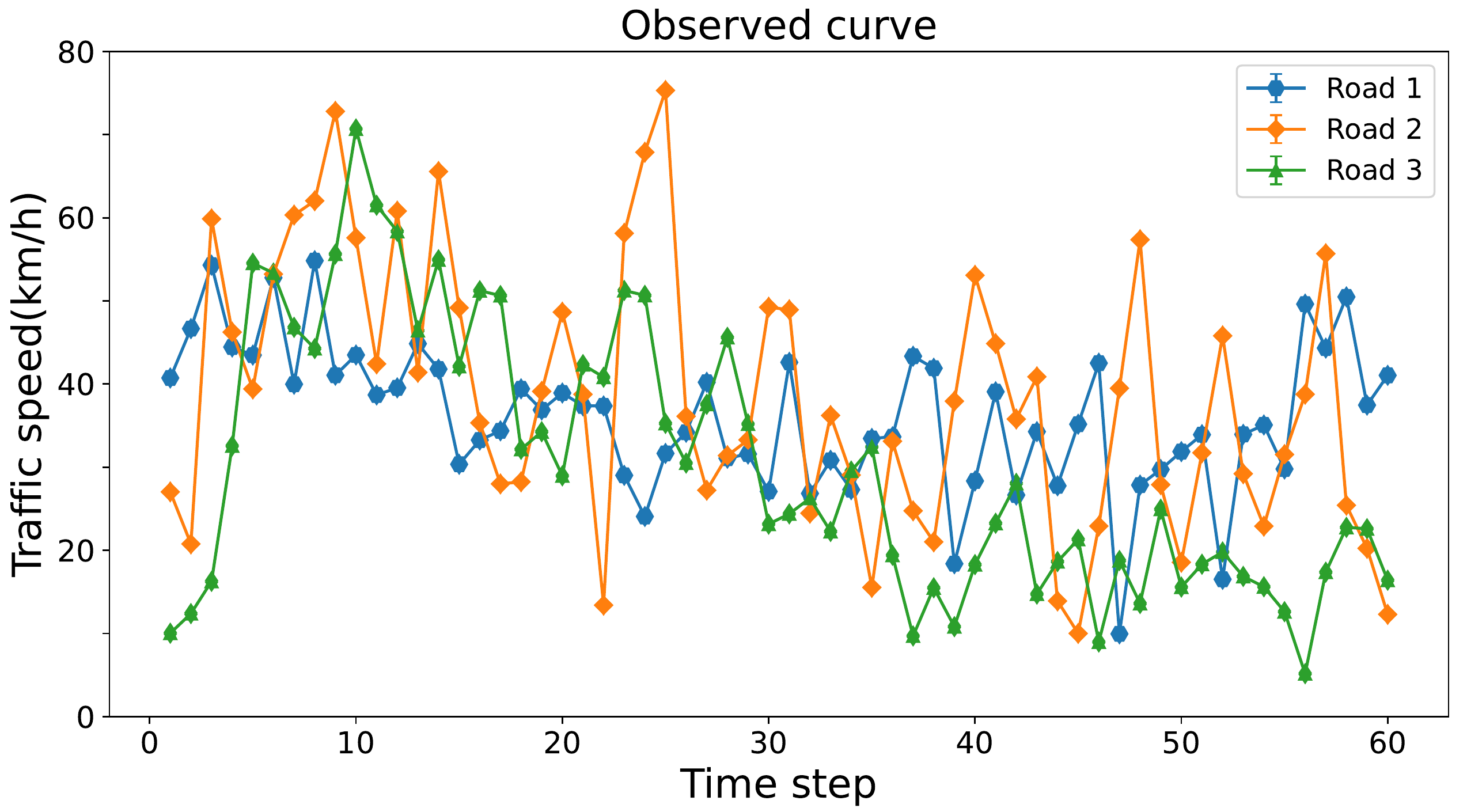}     \label{diagramc} 
 }       
 \caption{Dynamic correlation of neighbour nodes and noisy observation, the red arrow denotes the Origin and Destination (OD) transition.} 
 \label{concept} 
 \end{figure}

With the development of neural networks in the big data era, prediction accuracy has been improved greatly \cite{yang2016optimized}. Many scholars have applied deep learning methods to such time series forecasting and causal discovery problems. Temporal Causal Discovery Framework (TCDF) \cite{nauta2019causal} applied convolutional neural networks (CNN) and attention mechanism to discover the causal graph of time series. Neural Granger Causality (NGC) \cite{tank2021neural} modelled each time series by multilayer perceptron (cMLP) and long short-term memory (LSTM) for Granger causality and prediction. Causal Recurrent Variational Autoencoder (CR-VAE) \cite{li2023causal} integrated the concepts of Granger causality into a recurrent VAE framework for causal discovery and prediction. Nevertheless, these causal discovery models focus on capturing the temporal features, and often fail when dealing with non-Euclidean graph-structure data. Therefore, their prediction performance is low in spatio-temporal forecasting problems. To capture the non-Euclidean graph structure features of time series data, graph neural networks (GNNs) have been devised to apply neural networks for processing data with graph structure. The key design of GNNs is the use of pairwise message passing, which represent the node embedding by exchanging information with their neighbours \cite{wu2022graph}. Although GNN has achieved great success in extracting spatial features of graphs, traditional GNN methods often adopt static graphs, neglecting that the edges in reality are complex, constantly changing, and usually nonlinear. Fortunately, more and more scholars are aware of this problem and attempt to construct dynamic graphs by various methods to further represent spatial-temporal features.

However, most of the dynamic graph methods belong to the class of discriminative models and learn the deterministic node features, which ignores the uncertainty of graph-structure data and the underlying stochastic generative process of the time series. Also, various noises are ubiquitous in real time series data, which makes it difficult to obtain real causal relationship and high-accuracy prediction. For example, Fig. \ref{diagrama} depicts three connected roads, traffic flow shuttle between Road 2 and Road 3 while neither of them interacts with Road 1 as  Origin and Destination (OD) red arrows indicate. As time elapses, trip patterns between Road 2 and Road 3 will get a high correlation and cause changes between their respective traffic flows. In turn, that results in a strong edge connecting the two roads in a graph representation. On the other hand, Road 1 has little interaction with them, which results in weak edges, or complete disconnection to Roads 2 and 3. In addition, from the trajectories points of the vehicles shown as the red circles on the map, points of Road 2 are denser than those of Road 3 at time $t$ and mainly cause the ``positive'' traffic flow change of Road 3 at time $t+T$. The actual traffic speeds of the three roads are shown in Fig. \ref{diagramb}. Unfortunately, as shown in the observed traffic speeds curve of Fig. \ref{diagramc}, due to the reliability of sensors or other unstable factors of traffic condition, the node signals collected by the sensors show certain uncertainty and stochasticity in reality and are prone to fluctuation affected by the impact of noise, which make it hard to discover this causal relationship. Therefore, it is more reasonable to consider the transition of OD as a stochastic process with noise.

To address the questions mentioned above, inspired by stochastic processes in physics, we propose a novel generative and unsupervised learning dynamic graph framework, Dynamic Diffusion-Variational Graph Neural Network (DVGNN), for spatio-temporal forecasting. It discovers causal relationship by optimizing variational lower bound w.r.t. the variational parameters of the Stochastic Differential Equation (SDE) to obtain a more accurate forecasting and better explainability of the generated graphs. The main contributions of this paper are as follows: 
\begin{itemize}    
\item A novel DVGNN is proposed for spatio-temporal forecasting. DVGNN adopts two-layer GCNs in the encoder stage and discovers the causal relationship of the neigbour nodes by the diffusion process in the decoder stage.       
\item  Uncertainty and noise of spatio-temporal data are formulated by noise-driven SDE in the diffusion process. Dynamic causal relationship of neighbour nodes are deduced and approximated theoretically.     
\item We conduct experiments on four real-world datasets of different graph structures in different fields. Experimental results demonstrate the superiority of DVGNN in spatio-temporal forecasting compared with state-of-the-art models. Furthermore, we analyze the uncertainty of the predicted dynamic edges, the robustness of the model, and the interpretability of the generated graphs. 
\end{itemize}

\section{Related work}
\subsection{Spatio-temporal Forecasting}  
Accurate and timely spatio-temporal forecasting is essential in many fields. However, it has always been considered an “open” scientific issue due to its complex spatial-temporal correlation \cite{bui2022spatial}. Many scholars have invested a lot of time and energy in studying this field. With the development of deep learning, many forecasting models have been invented. They can be sorted as temporal-dependent models, spatial-dependent models, and spatio-temporal models, which are a hybrid of both formers. Temporal-dependent models focus on capturing temporal features of time series, RNNs, CNN, and attention mechanisms are often applied \cite{zhaowei2020short}. LSTM \cite{zhao2017lstm} and Gated Recurrent Unit (GRU) \cite{cho2014learning} are the typical RNN methods that are employed to exploit the temporal features. RNNs are naturally suited to modelling sequence data \cite{bandara2020lstm}. LSTM is designed to deal with the gradient vanishing problem, and GRU is for reducing the computational complexity of LSTM; 1D CNN and attention are used to exploit the temporal features of the time series. TCDF \cite{nauta2019causal} sampled the time series by N independent attention-based CNNs. Bi-STAT \cite{chen2022bidirectional} designed the cross-attention module, which bridged the future and past temporal states to the present temporal states. TraverseNet \cite{wu2022traversenet} built connections between each neighbor’s past states to the node’s present states by attention mechanism. Spatial-dependent models focus on CNNs in the early research stage in light of their success in image. Deep Spatio-Temporal Residual Networks (ST-ResNet)\cite{zhang2017deep} converted the city into grid-shaped data, and then used CNN to capture spatial dependencies. CNN is very suitable for Euclidean structured data. However, most graphs of spatio-temporal data naturally belong to non-Euclidean structured data. As a result, it is difficult to extract all the relevant spatial features of these graphs by CNN. Therefore, GNN is devised to deal with these problems, which will be discussed in the following section. Spatial-temporal models combine temporal and spatial models to exploit  spatio-temporal features. CNN-LSTM (CLM) \cite{cao2020cnn} extracted the spatio-temporal features together with the output of CNN and LSTM layers. Spatial-temporal feature selection algorithm (STFSA) integrated a CNN and a GRU based on spatial-temporal analysis for short-term prediction \cite{ma2022novel}.
\subsection{Graph Neural Networks} 
In order to deal with non-Euclidean structured data, GNN is devised. Bruna et al. \cite{bruna2013spectral} attempted to apply convolution operation to the spectrum of the graph. Defferrard et al. approximated it by  singular value decomposition (SVD) and the Chebyshev polynomial \cite{defferrard2016convolutional}. Kipf and Max Welling \cite{kipf2016semi} proposed GCN via a localized first-order approximation of spectral graph convolutions. Since GCN has natural advantages in spatial information mining of graphs, GCN has become a promising spatial-dependent method in spatio-temporal forecasting. Also, to better exploit the spatio-temporal features of the time series data with the graph structure, a hybrid spatio-temporal dependent methods such as GCN-LSTM \cite{ali2022exploiting,cao2020cnn}, or GCN-GRU \cite{lv2020temporal,wang2020traffic} are often applied in many real application scenarios. T-GCN model applied GCN and GRU methods to exploit spatio-temporal features \cite{Zhao2020}. Diffusion Convolutional Recurrent Neural Network (DCRNN) \cite{li2017diffusion} modelled the problem on a directed graph and adopted bidirectional GCN combined with GRU for spatio-temporal forecasting.

Most of the models apply a static graph to GCN, ignoring the dynamics of the edges. Thus, these methods restrict the capacity of GCN to exploit the dynamics of spatio-temporal data.

 \subsection{Dynamic Graph Neural Networks} 
Recently, the importance of dynamic graphs in spatio-temporal forecasting problems has been recognized \cite{zhang2022dynamic,wang2023tyre,li2022spatial,chen2022novel}. Many methods have been developed to construct dynamic graphs. Graph Attention Network (GAT) \cite{Veli2018Gan} applied attention mechanism to construct the dynamic edge with variable weight. Attention-based Spatio-temporal GCN (ASTGCN)  \cite{guo2019attention} applied spatial and temporal attention mechanism to capture the spatio-temporal features and construct the dynamic matrix for GCN. Dynamic Graph Convolutional Recurrent Network (DGCRM) filtered the node embeddings and then used node embeddings and pre-defined static graph to generate dynamic graphs \cite{li2021dynamic}. Attention-based spatiotemporal graph attention network (ASTGAT) \cite{wang2022attention} used GAT and temporal attention to deal with dynamic spatio-temporal information. Semantics-aware Dynamic GCN (SDGCN) \cite{liang2023semantics} noticed the semantics of the dynamic graphs and extracted semantic dynamic graphs through HMM. However, this method requested the corresponding vehicle trajectory dataset to construct the dynamic graphs. Zheng et al. proposed STGODE \cite{fang2021spatial} which calculated the dynamic semantic adjacency matrix by Dynamic Time Warping (DTW), then, incorporated the semantic adjacency matrix and spatial adjacency matrix into GCN through Ordinary Differential Equations (ODE) for prediction. Variational Graph Recurrent Attention Neural Networks (VGRAN) \cite{zhou2020variational} introduced a novel Bayesian framework to calculate the posterior distributions of latent variables by GRU for dynamic graph convolution operations. Li et al. developed DSTGN, which inferred dynamic relevance between variable node signals and static topology graph to construct the dynamic graphs for prediction \cite{li2023dynamic}. Dynamic Spatial-Temporal Aware Graph Neural Network (DSTAGNN) introduced the Spatial-Temporal Aware Distance (STAD) component to construct the dynamic graphs and combined attention mechanism to exploit the spatio-temporal features for higher prediction \cite{lan2022dstagnn}.

Although the dynamic graph methods have improved the performance of GCN, they rarely consider the causal relationship of the generated graphs. Moreover, most dynamic graph construction models adopt supervised discriminative models, which are inadequate for modelling uncertainty. In addition, high-accuracy real-time spatio-temporal forecasting is extremely difficult due to the effect of noise or outliers \cite{lou2020probabilistic}. Discriminative algorithms learn $p(\mathbf{A}|\mathbf{X})$ directly from the node signals and then construct the dynamic graph, which does not need to model the distribution of the observed variables. They can not generally express complex relationships between the observed and target variables. Hence, certain deviations exist in practical applications. Inspired by the stochastic diffusion process in the physics domain, this paper proposes a novel unsupervised generative model (DVGNN) to construct the dynamic graph, where the interaction of neighbour nodes is considered as a stochastic diffusion process. The model encodes the latent variable by two-layer GCNs and infers the causal relationship of hidden variables at different times by SDE of the diffusion process in the decoder stage. Consequently, more explanatory and robust results can be obtained in this way.

\section{Preliminaries}
In order to facilitate the subsequent discussion, we introduce key definitions as follows: 
\begin{definition}\label{definition_1}   
\textit{\textbf{Graph} $\boldsymbol{\mathcal{G}}$}. Most spatio-temporal networks can be  denoted as a graph $\mathbf{\mathcal{G}} = (\mathbf{V},\mathbf{E},\mathbf{A})$, where $\mathbf{V}$ is a set of nodes in the graph, $\mathbf{V}=\{v_1,v_2,...,v_N\}$ and $N$ is the total number of nodes in the graph. $\mathbf{E}$ is a set of edges between the nodes. We adopt the adjacency matrix $\mathbf{A}$ to represent the connections between the nodes. To facilitate the subsequent discussion, $\Bar{\mathbf{A}}$ is defined to represent the graph structure of the observational data for the encoder stage, and $\mathbf{A}_{causal}$ is used to represent the causal relationship graph, while $\mathbf{A}_{trans}$ for the transition density probability graph for the multi-step prediction. In particular, $\Bar{\mathbf{A}}$, $\mathbf{A}_{causal}$  and $\mathbf{A}_{trans}$ $\in \mathbb{R}^{N \times N} $ for a weighted graph.  \end{definition}

\begin{definition} 
\label{definition_2}    
\textit{\textbf{Feature matrix} $\boldsymbol{X}$}. Node information contains various information, which can be denoted as a matrix, $\mathbb{X} \in \mathbb{R}^{N \times F}$, where $F$ is the dimension of input features. In this study, to represent the feature of time series, $\mathbf{X}_{t} \in \mathbb{R}^{N \times F}$ is used to denote all the input features of all the nodes $\mathbf{V}$ at time $t$. Thus, the values of the historical $p$ time steps of every attribute form the feature matrix $\mathbf{X}_{t-p:t-1} = [X_{t-p},...,X_{t-2},X_{t-1}]$, correspondingly $\mathbf{X}_{t-p:t-1} \in \mathbb{R}^{N \times F\times p}$. 
\end{definition}

In this paper, the goal of spatio-temporal forecasting with observational graph structure data is to infer the causal graph of time series and predict the future $T'$ time steps based on the past $p$ historical time steps. It can be formulated as follows: 
\begin{equation}
\label{Eq:problem definition1}     \left [ X_{t-p} ,...X_{t-1};\mathbf{\Bar{A}}\right ]\xrightarrow{g(.)} \left [ \mathbf{A}_{causal}^{t-p+1:t-1} \right ]. 
\end{equation}

\begin{equation} 
\label{Eq:problem definition3}   \left [ \mathbf{A}_{causal}^{t-p+1:t-1}\right ]\xrightarrow{} \left [ \mathbf{A}_{trans}^{t-p+1:t-1} \right ].  \end{equation}  \begin{equation}\label{Eq:problem definition2}     \left [ \mathbf{X}_{t-p+1} ,...X_{t-1};\mathbf{A}_{trans}^{t-p+1:t-1}\right ]\xrightarrow{f(.)} \left [ \mathbf{X}_{t},...,\mathbf{X}_{t+T'-1} \right ].
\end{equation}

 \begin{figure*}[h!]     
 \centering     
 \includegraphics[width=\linewidth]{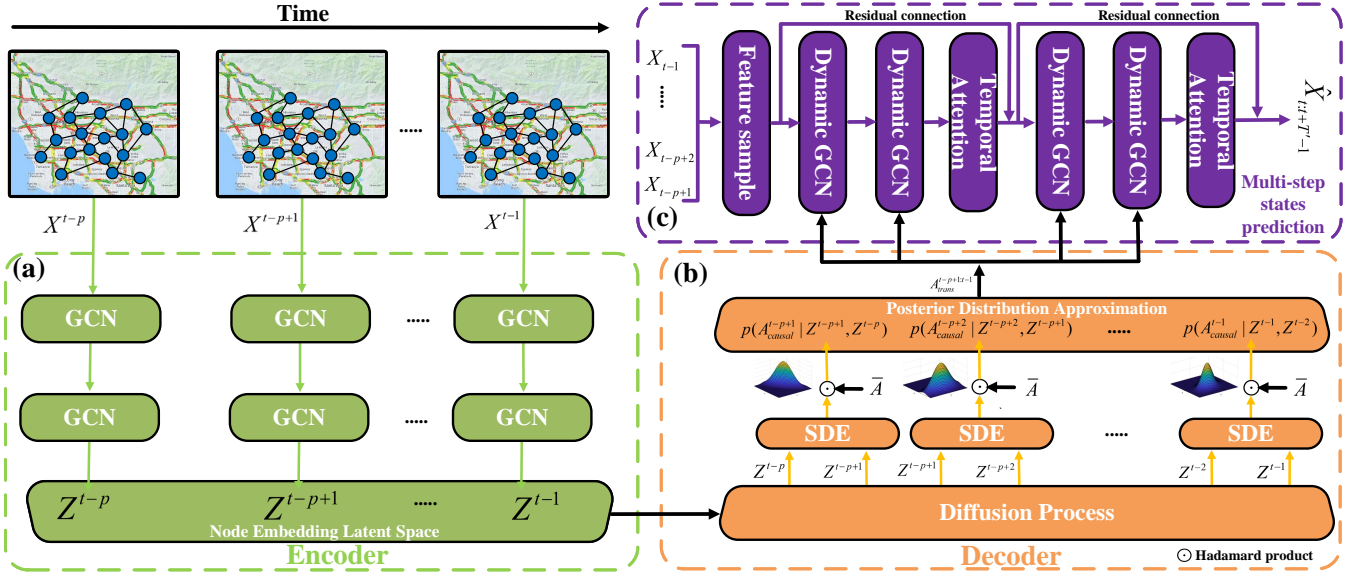}  
 \caption{Overview of the DVGNN architecture: (a) is the encoder component, consisting of two GCN layers, which is applied to learn latent node embeddings $\mathbf{Z}^t$. (b) is the decoder component, processing the latent variables $\mathbf{Z}^t$ and generating dynamic graphs by SDE. (c) is the multi-step states prediction component, consisting of dynamic GCNs and temporal attention to capture the spatial-temporal features of graph node signals and predict the multi-step states in the future.}   
 \label{diagram}  
 \end{figure*}

\section{Network architecture} 
The overview framework of DVGNN is shown in Fig. \ref{diagram}. In the encoder stage, the feature matrix of historical $p$ time steps $\mathbf{X}_{t-p:t-1}$ are fed into two GCN layers to learn the mean vectors $\mathbf{u}$ and variance vectors $\mathbf{log\sigma}$ of the latent variables $\mathbf{Z}_{t-p:t-1}$, all the time series are sharing the same parameters of GCNs, while $\mathbf{u}$ and $\mathbf{log\sigma}$ only share the first layer GCN parameters but keep their own second layer parameters. Therefore, the posterior probability distribution of latent states can be obtained in this way. In the decoder stage, the relationship of the latent states is regarded as a diffusion process and subjected to the SDE, which is considered as a priori that the transitions of the sequence latent states conform to the Markov and Gaussian process. The internal causal relationship of neighbour nodes can be inferred with this adaptive mechanism. Thus, the dynamic link probability can be estimated by approximation, and the dynamic causal graphs $A_{causal}$ are formed as $p_{\theta}(\mathbf{A}_{causal}^t|\mathbf{Z}^t,\mathbf{Z}^{t-1})$. To reduce the randomness of the predicted links, regularization is applied by the Hadamard product between the dynamic graphs and the static graph. In order to facilitate the subsequent discussion, DVGNN-mask is applied to represent the model with the Hadamard product, while DVGNN is without the Hadamard product. Then the transition graphs of transition density probabilities $A_{trans}^{t-p:t-1}$ are calculated by Gaussian distribution. Finally, $A_{trans}^{t-p:t-1}$, together with the historical $p-1$ time series feature matrix, are sampled and input to Dynamic GCNs for capturing the spatial features, while temporal attention is adopted to capture the temporal features. At the end of the neural network, residual connections are applied to produce the predicted value $\mathbf{\hat{X}}_{t:t+{T}'-1}$. 
\subsection{\textbf{Graph Convolution Network}} 
A GCN defines a first-order approximation based on the localized spectral filter on graphs. GCN can be formulated as \cite{kipf2016semi}:
\begin{equation}
\label{gcn}
\mathbf{h}^{(l+1)}=GCN\left(\mathbf{A}, \mathbf{h}^{(l)}\right)=\sigma\left(  \mathbf{L h}^{(l)} \mathbf{w}^{(l)}\right), 
\end{equation} 
where Laplacian matrix $\mathbf{L}=\mathbf{D}^{-\frac{1}{2}} (\mathbf{A}+\mathbf{I})\mathbf{D}^{\frac{1}{2}}$, $\mathbf{I}$ is the identity matrix, $\mathbf{D}$ is the degree matrix.  $\mathbf{h}^{(l)} \in \mathbb{R}^{N \times F}$ is the input of $l$ layer, while $\mathbf{h}^{(l+1)} \in \mathbb{R}^{N \times E}$ is output of $l$ layer with embedding features $E$. In addition, $\mathbf{w} \in \mathbb{R}^{F \times E}$ is the learning parameter, and $\sigma$ is the activation function.

\subsection{Variational Graph Autoencoder} 
Variational Graph Autoencoder (VGAE) is a powerful generative and unsupervised learning method in link prediction and graph generation tasks. VGAE applies the encoder to project the prior into a latent space from which the decoder can sample \cite{kipf2016variational}. VGAE consists of two models called the inference and generative model or encoder and decoder stages. 

In the graph inference model, the posterior distribution of latent variable $\mathbf{Z}$ is approximated through a Gaussian distribution by two-layer GCNs as following Equation \cite{kipf2016variational}:
\begin{equation} 
\label{equation 1}    
\begin{aligned}     & q_{\phi }(\mathbf{Z}|\mathbf{X})=\prod_{i=1}^{N}q_{\phi }(z_{i}|\mathbf{X}, \mathbf{\bar{A}})\\      & \mathrm{with} \quad q_{\phi }(z_{i}|\mathbf{X},\mathbf{\bar{A}})=\mathcal{N}(z_{i}|u_{i}, diag(\sigma_{i}^{2})),     \end{aligned} 
\end{equation}
where $\mathbf{u}=GCN_{\mathbf{u}}(\mathbf{X},\mathbf{\bar{A}}), \quad \mathbf{\log \sigma} = GCN_{\mathbf{\sigma}}(\mathbf{X},\mathbf{\bar{A}})$. In this paper, GCN layers are defined as $GCN(\mathbf{X},\mathbf{\bar{A}}) = Sigmoid (\mathbf{L} (ReLU(\mathbf{LXW_{0}))W_{1}})$. 

In the generative model, the link probability of the graph is given by an inner product between latent variables as follows:
\begin{equation}
\begin{aligned} 
& p(\mathbf{A}_{causal}|\mathbf{Z})=\prod_{i=1}^{N}\prod_{j=1}^{N}p(A_{ij}|z_i,z_j),  \\  & \mathrm{with} \quad p(A_{ij}=1|z_i,z_j)=\sigma(z_i^{T}z_j),   
\end{aligned}
\end{equation}
where $A_{ij}$ are the elements of $\mathbf{A}_{causal}$, and $\sigma$ is the activation function such as the sigmoid or $ReLU$ function.

\subsection{VGAE2Seq model} 
VGAE is a powerful unsupervised and generative learning tool in link prediction, but it is not suitable to deal with time series problems. To deal with the time series data, we extend the previous Definition \ref{equation 1} in the encoder stage as follows:

\begin{equation} 
\label{encoder}    
\begin{aligned}     & q_{\phi }(\mathbf{Z}^t|\mathbf{X}^{t})=\prod_{i=1}^{N}q_{\phi }(z_{i}^{t}|\mathbf{X}^{t}, \Bar{\mathbf{A}}),\\     &\mathrm{with} \quad q_{\phi }(z_{i}^t|\mathbf{X}^t,\Bar{\mathbf{A}})=\mathcal{N}(z_{i}^{t}|u_{i}^{t}, diag(\sigma_{i}^{t})^{2}).     
\end{aligned} 
\end{equation}

Also, we assume that the latent variable $\mathbf{Z}^t$ obeys the Gaussian distribution. In the decoder stage, traditional VGAE belongs to a non-probabilistic variant, which predicts the link only by an inner product between latent variables. This graph generative method is unsuitable for the spatio-temporal data, since the link pattern between nodes can not be simply classified by the similarity of the latent variable of neighbour nodes. Thus, the traditional VGAE method is limited to exploiting spatial-temporal features such as time series with complex graph structures. In order to discover the dynamic causal relationship of the graph, inspired by the Bayesian framework and assuming that all the edges are independent of each other, the dynamic link probability can be formulated as follows:
\begin{equation} 
\label{transition density} 
\begin{aligned}     
& p_{\theta}(\mathbf{A}_{causal}^t|\mathbf{Z}^t,\mathbf{Z}^{t-1})=p_{\theta}(\mathbf{Z}^t,\mathbf{Z}^{t-1}),\\      & \mathrm{with} \quad p_{\theta}(\mathbf{Z}^t,\mathbf{Z}^{t-1})=\prod_{i}\prod_{j}p_{\theta}(z_j^t,z_i^{t-1}). \end{aligned} 
\end{equation}

Similar to the traditional VGAE, we assume that the latent variables $\mathbf{Z}^t$ obey standard Gaussian prior distribution at each time interval $p(\mathbf{Z}^t)=\prod_{i}p(z_i^t)=\prod_{i}\mathcal{N}(z_i|0,1)$. In the same way as VGAE, the time sequence VGAE maximizes the Evidence Lower Bound (ELBO) to learn the parameters $\theta$ and $\phi$:
\begin{equation}    
\begin{aligned}     
\mathcal{L} & =\mathbb{E}_{q_{\phi}(\mathbf{Z}^t|\mathbf{X}^t,\Bar{\mathbf{A}})}\mathbb{E}_{q_{\phi}(\mathbf{Z}^{t-1}|\mathbf{X}^{t-1},\Bar{\mathbf{A}})}[\log p_{\theta}(\mathbf{Z}^t,\mathbf{Z}^{t-1})]\\     &-\mathbf{KL}[q_{\phi}(\mathbf{Z}^t|\mathbf{X}^t,\Bar{\mathbf{A}})||p(\mathbf{Z}^t)]\\     &-\mathbf{KL}[q_{\phi}(\mathbf{Z}^{t-1}|\mathbf{X}^{t-1},\Bar{\mathbf{A}})||p(\mathbf{Z}^{t-1})],     \end{aligned} 
\end{equation}
where $\mathbf{KL}[q_{\phi}(\cdot)||p(\cdot)]$ is the Kullback-Leibler divergence between $q_{\phi}(\cdot)$ and $p(\cdot)$. Considering the $p$ time series, the total objective function in the training stage will be:
\begin{equation}
\label{objective fuction4}   
\begin{aligned}     
\mathcal{L}= \sum_{t=2}^{t=p}&\mathbb{E}_{q_{\phi}(\mathbf{Z}^t|\mathbf{X}^t,\Bar{\mathbf{A}})}\mathbb{E}_{q_{\phi}(\mathbf{Z}^{t-1}|\mathbf{X}^{t-1},\Bar{\mathbf{A}})}[\log p_{\theta}(\mathbf{Z}^t,\mathbf{Z}^{t-1})]-\\      \mathbf{KL}[q_{\phi}(\mathbf{Z}&^1|\mathbf{X}^1,\Bar{\mathbf{A}})||p(\mathbf{Z}^1)]-\mathbf{KL}[q_{\phi}(\mathbf{Z}^p|\mathbf{X}^p,\Bar{\mathbf{A}})||p(\mathbf{Z}^p)] \\     &-2\sum_{t=2}^{t=p-1}\mathbf{KL}[q_{\phi}(\mathbf{Z}^t|\mathbf{X}^t,\Bar{\mathbf{A}})||p(\mathbf{Z}^t)].     \end{aligned}
\end{equation}

We develop a new VGAE to deal with the spatial-temporal data, but how to find out the link probability of the neighbour nodes becomes the key problem.

\subsection{Dynamic link probability with diffusion process}
Time series data in real scenarios is inevitably uncertain and accompanied by noise caused by the data-collecting sensors, which is ubiquitous in the data-collecting stage. Therefore, taking the interaction of neighbour nodes as a stochastic process will be more appropriate to determine their causal relationships of them, which are the key in the decoder stage. Inspired by the diffusion model in the physics domain, we introduce the inference diffusion process in the decoder stage. This diffusion process can be modelled as the solution to a noise-driven Stochastic Differential Equation (SDE) \cite{sarkka2019applied}. As to the spatio-temporal forecasting problem, consider a linear stochastic differential equation of the general form as follows \cite{watson2021learning}:
\begin{equation} 
\label{SDE} d\mathbf{Z}^t=\mathbf{F}(t)\mathbf{Z}^tdt + \mathbf{u}(t)dt + \mathbf{L}(t)d\mathbf{\beta}(t), 
\end{equation}
where $\mathbf{Z}^t \in \mathbb{R}^N$ is the state, $\mathbf{F}(t)$ and $\mathbf{L}(t)$ are matrix valued function of time, $\mathbf{u}(t) \in \mathbb{R}^N$ is an input vector, and $\mathbf{\beta}(t) \in \mathbb{R}^N$ is a Brownian motion with diffusion matrix $\mathbf{Q}$. In this paper, we assume that the latent variable $\mathbf{Z}^t$ follow Markov process $p(\mathbf{Z}^t|\mathbf{Z}^{t-1},\mathbf{Z}^{t-2},...\mathbf{Z}^{1})=p(\mathbf{Z}^t|\mathbf{Z}^{t-1})$ and the input vector is zero ($\mathbf{u}(t)=0$) in the whole  graph. Then, the dynamic link probability of Equation \ref{transition density} will be regarded as the joint probability of $\mathbf{Z^t}$ and $\mathbf{Z^{t-1}}$ if we take the latent variable $\mathbf{Z}$ as a diffusion process which is governed by Equation \ref{SDE}. The transition density $p_\theta(\mathbf{Z}^t|\mathbf{Z}^{t-1})$ obeys Gaussian Process (GP). Thus the solution is a linear transformation of Brownian motion, which is a Gaussian distribution. The solution can be formulated as \cite{oksendal2003stochastic}:
\begin{equation}      
p_{\theta}(\mathbf{A}_{causal}^t|\mathbf{Z}^t,\mathbf{Z}^{t-1}) = p_{\theta}(\mathbf{Z}^t,\mathbf{Z}^{t-1})=\mathcal{N}(\mathbf{Z}^{t},\mathbf{Z}^{t-1}) 
\label{causal eq}
\end{equation}
\begin{equation}
\label{gp trans} 
\begin{aligned}     
& \mathbf{A}_{trans}=p_\theta(\mathbf{Z}^t|\mathbf{Z}^{t-1})=\mathcal{N}(\mathbf{\Psi_\theta(t,t-1)}\mathbf{Z}^{t-1},\mathbf{\Sigma}_\theta)\\       & with \quad \mathbf{\Sigma}_\theta=\int_{t-1}^t\mathbf{\Psi_\theta(t,\tau)}\mathbf{L}_\theta(\tau)\mathbf{Q}_\theta\mathbf{L}_\theta^T(\tau)\mathbf{\Psi}_\theta^T(t,\tau)d\tau),  
\end{aligned}
\end{equation}
where $\mathbf{\Sigma}_\theta$ is the covariance matrix of the Gaussian process, $\mathbf{Q}=\mathbf{L_\theta L_\theta^T}$ is the diffusion matrix and $\mathbf{\Psi}$ is the transition matrix, which does not have a closed-form expression in general but is defined via the properties:
\begin{equation} 
\label{property of transition matrix}   
\begin{aligned}     \frac{\partial \mathbf{\Psi}(\tau,t)}{\partial \tau}&=\mathbf{F}(\tau)\mathbf{\Psi}(\tau,t)\\     \frac{\partial \mathbf{\Psi}(\tau,t)}{\partial \tau}&=-\mathbf{\Psi}(\tau,t)\mathbf{F}(t)\\     \mathbf{\Psi}(t,t)&=\mathbf{I}.     \end{aligned}
\end{equation}

To simplify the calculation, $\mathbf{L}_\theta(t)=\sqrt{2}\delta_{ij}$ is used as most models often adopted, where $\delta_{ij}$ is the Kronecker delta function \cite{rezende2015variational}. Then the variance of Equation \ref{gp trans} can be deduced as follows:
\begin{equation} 
\begin{aligned}
\mathbf{\Sigma}_\theta=&\int_{t-1}^t\mathbf{\Psi_\theta(t,\tau)}\mathbf{L}_\theta(\tau)\mathbf{Q}_\theta\mathbf{L}_\theta^T(\tau)\mathbf{\Psi}_\theta^T(t,\tau)d\tau\\
&=\int_{t-1}^t4\mathbf{\Psi}_\theta(t,\tau)\mathbf{\Psi}_\theta^T(t,\tau)d\tau. 
\end{aligned}
\end{equation}

Considering the discrete data of time series data and the property of Equation \ref{property of transition matrix}, the variance of Equation \ref{gp trans} can be approximated as:

\begin{equation}   
\begin{aligned}  
&\mathbf{\Sigma}_\theta = \int_{t-1}^t4\mathbf{\Psi}_\theta(t,\tau)\mathbf{\Psi}_\theta^T(t,\tau)d\tau\\
&\approx 2(\mathbf{I}+\mathbf{\Psi}_\theta(t,t-1)\mathbf{\Psi}_\theta^T(t,t-1)).
\end{aligned}
\end{equation}

This further leads to the approximation of the link prediction that can be formulated as follows:

\begin{equation}
\label{approx A}   p_{\theta}(\mathbf{Z}^t,\mathbf{Z}^{t-1}) \approx \mathcal{N}(\mathbf{Z}^{t},\mathbf{Z}^{t-1},\mathbf{\Sigma}_\theta). 
\end{equation}

Substituting Equation \ref{approx A} into Equation \ref{objective fuction4}, the objective function can be formulated as follows:
\begin{equation} 
\label{objective fuction1}    
\begin{aligned}     &\mathcal{L} \approx \sum_{t=2}^{t=p}\mathbb{E}_{q_{\phi}(\mathbf{Z}^t|\mathbf{X}^t,\Bar{\mathbf{A}})}\mathbb{E}_{q_{\phi}(\mathbf{Z}^{t-1}|\mathbf{X}^{t-1},\Bar{\mathbf{A}})}[\log \mathcal{N}(\mathbf{Z}^{t},\mathbf{Z}^{t-1},\mathbf{\Sigma}_\theta)]\\     &-\mathbf{KL}[q_{\phi}(\mathbf{Z}^1|\mathbf{X}^1,\Bar{\mathbf{A}})||p(\mathbf{Z}^1)]-\mathbf{KL}[q_{\phi}(\mathbf{Z}^p|\mathbf{X}^p,\Bar{\mathbf{A}})||p(\mathbf{Z}^p)]\\     &-2\sum_{t=2}^{t=p-1}\mathbf{KL}[q_{\phi}(\mathbf{Z}^t|\mathbf{X}^t,\Bar{\mathbf{A}})||p(\mathbf{Z}^t)].     \end{aligned}
\end{equation}
To facilitate the subsequent discussion, we change the objective function into the form of matrix elements, considering that all the edges are independent of each other as Equation~\ref{transition density}. 
\begin{equation}
\label{objective fuction2}   
\begin{aligned}     \mathcal{L} &\approx \sum_{t=2}^{t=p} \sum_j^N \sum_i^N \mathbb{E}_{q_{\phi} (z_j^t|\mathbf{X}^t,\Bar{\mathbf{A}})}\mathbb{E}_{q_{\phi}(z_i^{t-1}|\mathbf{X}^{t-1},\Bar{\mathbf{A}})} [-\log{2\pi}-0.5\\     &\log{(\sigma_{i}^2\sigma_{j}^2-\Sigma_{ij}^2)}-\frac{\sigma_{j}^2(z_i^{t-1}-\mu_i^{t-1})^2}{2(\sigma_{i}^2\sigma_{j}^2-\Sigma_{ij}^2)}-\frac{\sigma_{i}^2(z_j^{t}-\mu_j^{t})^2}{2(\sigma_{i}^2\sigma_{j}^2-\Sigma_{ij}^2)}+\\     &\frac{\Sigma_{ij}(z_i^{t-1}-\mu_i^{t-1})(z_j^{t}-\mu_j^{t})}{(\sigma_{i}^2\sigma_{j}^2-\Sigma_{ij}^2)}]-\sum_{i}^{N} [\mathbf{KL}(q_{\phi}(z_i^1|\mathbf{X}^1,\Bar{\mathbf{A}})||p(z_i^1))-\\     &\mathbf{KL}(q_{\phi}(z_i^p|\mathbf{X}^p,\Bar{\mathbf{A}})||p(z^p))-2\sum_{t=2}^{t=p-1}\mathbf{KL}[q_{\phi}(z_i^t|\mathbf{X}^t,\Bar{\mathbf{A}})||p(z_i^t)].   
\end{aligned}
\end{equation}
However, stochastic gradient descent via backpropagation can handle stochastic inputs, but not stochastic units within the network. The solution, called the “reparameterization trick” \cite{kingma2013auto}, is to move the sampling to an input layer. We reparameterize the latent variable in terms of a known base distribution as follows:
\begin{equation} 
\label{reparameterization} 
\begin{aligned}      z_i^{t-1} & =u_i+\sigma_i\epsilon_i,\quad \epsilon_i \sim \mathcal{N}(0,1).\\     z_j^t & =u_j+\sigma_j\epsilon_j,\quad \epsilon_j \sim \mathcal{N}(0,1).    
\end{aligned} 
\end{equation}
Then, the expectation of the objective function in Equation \ref{objective fuction2} can be simplified as:
\begin{equation} 
\label{reparameterization2}    
\begin{aligned}    
&\mathbb{E}_{q_{\phi}(z_j^t|\mathbf{X}^t,\mathbf{\Bar{A}})}\mathbb{E}_{q_{\phi}(z_i^{t-1}|\mathbf{X}^{t-1},\mathbf{\Bar{A}})}     [-\frac{\sigma_{j}^2(z_i^{t-1}-\mu_i^{t-1})^2}{2(\sigma_{i}^2\sigma_{j}^2-\Sigma_{ij}^2)}\\ 
&-\frac{\sigma_{i}^2(z_j^{t}-\mu_j^{t})^2}{2(\sigma_{i}^2\sigma_{j}^2-\Sigma_{ij}^2)}+\frac{\Sigma_{ij}(z_i^{t-1}-\mu_i^{t-1})(z_j^{t}-\mu_j^{t})}{(\sigma_{i}^2\sigma_{j}^2-\Sigma_{ij}^2)}] \\  
& =\mathbb{E}_{\epsilon_{i} \sim \mathcal{N}(0,1)}\mathbb{E}_{\epsilon_{j} \sim \mathcal{N}(0,1)} [\frac{2\Sigma_{ij}\sigma_i\sigma_j\epsilon_i\epsilon_j-\sigma_{i}^2\sigma_{j}^2(\epsilon_i^2+\epsilon_j^2)}{2(\sigma_{i}^2\sigma_{j}^2-\Sigma_{ij}^2)}].    
\end{aligned} 
\end{equation}
 In addition, we add 0.0001 to Equation \ref{objective fuction2} to avoid the NaN problem. Then the objective function can be obtained by substituting Equation \ref{objective fuction2} by reparameterization of Equation \ref{reparameterization2} as:
 \begin{equation}    
 \label{objective fuction3}   
 \begin{aligned}      
 &\mathcal{L} \approx \sum_{t=2}^{t=p} \sum_j^N \sum_i^N \mathbb{E}_{\epsilon_{i}}\mathbb{E}_{\epsilon_{j}}[-\log{2\pi}-0.5\log{(\sigma_{i}^2\sigma_{j}^2-\Sigma_{ij}^2}\\     
 &+0.0001)+\frac{2\Sigma_{ij}\sigma_i\sigma_j\epsilon_i\epsilon_j-\sigma_{i}^2\sigma_{j}^2(\epsilon_i^2+\epsilon_j^2)}{2(\sigma_{i}^2\sigma_{j}^2-\Sigma_{ij}^2+0.0001)}]-\\
 &\sum_{i}^{N} [\mathbf{KL}(q_{\phi}(z_i^1|\mathbf{X}^1,\Bar{\mathbf{A}})||p(z_i^1))-\mathbf{KL}(q_{\phi}(z_i^p|\mathbf{X}^p,\Bar{\mathbf{A}})||p(z^p))\\
 &-2\sum_{t=2}^{t=p-1}\mathbf{KL}[q_{\phi}(z_i^t|\mathbf{X}^t,\Bar{\mathbf{A}})||p(z_i^t)].    
 \end{aligned} 
 \end{equation}
 Finally, we will optimize the objective function of Equation \ref{objective fuction3} to learn and predict the dynamic causal graphs at the first step.
\subsection{Temporal attention} 
To capture the temporal features, the attention mechanism is applied. Attention is a powerful tool, which has been widely adopted by many famous models such as ASTGCN. It is proven that temporal attention can effectively explore the temporal correlations under different situations in different time slices adaptively. The mechanism of its operation can be expressed as \cite{guo2019attention}:
\begin{equation} 
\label{eq11}  \mathbf{E}=\mathbf{V}_e \sigma ((\mathbf{h}^{T} \mathbf{U}_1)\mathbf{U_2} (\mathbf{U_3}\mathbf{h})+\mathbf{b}_e) 
\end{equation}  

\begin{equation} 
\label{eq12} \mathbf{E}_{i,j}^{'} = \frac{\exp(\mathbf{E}_{i,j})}{\sum_{j=1}^{Tr}\exp(\mathbf{E}_{i,j})}
\end{equation}  
where $\mathbf{V}_e$, $\mathbf{b}_e \in \mathbb{R}^{T_r \times T_r }$, $\mathbf{U}_1 \in \mathbb{R}^{N}$, $\mathbf{U}_2 \in \mathbb{R}^{F \times N}$, $\mathbf{U}_3 \in \mathbb{R}^{F}$ are learnable parameters, $\sigma(.)$ is the activation function and $T_r$ is the length of the temporal dimension, $\mathbf{h}$ represents the output of the two dynamic GCN layers, the temporal semantic dependencies can be represented by the value of $\mathbf{E}_{i,j}$ by Equation \ref{eq11}. In addition, $\mathbf{E}^{'}$ is a normalized temporal attention matrix by the softmax function. Then, the output of temporal attention is calculated as $\mathbf{E}^{'}\mathbf{h}$.

In sum, with the definitions and mathematical reasoning explained above, the input time series feature $\mathbf{X}$, as well as the dynamically generated transition graphs $\mathbf{A}_{trans}$ by Equation \ref{gp trans}, is fed to dynamic GCNs to capture the spatial features, then the output of dynamic GCNs is input to the temporal attention to capture the temporal features and predict the future states.

Also, mask means the parameter covariance matrix $\mathbf{\Sigma}_\theta$ makes Hadarmard product with the pre-defined adjacent matrix $\bar{\mathbf{A}}$ (where connected link equals 1, otherwise will be 0) $\mathbf{\Sigma}_\theta = \mathbf{\Sigma}_\theta \odot \bar{\mathbf{A}}$, since we assume that only connected nodes can get a chance to interact with each other directly while not connected nodes can not, if the dataset owns pre-defined graph structure. As a result, mask employment is akin to applying the regularization on the parameter $\mathbf{\Sigma}_\theta$. Algorithm \ref{algorithm} summarizes the procedure of DVGNN.

\begin{algorithm}[]   
\SetKwInOut{Input}{Input}   
\Input{historical feature matrix $\mathbf{X}_{t-p:t-1}$ and pre-defined adjacent matrix $\Bar{\mathbf{A}}$}     
\SetKwInOut{Output}{Output}  
\Output{predicted future states $\mathbf{\hat{X}}_{t:t+T'-1}$,  causal graph adjacent matrix $\mathbf{\hat{A}}_{causal}^{t}$ and model parameters$\mathbf{\Theta}=$ \{$\phi,\theta,\mathbf{w1},\mathbf{w2}$\} }  
Randomly initialize $\mathbf{\Theta}$;\\  
\For{each epoch}{ 
\For{each batch of $\mathbf{X}_{t-p:t-1}$ in dataset}{ 
/*\quad  causal graph generation  \quad  \quad   */\\ 
\For{t=$(t-p+1) \rightarrow (t-1)$}{ 
/*\quad  Encoder stage  \quad  \quad \quad \quad \quad    */\\ 
Compute $q_{\phi }(\mathbf{Z}^t|\mathbf{X}^{^{t}})$ and $q_{\phi }(\mathbf{Z}^{t-1}|\mathbf{X}^{^{t-1}})$ by Eq.\ref{encoder};\\ 
Reparameterize $\mathbf{Z}^{t}$ and $\mathbf{Z}^{t-1}$ by Eq. \ref{reparameterization};\\    
/*\quad  Decoder stage  \quad  \quad \quad \quad \quad    */\\ 
\If{$MASK$} 
{   
$\mathbf{\Sigma}_\theta = \mathbf{\Sigma}_\theta \odot \bar{\mathbf{A}}$; 
}     
Approximate $p_{\theta}(\mathbf{Z}^{t},\mathbf{Z}^{t-1})$ by Eq. \ref{approx A}; \\  
Optimize approximate objective function $\mathcal{L}$ in Eq. \ref{objective fuction3};\\ Update parameters $\phi,\theta$;  
}   
Infer causal graph $\mathbf{\hat{A}}_{causal}^{t-p+1:t-1}$ by Eq. \ref{causal eq};\\      
Calculate transition density probability graph        
$\mathbf{\hat{A}}_{trans}^{t-p+1:t-1}$ by Eq. \ref{gp trans};\\  
/*\quad  Future states prediction  \quad  \quad \quad */\\     
Calculate Laplacian matrix $\mathbf{\hat{L}}^{t-p+1:t-1}$ with $\mathbf{\hat{A}}_{trans}^{t-p+1:t-1}$;\\    
Feed $\mathbf{X}_{t-p+1:t-1}$ and Laplacian matrix $\mathbf{\hat{L}}^{t-p+1:t-1}$ to dynamic GCNs with $\mathbf{w1}$ ;\\    
Feed output of dynamic GCNs to temporal attention with $\mathbf{w2}$ with residual connection and obtain predicted $\mathbf{\hat{X}}_{t:t+{T}'-1}$;\\ 
Update $\mathbf{w1,w2}$ by minimize $L2$ loss $\left \| \mathbf{X}_{t:t+{T}'-1} -\mathbf{\hat{X}}_{t:t+{T}'-1}\right \|_2$;\\ 
}
}
return predicted result $\mathbf{\hat{X}}$, $\mathbf{\hat{A}}_{causal}^{t}$ and parameters $\mathbf{\Theta}$;  \\   
\caption{DVGNN algorithm}   
\label{algorithm} 
\end{algorithm}

\section{Experiments}
\subsection{Datasets}
To evaluate the performance of DVGNN, four public datasets of different domains with three different graph structures are collected. PeMS08 and Los-loop are the famous datasets of non-Euclidean structure with long and short time series in the transportation domain, while T-drive is a 2D grid dataset of Euclidean structure in the transportation domain. To test the generalization of our model, FMRI (Functional Magnetic Resonance Imaging) is used, which is a time series dataset with a non-pre-defined structured dataset in the healthcare domain. PeMS08 is released by the Caltrans Performance Measurement System (PeMS) \cite{chen2001freeway}, which samples the traffic flow every 5 minutes,  Los-loop is collected in the highway by loop detectors in Los Angeles, aggregated the traffic speed every 5 minutes and the missing values are supplemented by linear interpolation \cite{Zhao2020}. Since Los-loop dataset contains many edges of low value, we set the threshold as 0.5 to remove the edges of low value. T-drive dataset is collected by taxis GPS trajectory from Beijing city of China, which is released by Microsoft company \cite{yuan2011driving,yuan2010t}. Beijing city is divided into 8 $\times$ 8 grids, and the traffic flow speed is sampled every 5 minutes.

To test the generalization of our model in other domains, the FMRI dataset is used. FMRI contains realistic, simulated BOLD (Blood-oxygen-level dependent) datasets for 28 different underlying brain networks, which is introduced by FMRIB Analysis Group of Oxford and processed by Nauta \cite{nauta2019causal}. FMRI does not have an apparent spatial relationship but offers the ground-truth causal graph with a delay of 1 time step for evaluation. In this paper, we adopt the sub-dataset of FMRI-3, FMRI-4, and FMRI-13 of different nodes for the experiments. The detailed information of the four datasets is shown in Tables \ref{tab:dataset} and \ref{tab:FMRI}.
\begin{table}[]
\centering
\caption{Information of transportation datasets}
\label{tab:dataset} 
\begin{tabular}{ccccc}
\hline \multicolumn{1}{l}{Datasets} & \begin{tabular}[c]{@{}c@{}}Node Number\\ or Grid Number\end{tabular} & \begin{tabular}[c]{@{}c@{}}Edge\\ Number\end{tabular} & \begin{tabular}[c]{@{}c@{}}Time\\ Step\end{tabular} & \begin{tabular}[c]{@{}c@{}}Missing\\ Ratio\end{tabular} \\ \hline PeMS08                       & 170(roads)                                                                & 295                                                   & 17856                                               & 0.696\%                                                 \\ Los-loop                       & 207(roads)                                                                 & 1095                                                   & 2017                                               & 0\%                                                 \\ T-Drive                      & 64(grids)                                                                 & 481                                                   & 1767                                                & 2.794\%                                                 \\ \hline 
\end{tabular} 
\end{table}

\begin{table}[] 
\centering
\caption{Information of FMRI dataset} 
\label{tab:FMRI} 
\begin{tabular}{ccccc} \hline Dataset & Node Number & Time & \begin{tabular}[c]{@{}c@{}}Missing\\ Ratio\end{tabular} & \begin{tabular}[c]{@{}c@{}}Pre-defined\\ Graph\end{tabular} \\ \hline FMRI-3  & 15          & 200  & 0\%                                                     & None                                                \\ FMRI-4  & 50          & 200  & 0\%                                                     & None                                                \\ FMRI-13 & 5           & 200  & 0\%                                                     & None                                                \\ \hline 
\end{tabular} 
\end{table}

\subsection{Experiment Settings}
 All experiments are performed using TensorFlow 2.5.0 on a Linux server (CPU: Intel(R) Core(TM) i7-11800H @ 2.30GH, GPU: NVIDIA GeForce RTX 3080) with 32G memory. In the dynamic graph construction stage,  32 and 16 convolution kernels are used in the first and second GCNs layer for the encoder stage, while other datasets are 12 and 8. As for the pre-define graph-structure datasets of transportation, the dynamic graph construction module is trained on the incomplete graph structure, where 80\% edges are adopted as $\mathbf{\bar{A}}$ in the training period and 20\% edges are used to evaluate the dynamic links of the generated graphs, while 0\% for training and 100\% for evaluation to the FMRI dataset. The learning rate is set to 0.001, and $\log{\sigma_i\sigma_j}$ regularization is applied. For the multi-step states prediction stage, historical $p=60$ samples are used to predict the future states, the kernel size along the temporal dimension, and dynamic GCNs are set to 3. The learning rate is set to 0.0005. 

\subsection{Evaluation Metrics}
To evaluate the prediction performance, the following metrics are adopted. \begin{itemize}  
\item Root Mean Squared Error ($RMSE$) 
\begin{equation} 	
\label{rmse} 
RMSE=\sqrt{\frac{\sum_j^{T'}\sum_i^N (X_i^j-\hat{X}_i^j)^2}{T' \times N}} 	\end{equation}   
\item Mean Absolute Error ($MAE$) 
\begin{equation} 
\label{mae} 
MAE=\frac{\sum_j^{T'}\sum_i^N |X_i^j-\hat{X}_i^j|}{T' \times N} 	
\end{equation}   
\end{itemize}
where $\hat{X_i^j}$ and $X_i^j$ are the predicted value and ground truth value of $i$-th node at time $j$. $N$ is the total number of nodes in the graph, and $T'$ is the future time step that needs to be predicted. The smaller the $RMSE$ and $MAE$ values are, the more accurate the prediction is.

To evaluate the discovered causal relationships of generated causal graph, the following metrics are adopted.
\begin{itemize}    
\item \textit{Precision} 	
\begin{equation} 
\label{Precision}     
precision = \frac{TP}{TP+FP} 
\end{equation}  
\item \textit{F1-Score} 	
\begin{equation} 
\label{F1}       
F1 = \frac{2TP}{2TP+FP+FN}
\end{equation}   
\end{itemize}
where TP is the true positive while FP is the false positive. The larger the $precision$ and $F1$ values are, the more accurate the generated causal graphs are.

\subsection{Baselines}
The transportation datasets belong to the class of spatio-temporal datasets, which have a pre-defined graph structure. DVGNN is compared with the famous spatio-temporal forecasting models. The baseline A is described as follows:
\begin{itemize}     
\item T-GCN \cite{Zhao2020}, combines  GCN with GRU to exploit spatio-temporal dependencies for prediction.    
\item DCRNN \cite{li2017diffusion}, captures the spatial features by using bidirectional random walks on the graph while temporal features by  encoder-decoder architecture.     
\item ASTGCN \cite{guo2019attention}, uses temporal attention and spatial attention, together with GCN to predict the future states.   
\item STGODE \cite{fang2021spatial}, constructs GNN to capture spatio-temporal features through a tensor-based ordinary differential equation.   
\item DSTAGNN \cite{lan2022dstagnn}, utilizes spatial-temporal aware distance (STAD) and  multi-head attention to construct dynamic Laplacian matrices for prediction.    
\end{itemize}
The FMRI dataset belongs to the general class of time series data, which does not have a pre-defined graph structure. We compare our model with the latest causal discovery of time series models. The baseline B is described as follows: 

\begin{itemize}     
\item TCDF \cite{nauta2019causal}, applies CNN and attention mechanism to discover the causal graph of time series and prediction.
\item NGC \cite{tank2021neural}, models each time series by multilayer perceptron and LSTM for Granger causality and prediction. 
\item CR-VAE \cite{li2023causal}, integrates the concepts of Granger causality into a recurrent VAE framework for causal discovery and prediction.    
\end{itemize}

 We apply identity matrix $\mathbf{I}$ as the pre-define adjacent matrix $\mathbf{\Bar{A}}$ in training period, since it does not have a pre-define graph structure. The ground truth causal graphs are only applied to evaluate the performance of different models.

\subsection{Results}
We tested our model with the baselines on four datasets, and the results are shown in Tables \ref{result1} and \ref{result2}. Our DVGNN achieves the best accuracy. For the fast-changing spatio-temporal data, traditional hybrid GNN methods such as T-GCN and DCRNN use static graph, whose edges are fixed, and have the same aggregation effect as the fixed neighbour nodes with the time change, which is easy to ignore the effect of the dominant neighbour nodes. Hence, the methods can not better represent the dynamics of the spatio-temporal data. The dynamic graph construction methods can exploit the dynamic links or edges of graphs adaptively with time. As a result, their overall accuracy will be higher than the static graph methods. However, even though other dynamic graph methods can obtain high accuracy, it is hard to discover the causal relationship of interaction between neighbour nodes even using attention or multi-attention mechanism. Therefore, the performance and generalization capacity are easily affected by the changes of different datasets. But DVGNN achieves better robustness in the change of different graph-structure datasets even with no pre-defined graph-structure as the FMRI dataset shown in Table \ref{result2}. As for FMRI dataset, other causal discovery methods focus on the temporal features for prediction, and ignore the transition graphs of the nodes in the prediction step, which limit the improvement of prediction accuracy.

\begin{figure}[h!] 
\centering \subfloat[RMSE of Los-loop.]{ 
\centering 
\includegraphics[width=0.50\columnwidth]{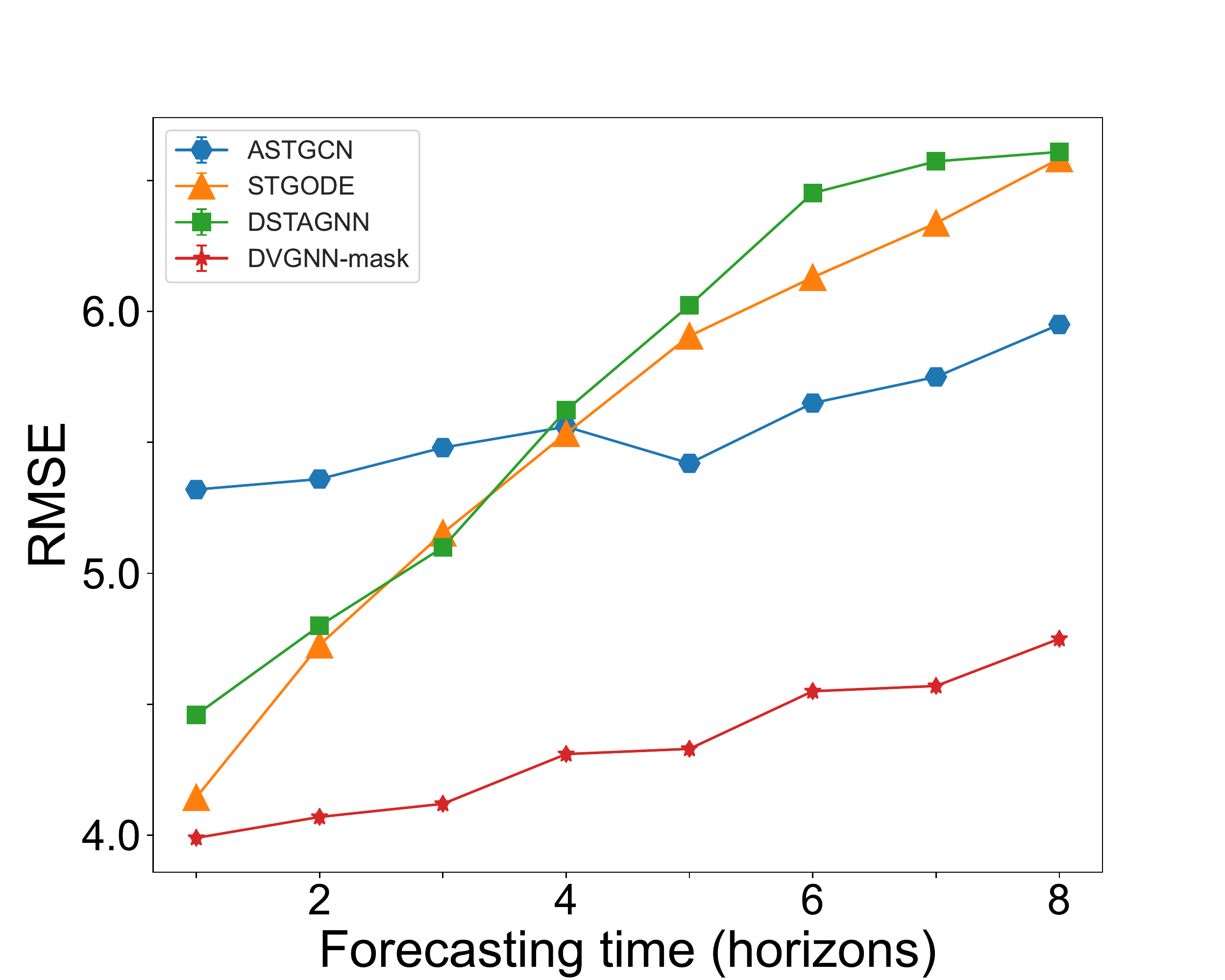} 	\label{baseline_curve_a} 
}
\subfloat[MAE of Los-loop.]{ 
\centering 
\includegraphics[width=0.50\columnwidth]{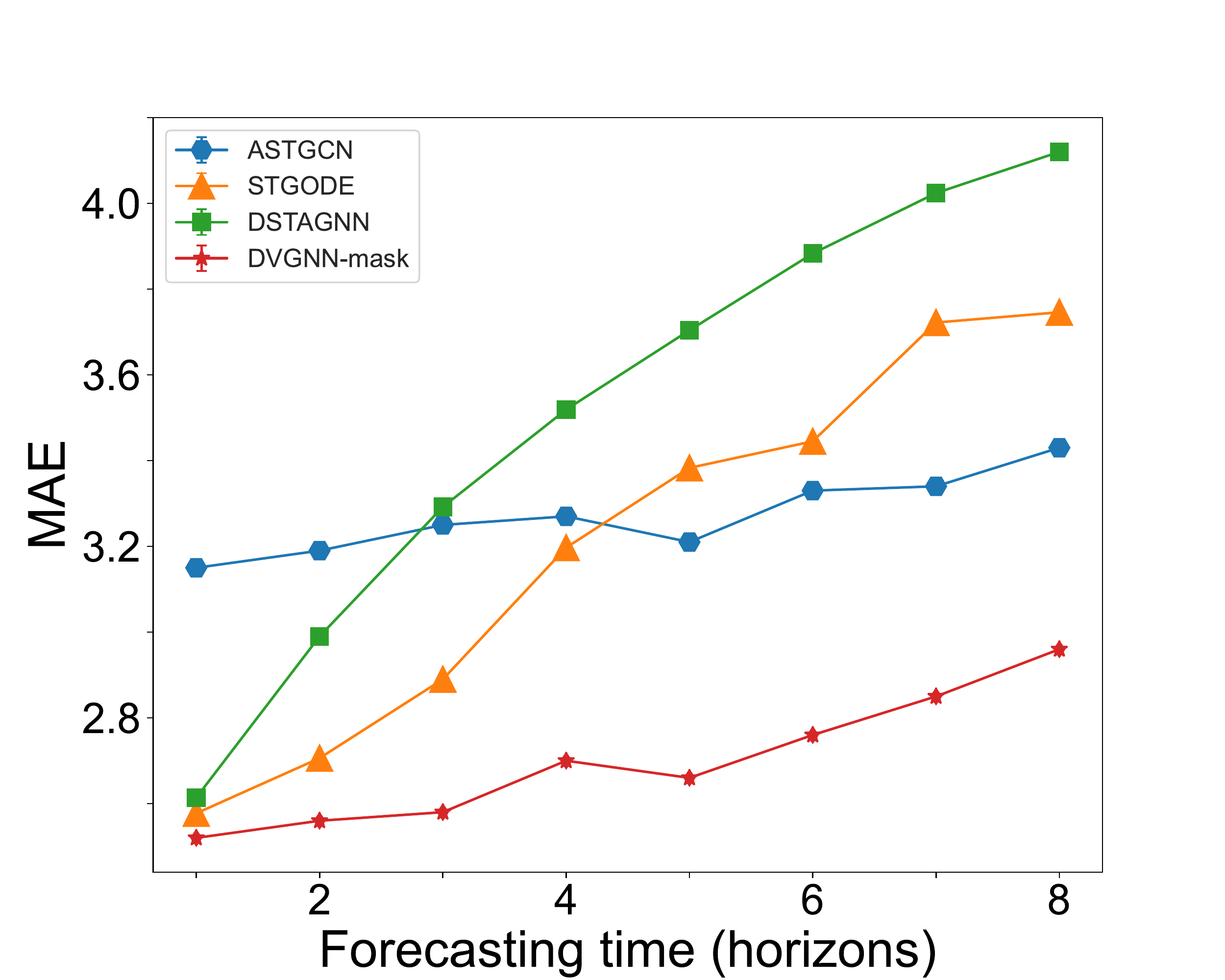} 
\label{baseline_curve_b}
} 

\subfloat[RMSE of PeMS08.]{ 
\centering 
\includegraphics[width=0.50\columnwidth]{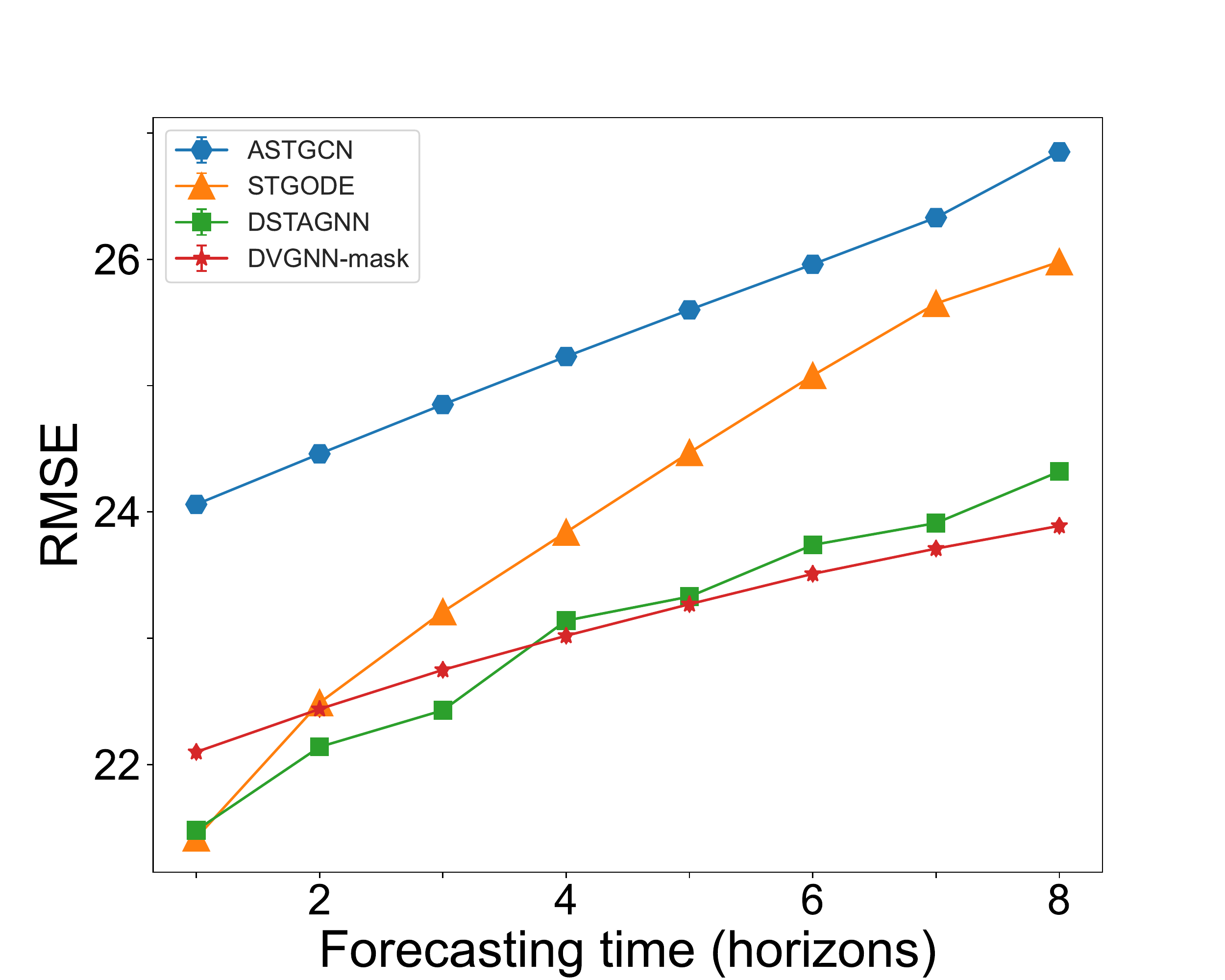} 	\label{baseline_curve_c} }
\subfloat[MAE of PeMS08.]{ 
\centering 	
\includegraphics[width=0.50\columnwidth]{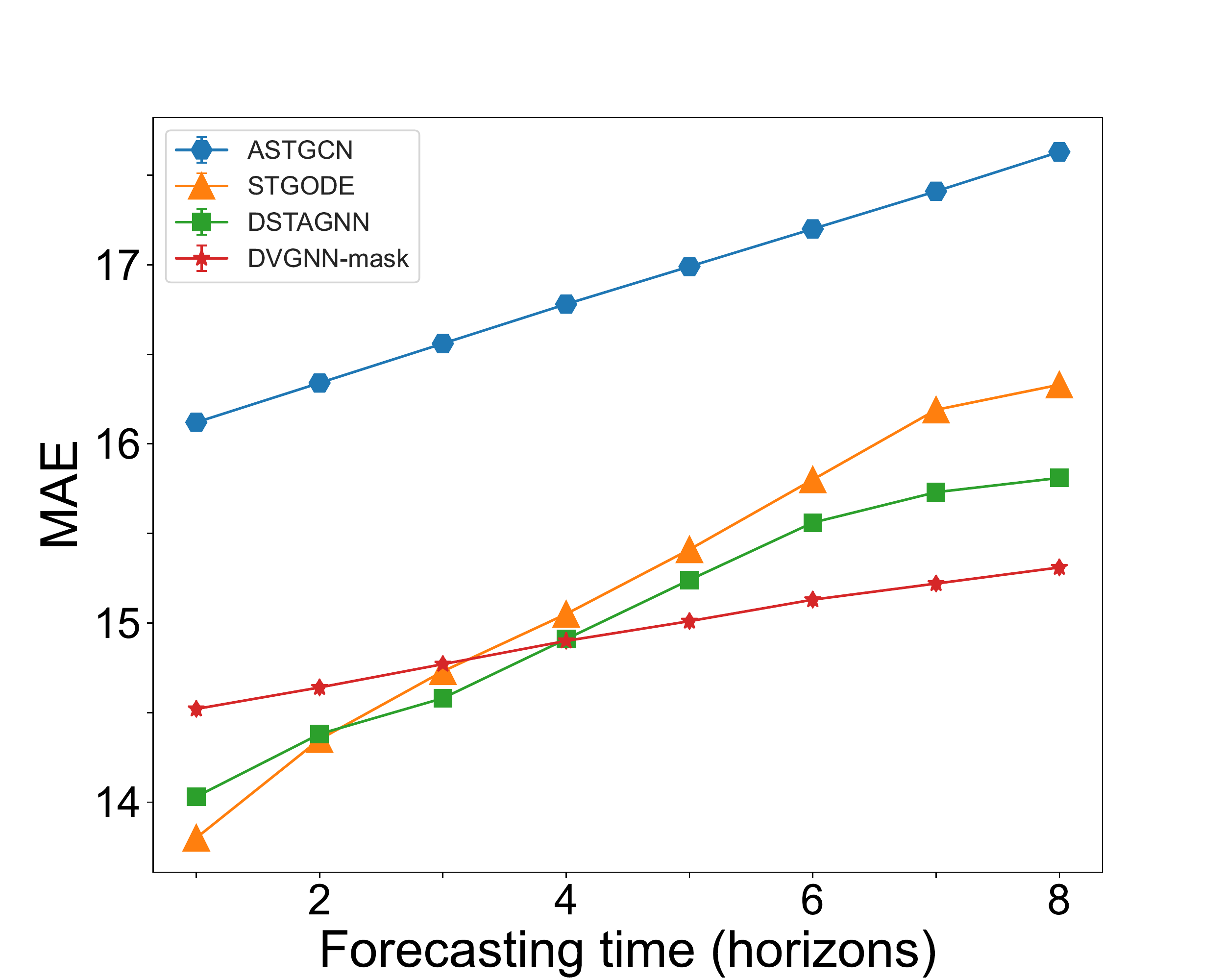} 	\label{baseline_curve_d} } 

\subfloat[RMSE of Tdrive.]{ 
\centering 
\includegraphics[width=0.50\columnwidth]{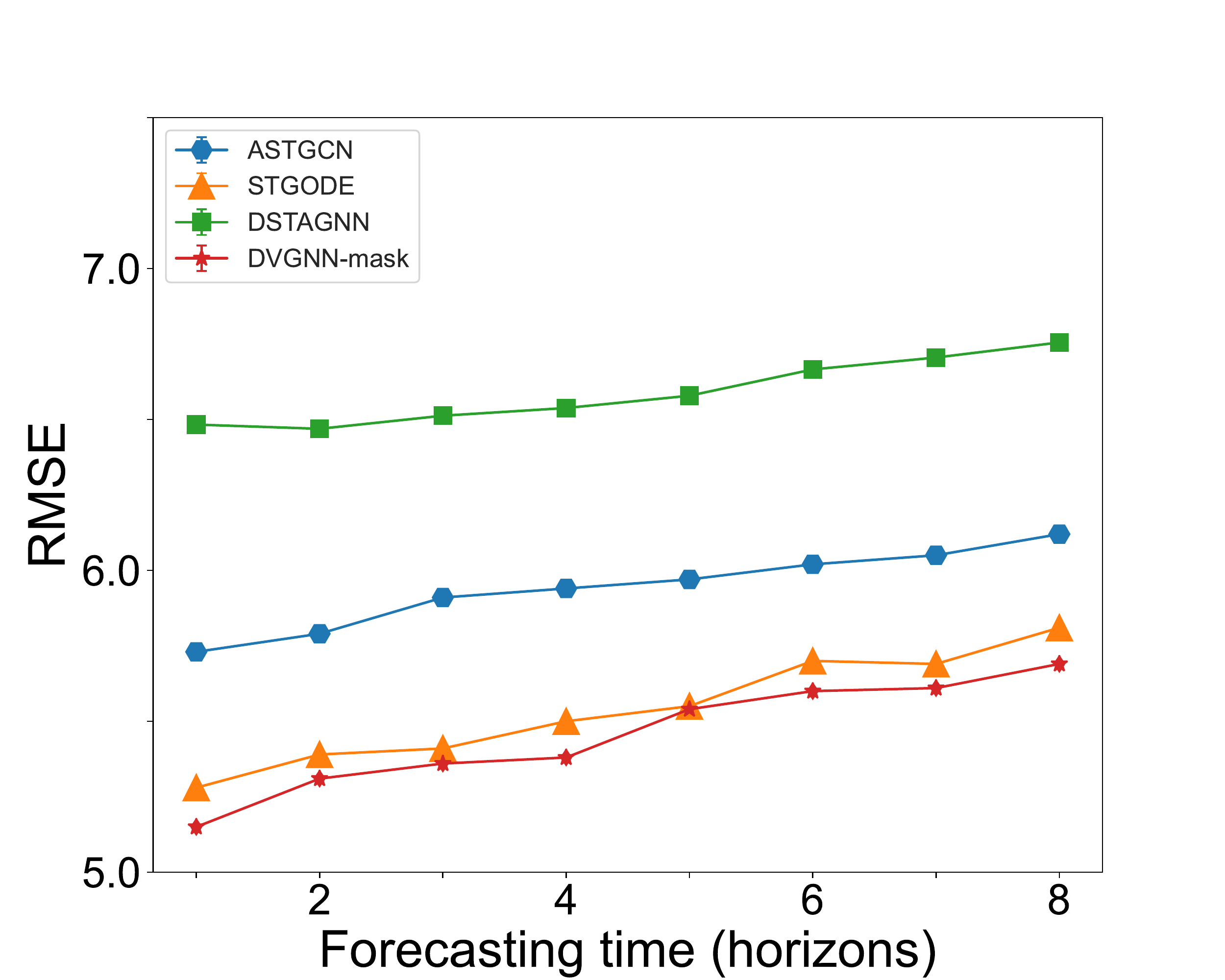} 	\label{baseline_curve_e} } 
\subfloat[MAE of Tdrive.]{ 
\centering 
\includegraphics[width=0.50\columnwidth]{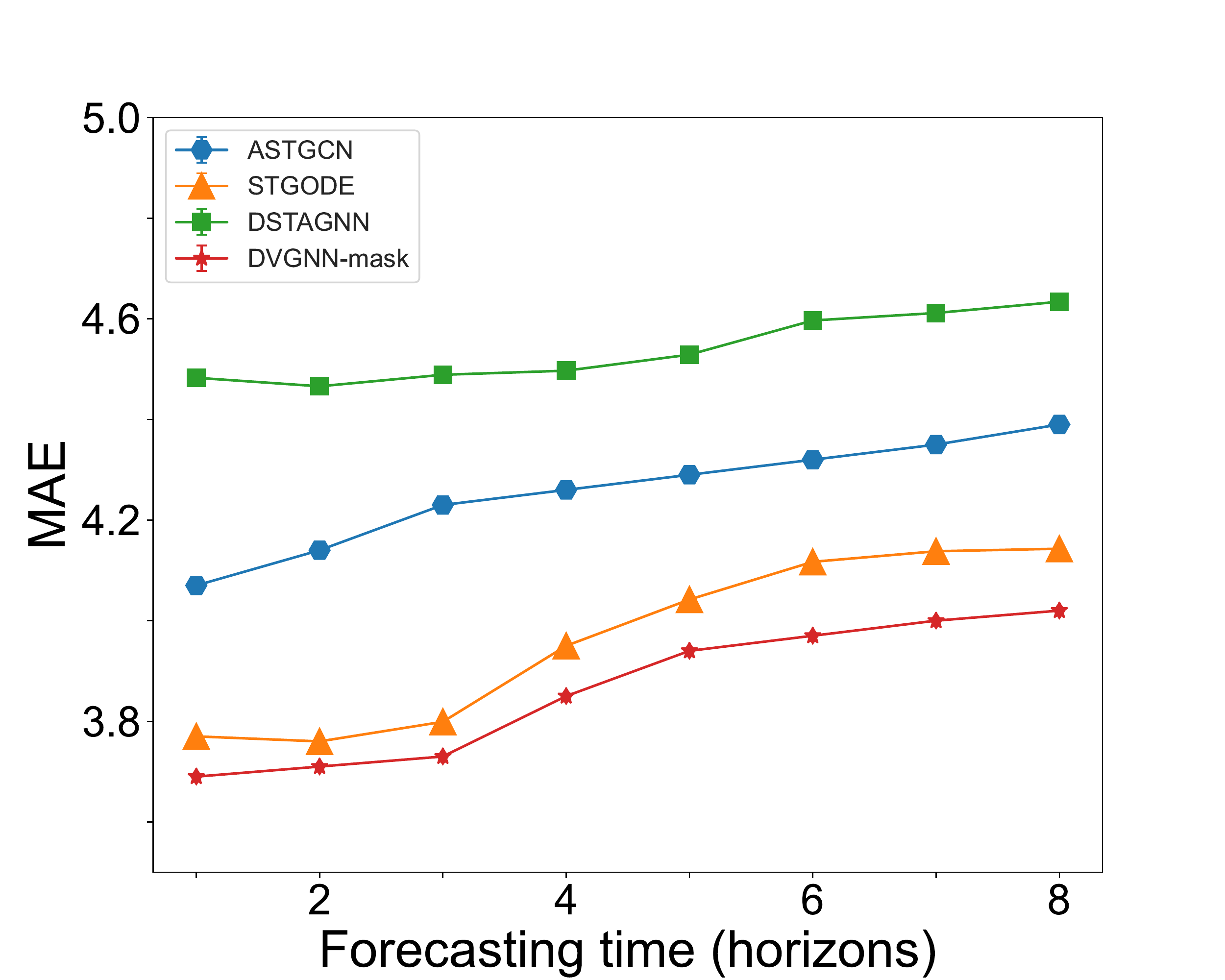} 	\label{baseline_curve_f} } 
\caption{The predictive results of dynamic graph methods based on three transportation datasets.} 
\label{baseline_curve}
\end{figure}

In order to better study the performance of DVGNN, dynamic graph methods are compared in 8 time horizons (1 horizon=5minutes, total 40 minutes) based on the transportation datasets in Fig. \ref{baseline_curve}. The prediction accuracy of all models declines with the increase in time. Clearly, long-term prediction is a difficult problem, but DVGNN can still obtain outstanding accuracy for all the datasets of different graph structures because it uses the dynamic generative model to represent the causal relationship between nodes, and it better represents the dynamic features of the spatio-temporal dataset and obtains a better interpretation of the dynamically generated graphs.

\begin{table}[] 
\caption{The prediction accuracy of DVGNN and baseline A models in 60 minutes on transportation datasets.}
\label{result1}
\begin{tabular}{ccccccc}
\hline 
\multirow{2}{*}{Models} & \multicolumn{2}{c}{Los-loop} & \multicolumn{2}{c}{PeMS08} & \multicolumn{2}{c}{T-Drive} \\    
& RMSE         & MAE         & RMSE         & MAE         & RMSE         & MAE          \\ \hline DCRNN\cite{li2017diffusion}         & 6.80        & 4.54       & 27.83        & 17.46       & 6.93         & 4.83       \\ T-GCN\cite{Zhao2020}                   & 7.26        & 4.60       & 29.95        & 20.98       & 6.88         & 4.99         \\ ASTGCN\cite{guo2019attention}             & 6.14        & 3.57       & 28.16        & 18.61       & 6.22         & 4.47         \\ STGODE\cite{fang2021spatial}                  & 6.71        & 4.17       & 25.91        & 16.81       & 5.96         & 4.34         \\ DSTAGNN\cite{lan2022dstagnn}                 & 6.79        & 4.54        & 24.77        & 15.67       & 6.91         & 4.75         \\ \textbf{DVGNN-mask}             & \textbf{5.49}        & \textbf{3.16}       & \textbf{24.51}        & \textbf{15.64}       & \textbf{5.92}         & \textbf{4.18}         \\ \hline
\end{tabular} 
\end{table}

\begin{table}[] 
\caption{The prediction results of DVGNN and causal discovery methods in the next horizon on the FMRI dataset of healthcare.}
\label{result2} 
\begin{tabular}{ccccccc} \hline \multirow{2}{*}{Models} & \multicolumn{2}{c}{FMRI-3} & \multicolumn{2}{c}{FMRI-4} & \multicolumn{2}{c}{FMRI-13} \\     
& RMSE         & MAE         & RMSE         & MAE         & RMSE         & MAE          \\ \hline TCDF\cite{nauta2019causal}                    & 2.87         & 2.20        & 3.39         & 2.57        & 2.61         & 2.10         \\ NGC\cite{tank2021neural}                     & 2.83         & 2.18        & 2.49         & 1.97        & 3.00         & 2.41         \\ CR-VAE\cite{li2023causal}                   & 2.67         & 2.08        & 2.11         & \textbf{1.66}        & 2.71         & 2.16         \\  \textbf{DVGNN}             & \textbf{2.35}         & \textbf{1.82}        & \textbf{2.09}         & \textbf{1.66}        & \textbf{2.43}         & \textbf{1.91}         \\ \hline 
\end{tabular} 
\end{table}

\begin{figure}[h!]    
\centering   
\subfloat[The transition probability of typical neighbour grids in T-drive dataset.]{     \centering   
\includegraphics[width=\columnwidth]{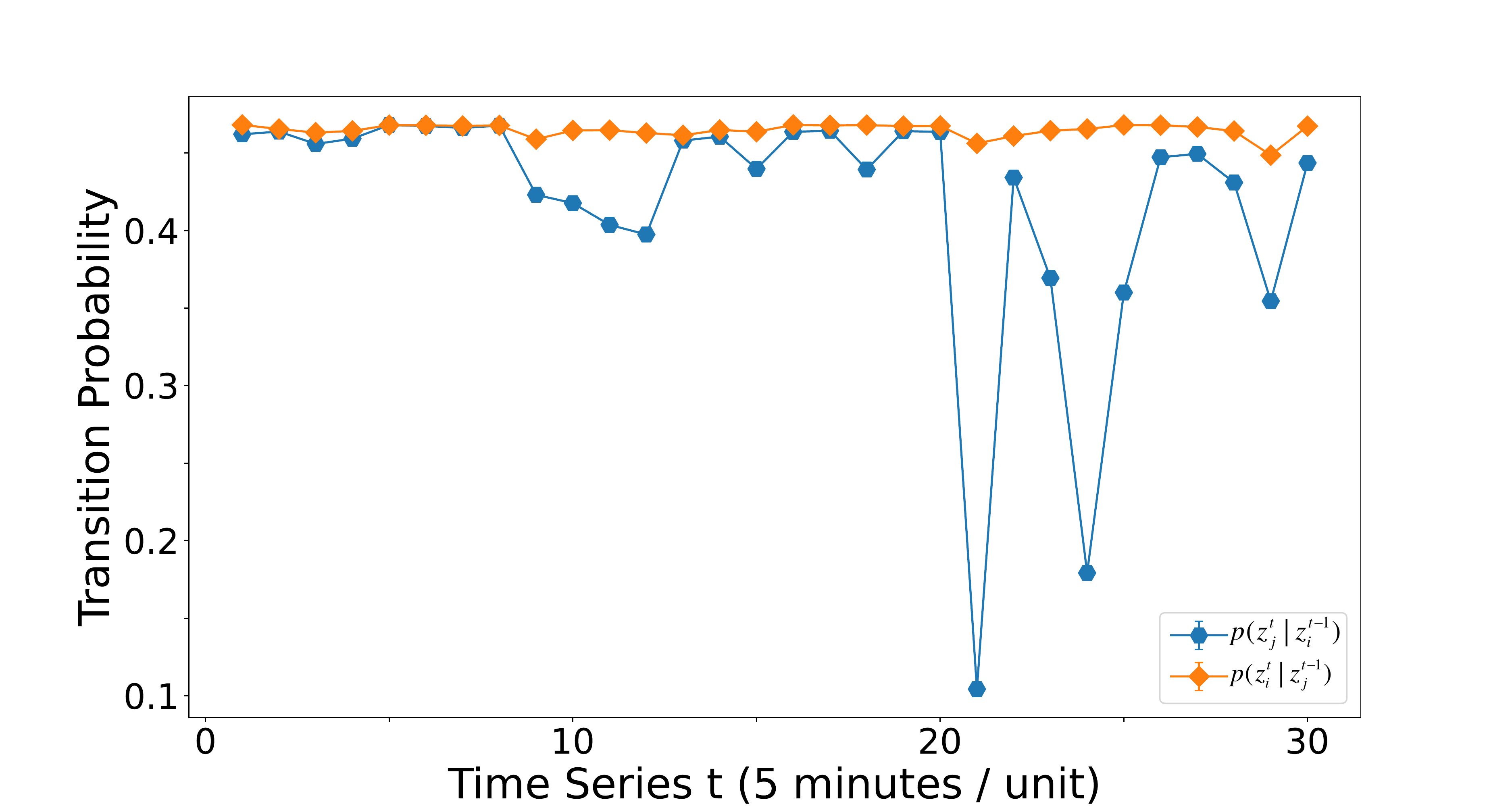}  
\label{transition probability_a}  
}  

\vfill 

\subfloat[The transition probability of typical neighbour roads in PeMS08 dataset.]{     \centering     
\includegraphics[width=\columnwidth]{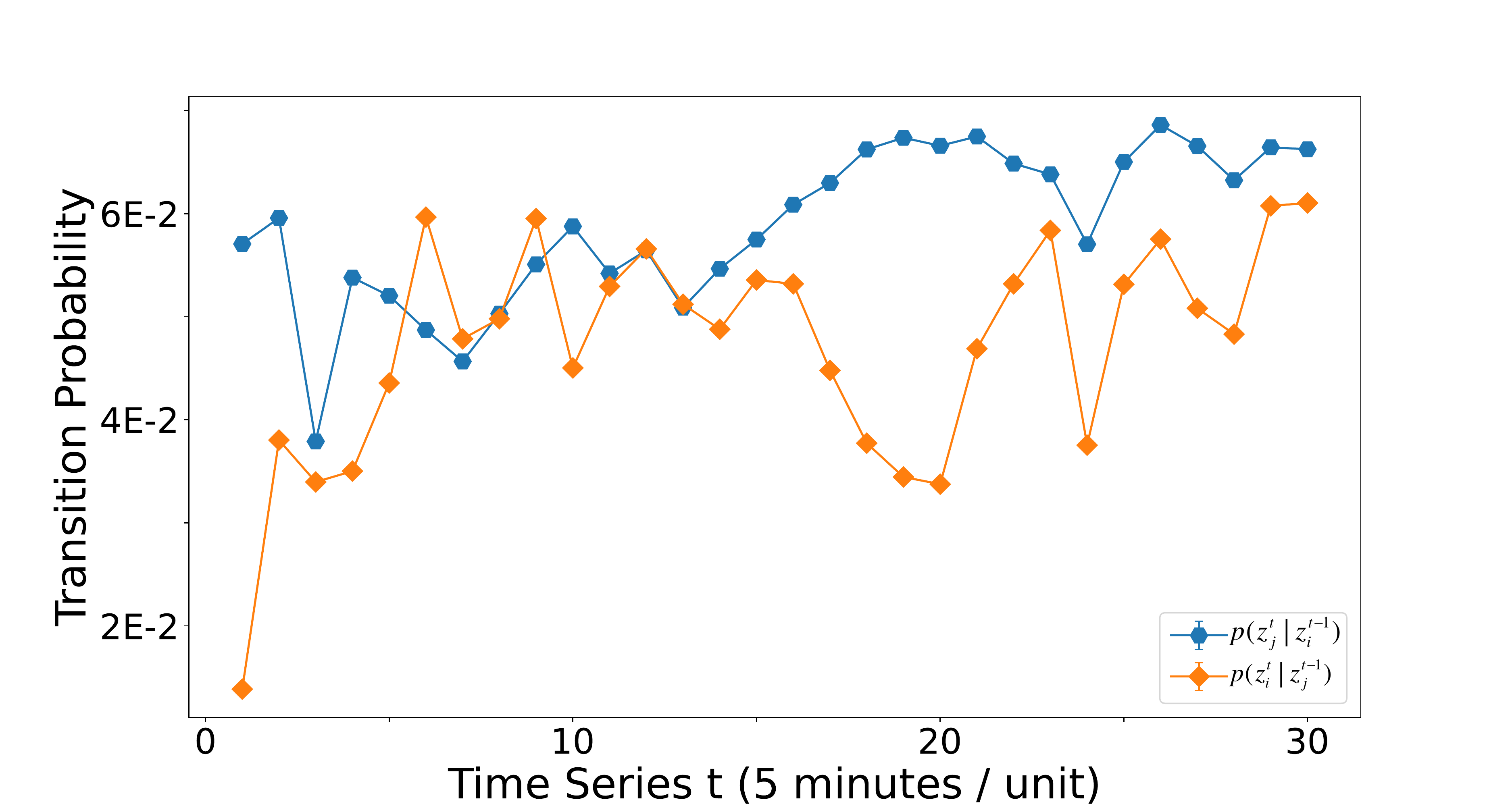}    
\label{transition probability_b}     
} 
\caption{The transition probability $p(z_j^{t}|z_i^{t-1})$ and $p(z_i^{t}|z_j^{t-1})$ of two correlated nodes in different time series of two datasets.} 
\label{transition probability_result}
\end{figure}

 \begin{figure*}[ht!]   
     \centering    
     \subfloat[The latent variable $Z$ distribution of typical neighbour grids in T-drive dataset.]{    
     \centering    
     \includegraphics[width=0.9 \linewidth]{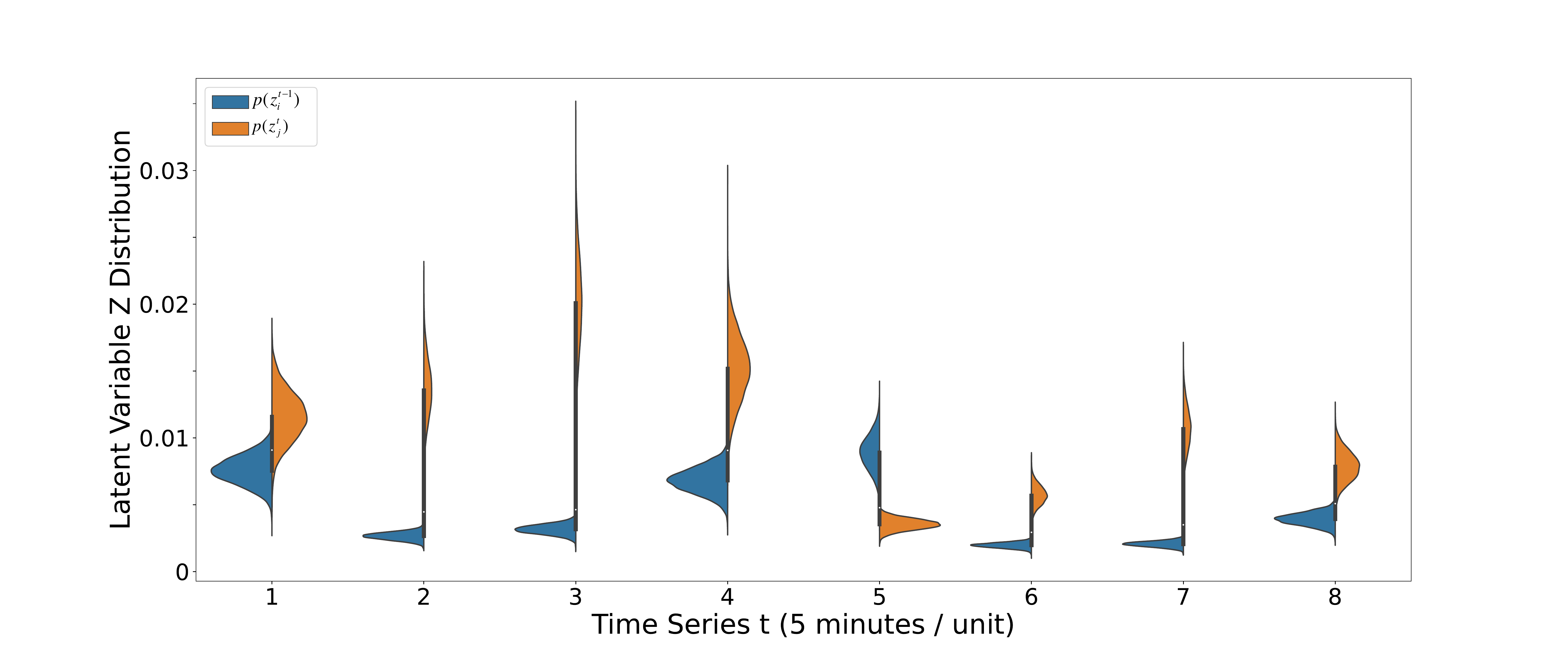}  
     \label{probability_a}  
     }

    \subfloat[Multivariate Gaussian distribution $p(z_j^{t},z_i^{t-1})$ of typical neighbour grids in T-drive dataset.]{    
    \centering    
    \includegraphics[width=0.7 \linewidth]{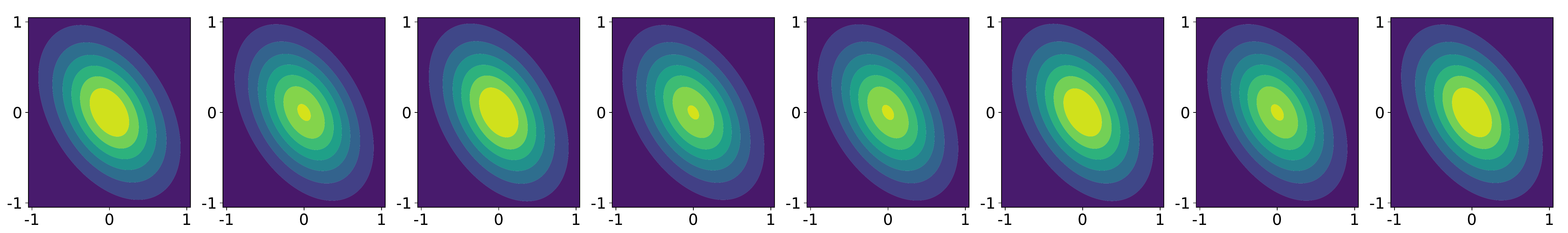}     \label{probability_b}   
    }   

    \subfloat[The latent variable $Z$ distribution of typical neighbour roads in PeMS08 dataset.]{    
    \centering    
    \includegraphics[width=0.9 \linewidth]{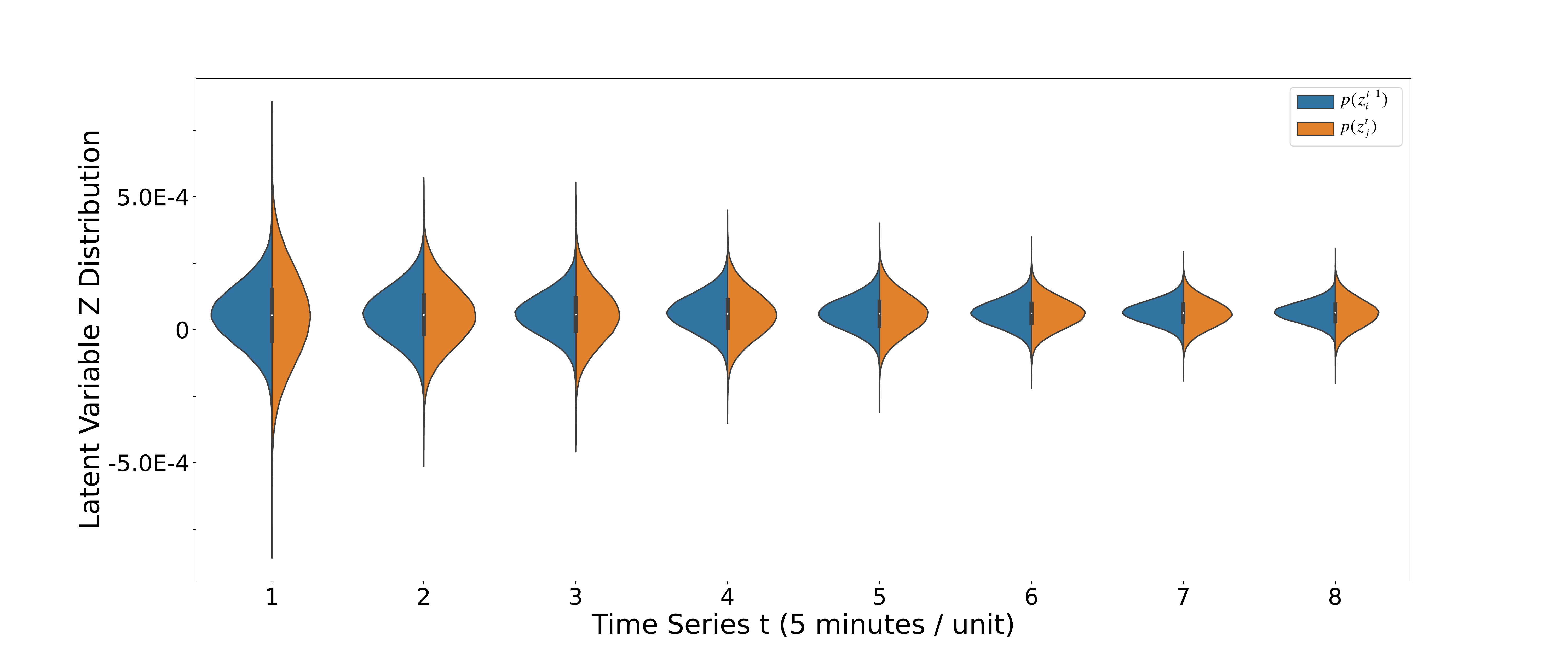}    
    \label{probability_c}    
    }

    \subfloat[Multivariate Gaussian distribution $p(z_j^{t},z_i^{t-1})$ of typical neighbour nodes in PeMS08 dataset.]{    
    \centering  
    \includegraphics[width=0.7 \linewidth]{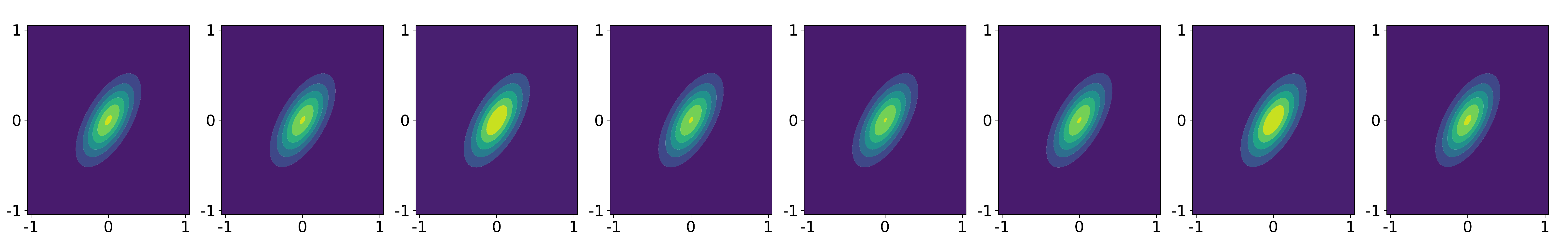}     \label{probability_d} 
    }
    \caption{The latent variable $Z$ distribution and Multivariate Gaussian distribution $p(z_j^{t},z_i^{t-1})$ of two correlated nodes in several time series of two datasets.} \label{latent distribution}  
\end{figure*}

\begin{table}[] 
\caption{The dynamic graph accuracy of DVGNN over the total test time series on test edges for all datasets.}
\label{F1score} 
\begin{tabular}{ccccc}
\hline 
Dataset                  & Model      & AUC          & Precision    & F1-Score     \\ \hline \multirow{3}{*}{T-Drive} & VGAE\cite{kipf2016variational}       & 53.95±8\%  & 53.91±13\% & 52.96±6\%  \\                          & DVGNN      & 58.35±2\%  & 61.32±3\%  & 57.51±2\%  \\                          & \textbf{DVGNN}-mask & \textbf{79.07}±3\%  & \textbf{75.32}±3\%  & \textbf{69.44}±3\%  \\ \hline \multirow{3}{*}{Los-loop}  & VGAE\cite{kipf2016variational}       & 48.37±7\% & 49.46±6\% & 50.50±5\% \\                          & DVGNN      & 58.89±3\%  & 59.76±3\%  & 57.36±2\%  \\                          & \textbf{DVGNN-mask} & \textbf{68.28}±3\%  & \textbf{67.84}±3\%  & \textbf{60.92}±1\%  \\ \hline \multirow{3}{*}{PeMS08}  & VGAE\cite{kipf2016variational}       & 52.90±17\% & 53.75±15\% & 52.96±11\% \\                          & DVGNN      & 49.14±4\%  & 46.84±4\%  & 47.85±3\%  \\                          & \textbf{DVGNN-mask} & \textbf{78.27}±2\%  & \textbf{68.97}±4\%  & \textbf{66.23}±3\%  \\ \hline \multirow{2}{*}{FMRI-3}  & TCDF \cite{nauta2019causal}      & 66.31±2\%  & 55.55±2\%  & 55.07±2\%  \\                          & NGC\cite{tank2021neural}        & 55.55±2\%    & 43.73±2\%    & 47.73±2\%    \\                                                    & CR-VAE\cite{li2023causal}     & 57.00±2\%    & 57.30±2\%    & 53.40±2\%    \\                          & \textbf{DVGNN}      & \textbf{68.63}±3\%  & \textbf{70.12}±2\%  & \textbf{65.42}±2\%  \\ \hline \multirow{2}{*}{FMRI-4}  & TCDF\cite{nauta2019causal}         & 56.28±2\%  & 19.08±2\%  & 20.66±2\%  \\                          & NGC\cite{tank2021neural}        & 55.00±2\%    & 36.51±2\%    & 43.19±2\%    \\                                                    & CR-VAE\cite{li2023causal}     & 55.70±2\%    & 52.20±2\%    & 51.10±2\%    \\                          & \textbf{DVGNN}      & \textbf{63.55}±2\%  & \textbf{60.12}±1\%  & \textbf{67.00}±1\%  \\ \hline \multirow{2}{*}{FMRI-13} & TCDF\cite{nauta2019causal}         & 54.38±3\%  & 60.00±4\%  & 52.17±4\%  \\                          & NGC\cite{tank2021neural}        & 55.00±3\%    & 66.22±4\%    & 57.80±4\%    \\                                                    & CR-VAE\cite{li2023causal}     & 53.30±3\%    & \textbf{67.30}±4\%    & 60.20±4\%    \\                          & \textbf{DVGNN}      & \textbf{61.97}±4\%  & 66.89±4\%  & \textbf{61.81}±3\%  \\ \hline 
\end{tabular} 
\end{table}

\subsection{Uncertainty Analysis and Interpretation}
Most dynamic graph methods belong to supervised and discriminative models, which learn deterministic node features and ignore the uncertainty of node attributes. Therefore, their graph representation capacity is limited. Moreover, they rarely analyze the uncertainty of spatio-temporal data and the causal relationship between the neighbour nodes of the generated graphs. As a result, they lack steady robustness and a strong interpretation of the generative graphs. DVGNN adopts the generative model to generate the dynamic graphs based on the causal relationship of the diffusion process at the adjacent time interval, which takes the uncertainty and noise into consideration. 

The dynamically generated graphs are evaluated over the total test time series on test edges on total datasets. The result is shown in Table \ref{F1score}. DVGNN can obtain the best accuracy in general. For the large graph with large nodes, DVGNN-mask adopts the pre-defined adjacent matrix as a mask obtains higher accuracy and stronger robustness than DVGNN, since the mask acts like regularization for preventing overfitting. In addition, if all time series data is reshaped and fed to VGAE for transportation datasets, area under the ROC curve (AUC), and F1-Score display large variability since traditional VGAE can not exploit the temporal features well. For FMRI datasets without a pre-defined graph structure, baseline B models display different performances in the causal discovery. Taking the FMRI-4, for example, the TCDF method obtains the lowest precision and F1-Score, as attention-based convolutional neural network methods are not skilled in dealing with short time series combined with a complex causal graph structure of large nodes. In addition, DVGNN achieves the outstanding performance of different causal graphs with different nodes since it takes the interaction of nodes as a stochastic diffusion process, which would be a more robust way to discover the causal graphs.

Taking the typical edge of two roads or grids in the different datasets, for example, the transition probability at interval times is shown in Fig. \ref{transition probability_result}. The transition probability between the two neighbour nodes changes over time. For T-drive dataset, the probabilities of traffic flow transit between two neighbour grids is high, and the traffic flow from grid $v_i$ to $v_j$ is steady, while the traffic flow transition from $v_j$ to $v_i$ drops dramatically at times 21 and 24. For PeMS08 dataset, the transition probability between road $v_j$ to road $v_i$ displays a bit of lag behind at first, but synchronization happens at time 23. DVGNN can discover dynamic relationship clearly.

\begin{figure*}[ht!]   
 \centering    
 \subfloat[Transition probability matrix at time 1]{      
 \centering   
 \includegraphics[width=0.33 \linewidth]{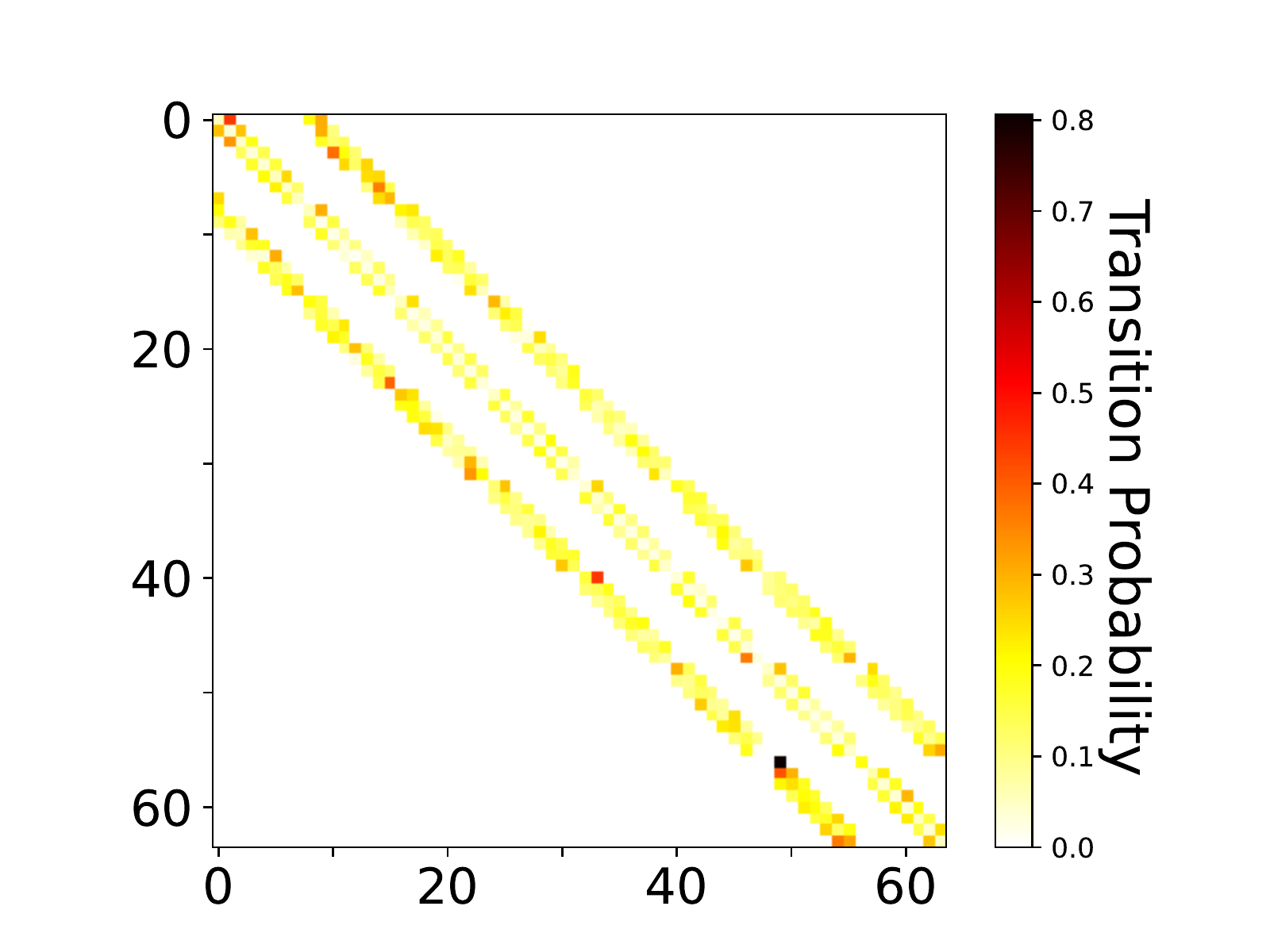}       \label{visualization_a} }   
 \subfloat[Transition probability matrix at time 2]{ 
 \centering      
 \includegraphics[width=0.33   \linewidth]{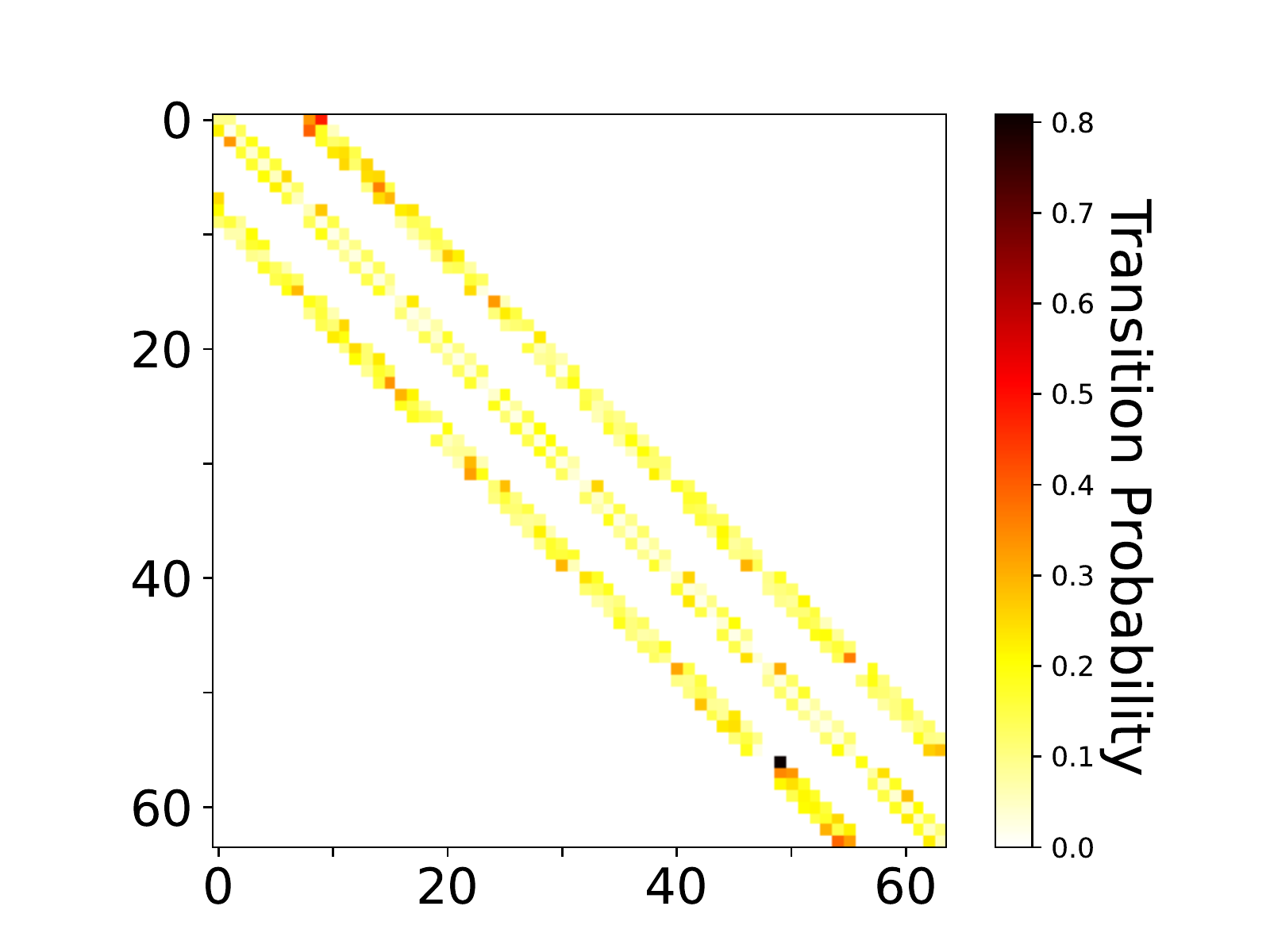}       \label{visualization_b}     } 
 \subfloat[Transition probability matrix at time 3]{  
 \centering      
 \includegraphics[width=0.33   \linewidth]{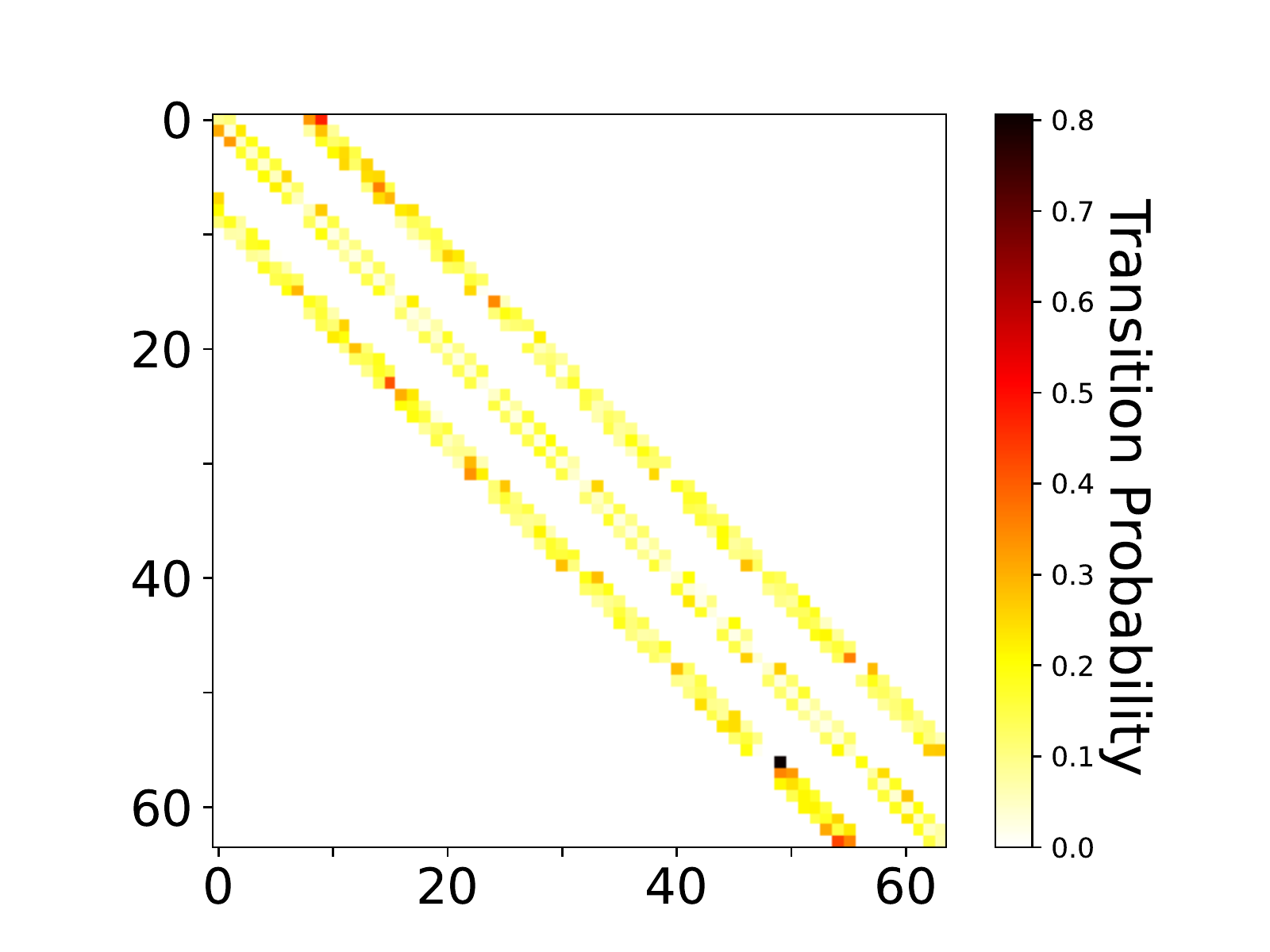}     \label{visualization_c}     }  
 \caption{Visualization of generated transition graphs of T-drive dataset in 3 adjacent time intervals.} 
 \label{mattrix visualization}   
 \end{figure*}

\begin{figure*}[ht!]    
 \centering      
 \subfloat[Transition probability matrix at the beginning of rush hour]{        \centering      
 \includegraphics[width=0.33 \linewidth]{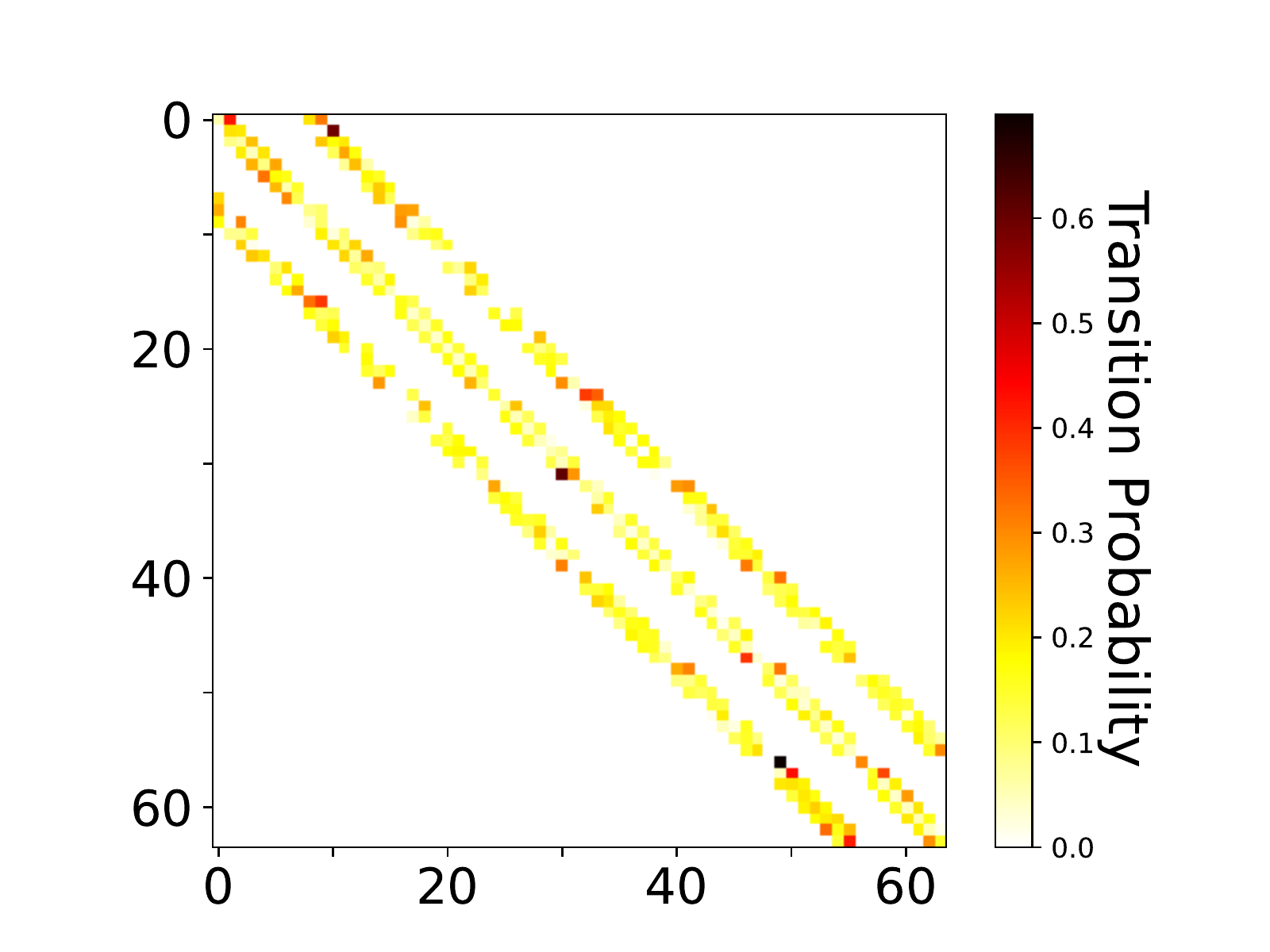}       \label{visualization2_a} }  
 \subfloat[Transition probability matrix at the end of rush hour]{
 \centering       
 \includegraphics[width=0.33   \linewidth]{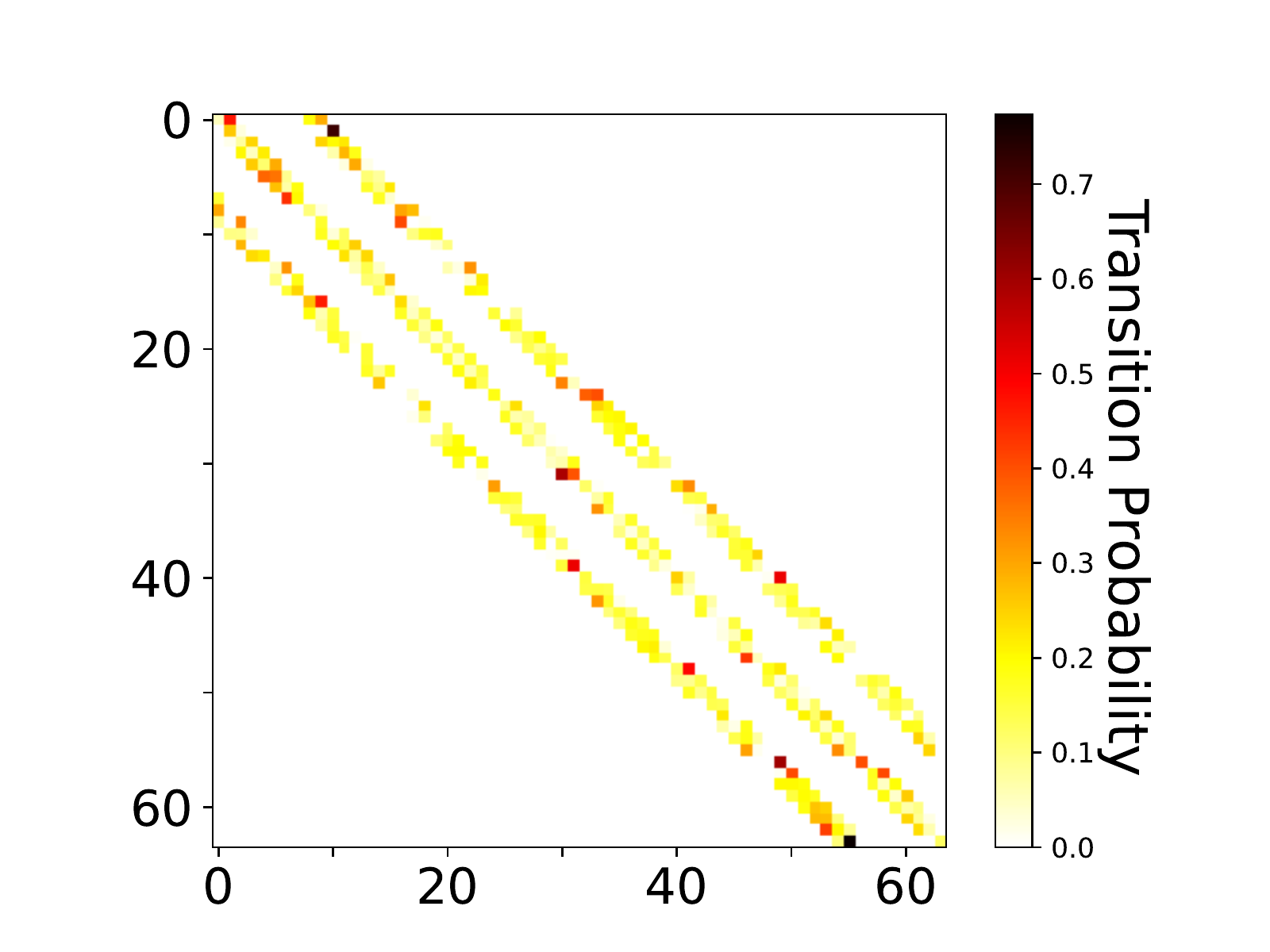}       \label{visualization2_b} 
 } 
 \subfloat[Transition probability matrix in the leisure hour]{  
 \centering     
 \includegraphics[width=0.33   \linewidth]{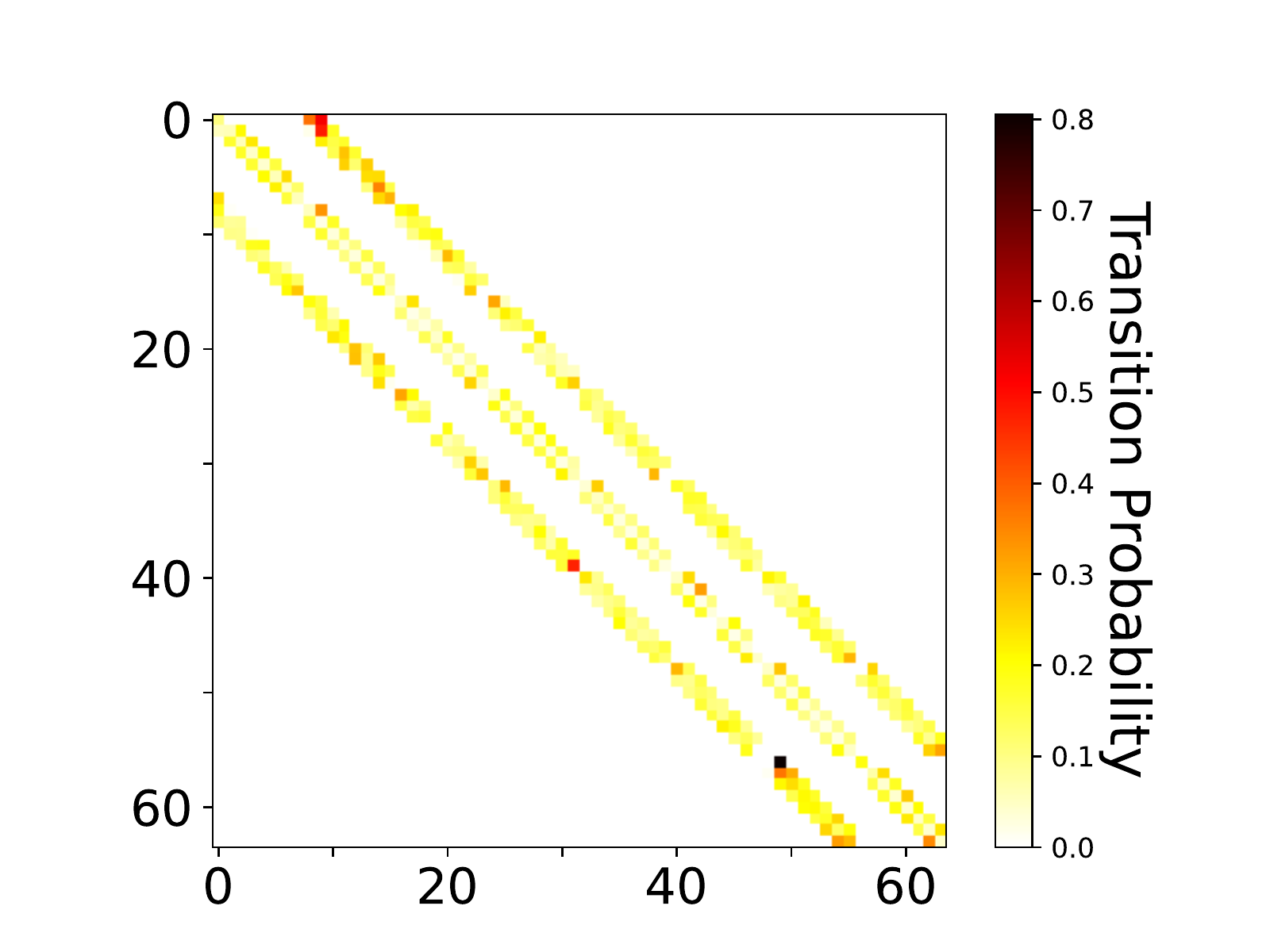}     \label{visualization2_c} 
 } 
 \caption{Visualization of generated transition graphs of T-drive dataset in 3 different time intervals.}  
 \label{mattrix visualization2}    
 \end{figure*}

To explore the interpretation of DVGNN, we plot the typical posterior probability distribution of latent variable $\{z_i^{t-1},z_j^{t}\}$ and corresponding multivariate Gaussian distribution of $p(z_j^{t},z_i^{t-1})$ at several time intervals as shown in Fig. \ref{latent distribution}, where $z_i^{t-1}$ and $z_j^{t}$ are the latent variables of connected neighbour nodes in the dynamic graphs. From Fig. \ref{probability_a}, the posterior probability distribution of latent variables $z_i^{t-1}$ and $z_j^{t}$ in T-drive dataset presents the opposite change trend. One changes sharp and steep while the other changes flattened, and vice versa. From Fig. \ref{probability_c}, the posterior probability distribution of latent variables $z_i^{t-1}$ and $z_j^{t}$ in PeMS08 dataset presents the same change trend. In order to further study their relationship, corresponding multivariate Gaussian distributions are plotted as shown in \ref{probability_b} and \ref{probability_d}. Latent variables $z_i^{t-1}$ and $z_j^{t}$ in Fig. \ref{probability_b} show a strong negative correlation, in which the probability density function rotates anti-clockwise. On the contrary, the latent variables $z_i^{t-1}$ and $z_j^{t}$ in Fig. \ref{probability_d} show a positive correlation, in which the probability density function rotates clockwise. Different from other dynamic graph methods, DVGNN can discover the causal relationship of latent variables by the physics-informed diffusion process. To display the dynamics of the generated graphs intuitively, Fig. \ref{mattrix visualization} plot generated transition density probability graphs of T-drive dataset in adjacent 3-time intervals. The three lines indicate the adjacent matrix of the static graph, while the colour is the value of transition probability $p(z_j^{t}|z_i^{t-1})$. The figures show that not all the connected grids of the static graph are highly correlated with each other all the time, their relationships are dynamic, and the whole graphs at adjacent times show some similarity since the fact that all the grids change together in a very short time is not usual in the common case. As for long time intervals, Fig. \ref{mattrix visualization2} indicates different states of three long time intervals. Fig. \ref{visualization2_a} and Fig. \ref{visualization2_b} are the beginning and end of the rush hour, the points of high transition probability will emerge more since more red points can be seen in the graphs, while fewer high-value colour points in leisure hour as shown in Fig. \ref{visualization2_c}. As a result, DVGNN can effectively represent the dynamics of spatio-temporal features.

\subsection{Ablation Study}
\begin{figure}[h!]     
\centering    
\subfloat[RMSE of Los-loop.]{ 	
\includegraphics[width=0.50\columnwidth]{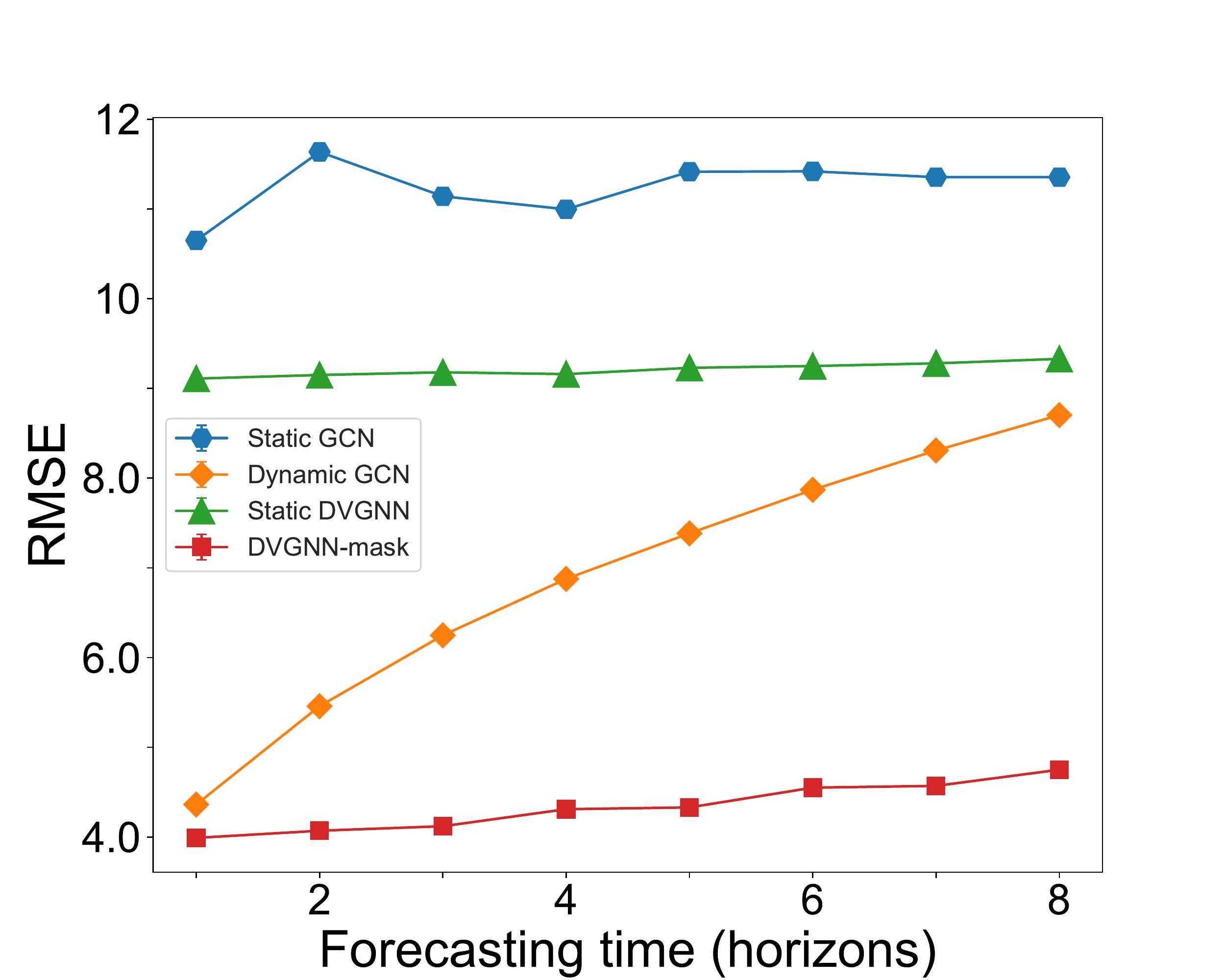} 
\label{ff}
} 
\subfloat[MAE of Los-loop.]{ 	
\includegraphics[width=0.50\columnwidth]{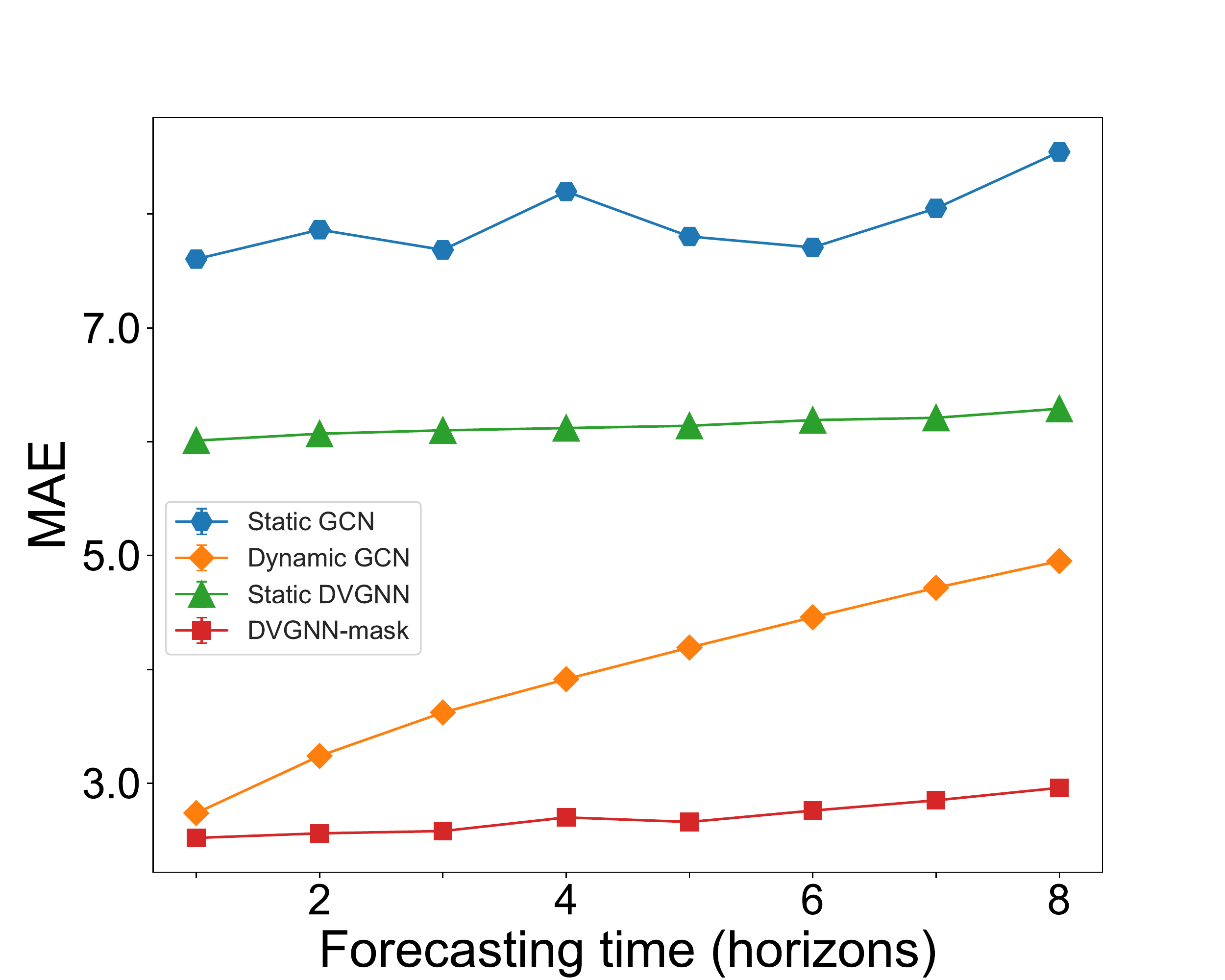}
\label{aa}
}   

\subfloat[RMSE of PeMS08.]{ 
\includegraphics[width=0.50 \columnwidth]{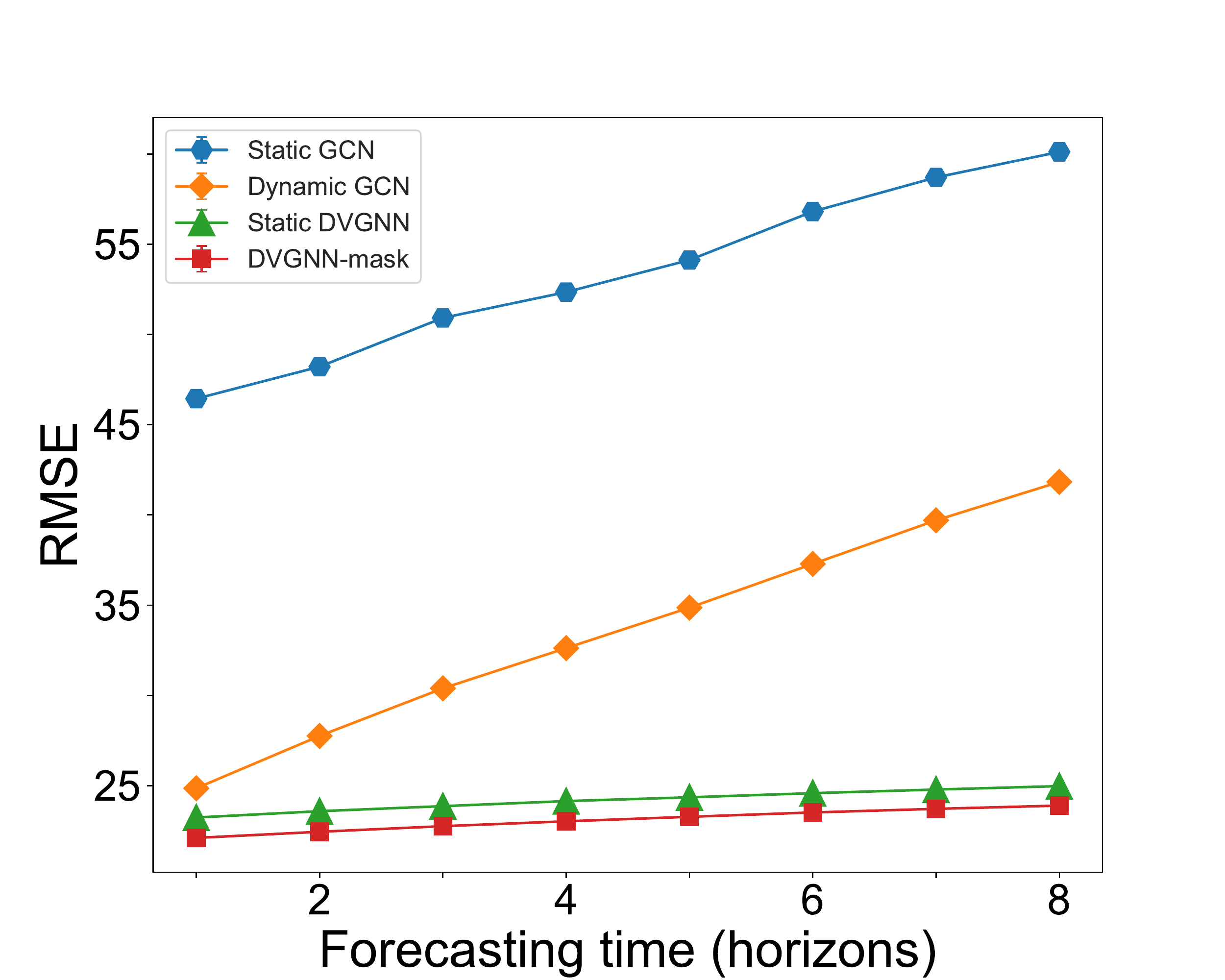} 	
\label{bb} 
} 
\subfloat[MAE of PeMS08.]{
\includegraphics[width=0.50\columnwidth]{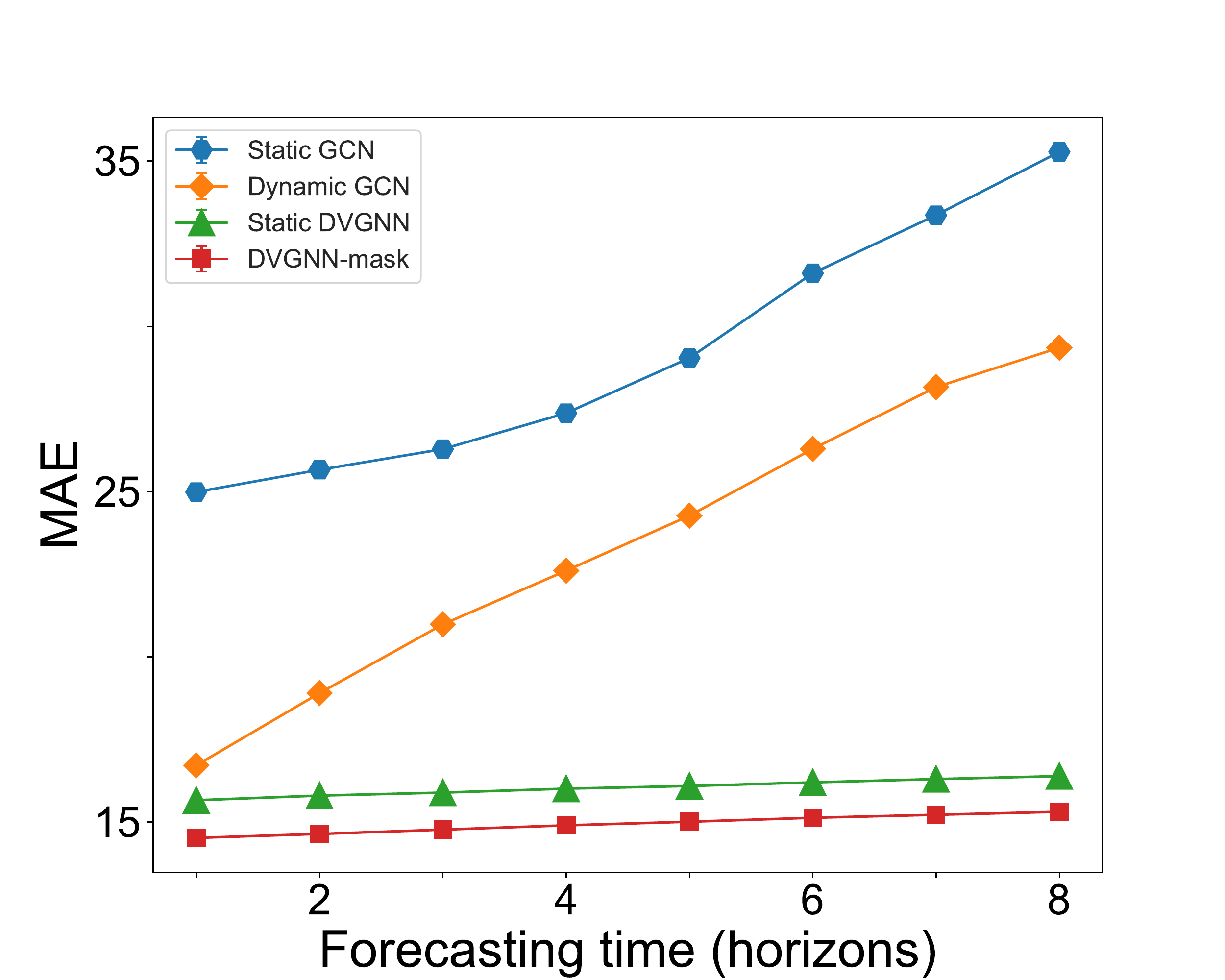} 
\label{cc}
}

\subfloat[RMSE of T-drive.]{ 
\includegraphics[width=0.50\columnwidth]{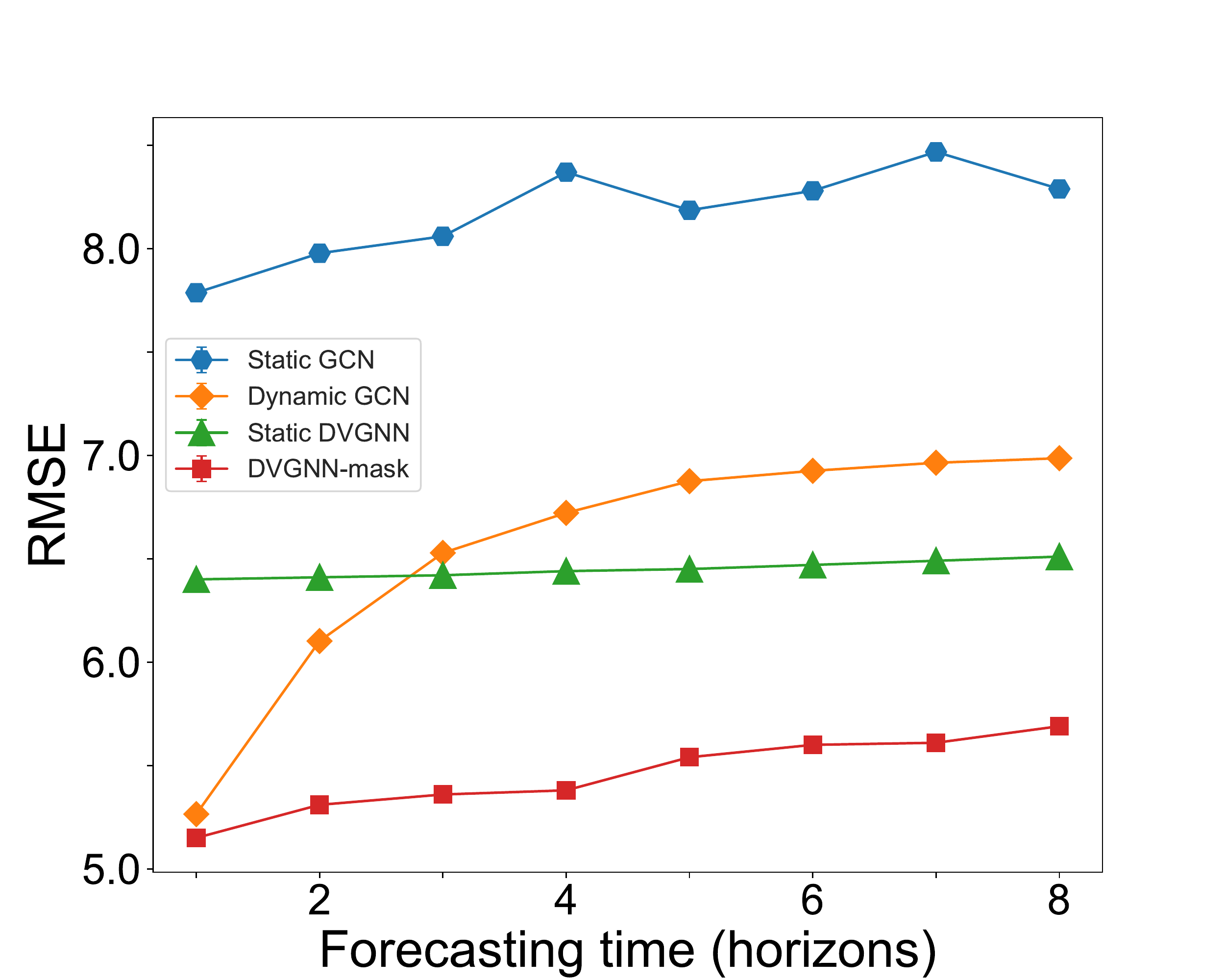} 
\label{dd} } 
\subfloat[MAE of T-drive.]{ 
\includegraphics[width=0.50\columnwidth]{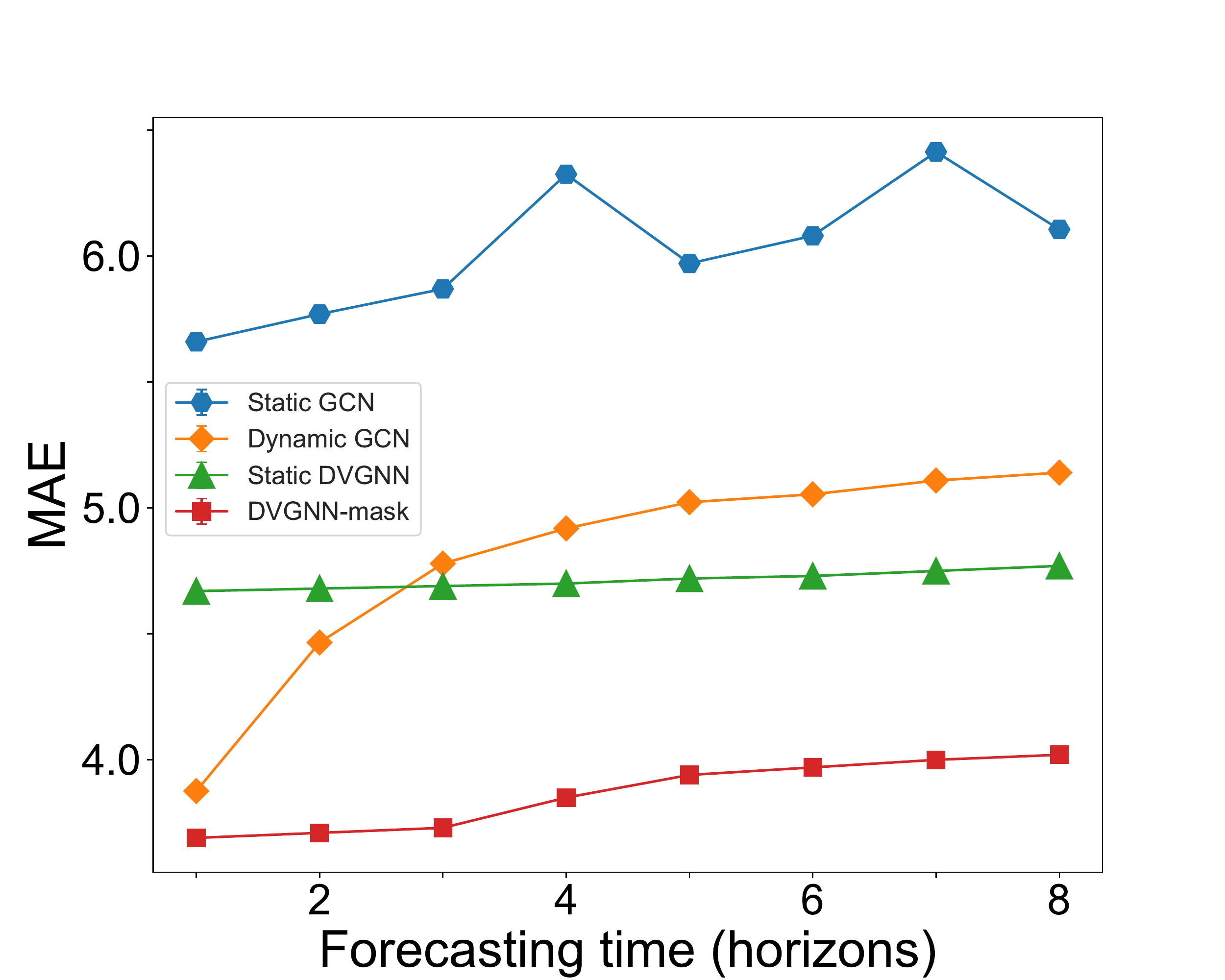} 
\label{ee}
}
\caption{Ablation experiments of module effectiveness} 
\label{ablation study}
\end{figure}

To verify the effectiveness of each component in DVGNN, we test variants of our model on three datasets. The variants are (1) Static DVGNN: replace the dynamic graph with the static graph but retain the other modules. (2) Static GCN: completely remove the temporal attention module and dynamic graph generator, but retain GCNs with the static graph. (3) Dynamic GCN: completely remove the temporal attention module, but retain the dynamic graph generator and apply GCNs with the dynamic graphs. It can be seen from Fig. \ref{ablation study} that the dynamic graph module has improved the prediction accuracy of both the Static GCN and Static DVGNN methods, which demonstrates the validity of the dynamic graph module. With the increase in the prediction time, the prediction accuracy is also decreasing, but the DVGNN model has little fluctuation and obtains the best accuracy. At the same time, it is noted that the accuracy of spatio-temporal models is higher than that of only using spatial methods. It is also proved that the key to obtaining high accuracy is effectively exploiting the spatio-temporal features of the datasets.

\begin{table}[]
\caption{The robustness experiments on different datasets.} 
\label{robustness2}
\begin{tabular}{ccccccc} \hline \multirow{2}{*}{Method} & \multicolumn{2}{c}{Poisson 1} & \multicolumn{2}{c}{Poisson 2} & \multicolumn{2}{c}{Poisson 3} \\                         & RMSE          & MAE           & RMSE          & MAE           & RMSE           & MAE           \\ \hline \multicolumn{7}{c}{T-Drive-60minutes}                                                                                              \\ \hline Static DVGNN            & 6.28          & 4.60          & 6.70          & 5.09          & 7.78           & 6.13          \\ DVGNN-mask              & 5.97          & 4.21          & 6.23          & 4.66          & 6.39           & 4.87          \\ \hline \multicolumn{7}{c}{Los-loop-60minutes}                                                                                               \\ \hline Static DVGNN            & 10.88         & 7.53         & 11.58         & 8.02         & 12.59          & 9.29         \\ DVGNN-mask              & 5.57         & 3.66         & 6.09         & 3.74         & 6.59          & 4.59         \\ \hline \multicolumn{7}{c}{FMRI-3 -next horizon}                                                                                               \\ \hline TCDF\cite{nauta2019causal}                    & 2.97          & 2.32          & 3.45          & 2.72          & 4.62           & 3.43          \\ DVGNN                   & 2.39          & 1.86          & 2.68          & 2.13          & 2.94           & 2.38          \\ \hline 
\end{tabular} 
\end{table}

\subsection{Robustness}
The spatio-temporal data in real application scenarios often contain various noises. In order to explore the robustness of our dynamic graph module, two datasets of transportation with different graph structures were chosen and compared with static DVGNN, while FMRI-3 of medium node number with non-pre-define graph structure was chosen and compared with the TCDF model, since TCDF achieve the best F1-score except our model. Moreover, the noise of Poisson distribution $P(\lambda)$ is the typical noise which exists in time series collecting sensors. Poisson distribution belongs to a discrete distribution, which defines the number of an event occurring within a given interval of time or space. The Poisson noise is set as $\lambda \in \{1,3,6\}$ to the training  datasets but not added to the test datasets. In addition, long-term prediction (12 horizons) is tested in transportation datasets, while short-term prediction (1 horizon) is tested in the FMRI dataset. The predicted results are shown in Table \ref{robustness2}. The accuracy of all datasets decreases as $\lambda$ increases. Thus, how to de-noise is still an important topic in spatio-temporal data mining and should be taken into consideration in the model construction, but DVGNN performance degrades the least. This is due to the introduction of the noise influence in the decoder stage by SDE, while other discriminative models, which learn deterministic node features, are sensitive to noise. Therefore, DVGNN achieves steady performance and exhibits better robustness. 

\section{Conclusion}
In this paper, a novel dynamic unsupervised generative graph construction model DVGNN is proposed for spatio-temporal forecasting. The inference model is adopted by two-layer GCNs in the encoder stage, and the causal graphs of the neighbour nodes are discovered by the diffusion model in the decoder stage, which are formulated by physics-informed noise-driven SDE. The objective function is deduced and approximated to simplify the calculation. Moreover, the model considers the uncertainty of causal graphs and the impact of noise. As a result, highly accurate and more robust results can be achieved by this dynamic generative graph model. The proposed model is evaluated and compared with other state-of-the-art methods based on four real time series datasets with different graph structures. DVGNN can not only outperform in terms of prediction accuracy but also discover the causal graphs adaptively, which provides better interpretability of the generated graphs in a robust way. 

Although DVGNN can achieve exciting results, it is still challenging to predict complex spatio-temporal data in the long term. In the future, we will explore better-suited prior distributions and more flexible generative models for more accurate prediction in complex scenarios.


\bibliographystyle{IEEEtran}
\bibliography{DVGNN}

\vfill

\newpage
\appendix
\section{Appendix}
\subsection{Transportation datasets} 
\begin{figure*}
\centering  
\subfloat[T-drive grid segment distribution.]{ 	
\centering	
\includegraphics[width=0.34\textwidth,height=4.5cm]{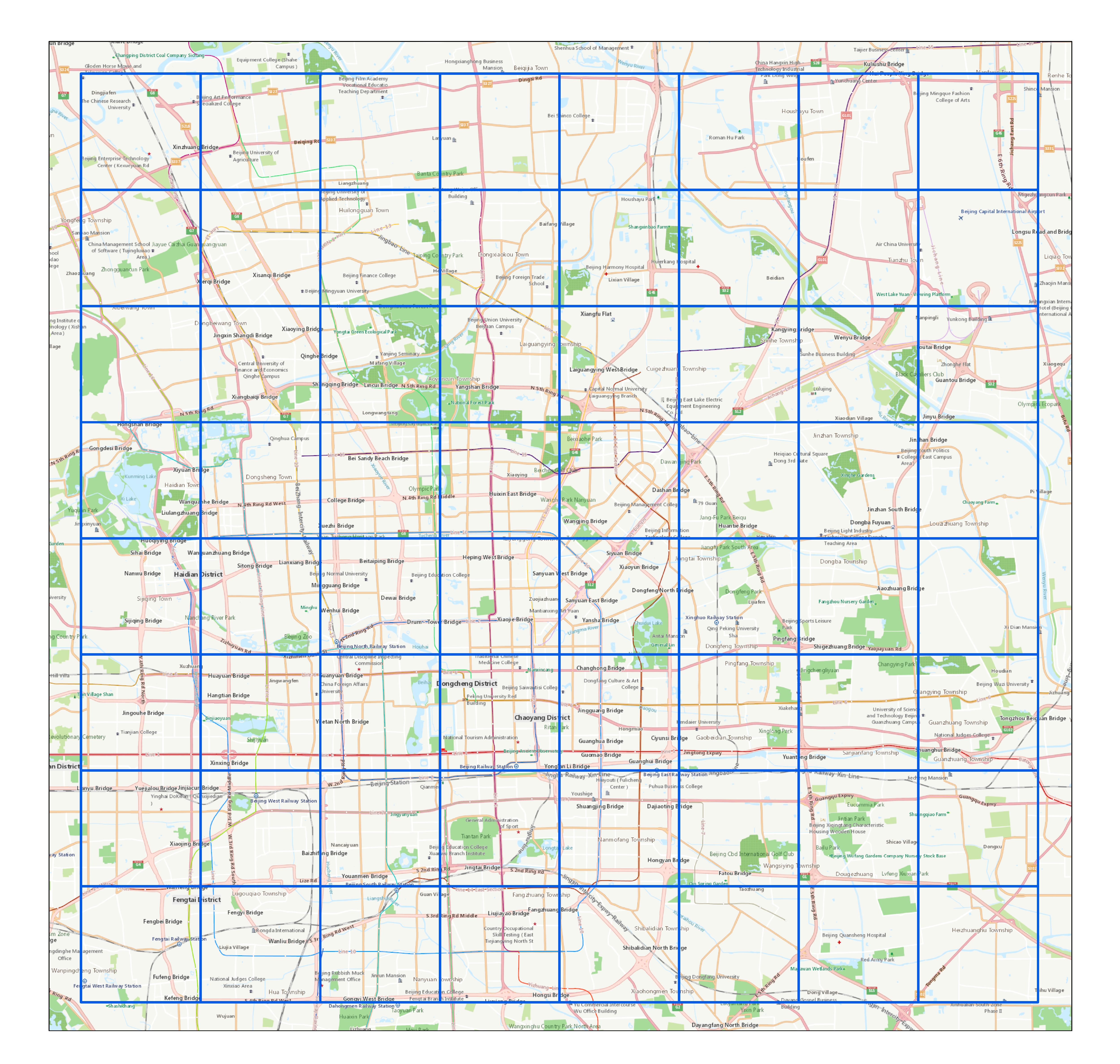} 	
\label{fig7_a} } 
\subfloat[PeMS08 road segment distribution.]{ 	
\centering 	
\includegraphics[width=0.34\textwidth]{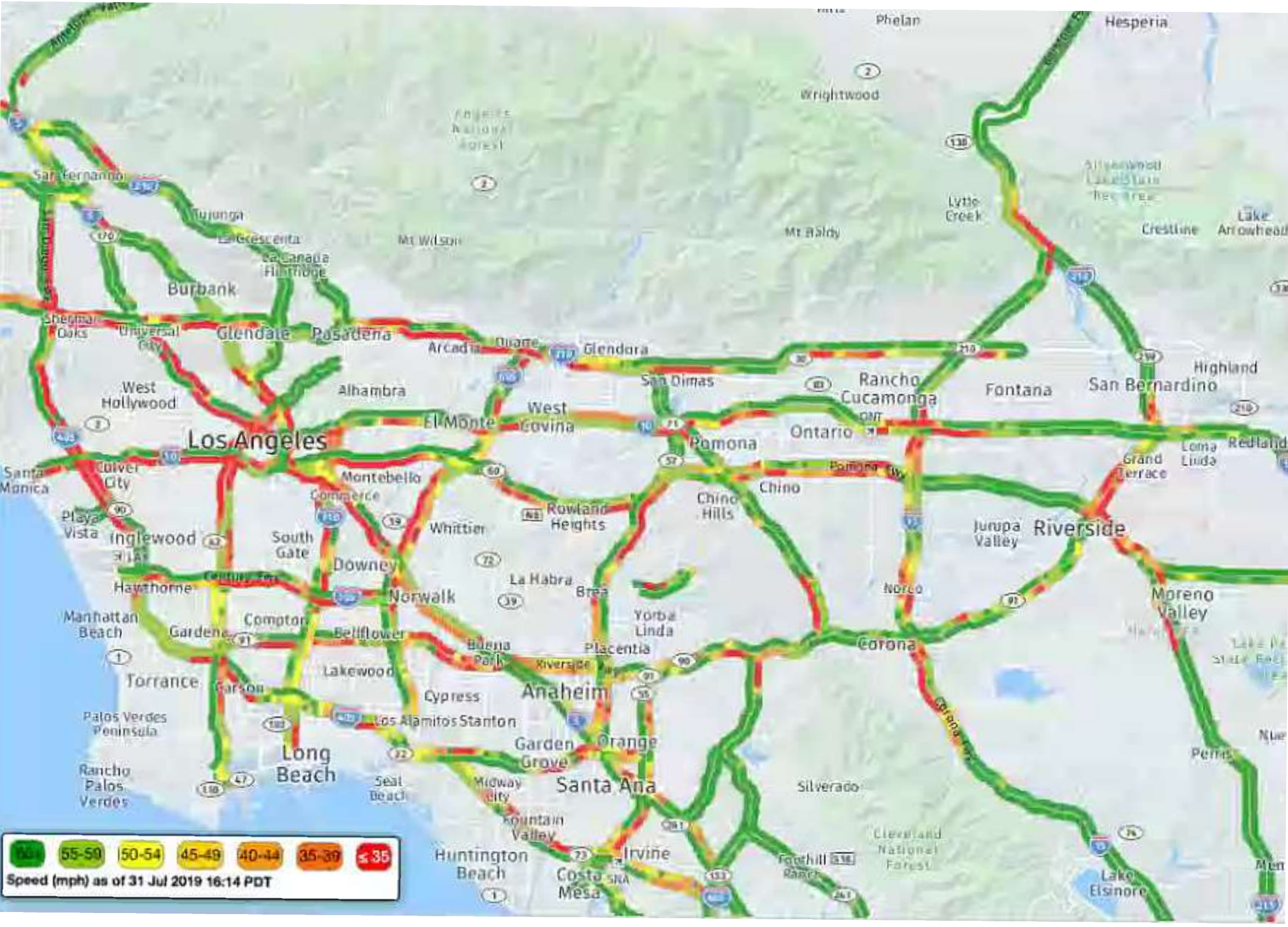} 
\label{fig7_b} } 
\subfloat[Los-loop road segment distribution.]{ 
\centering  
\includegraphics[width=0.34\textwidth] {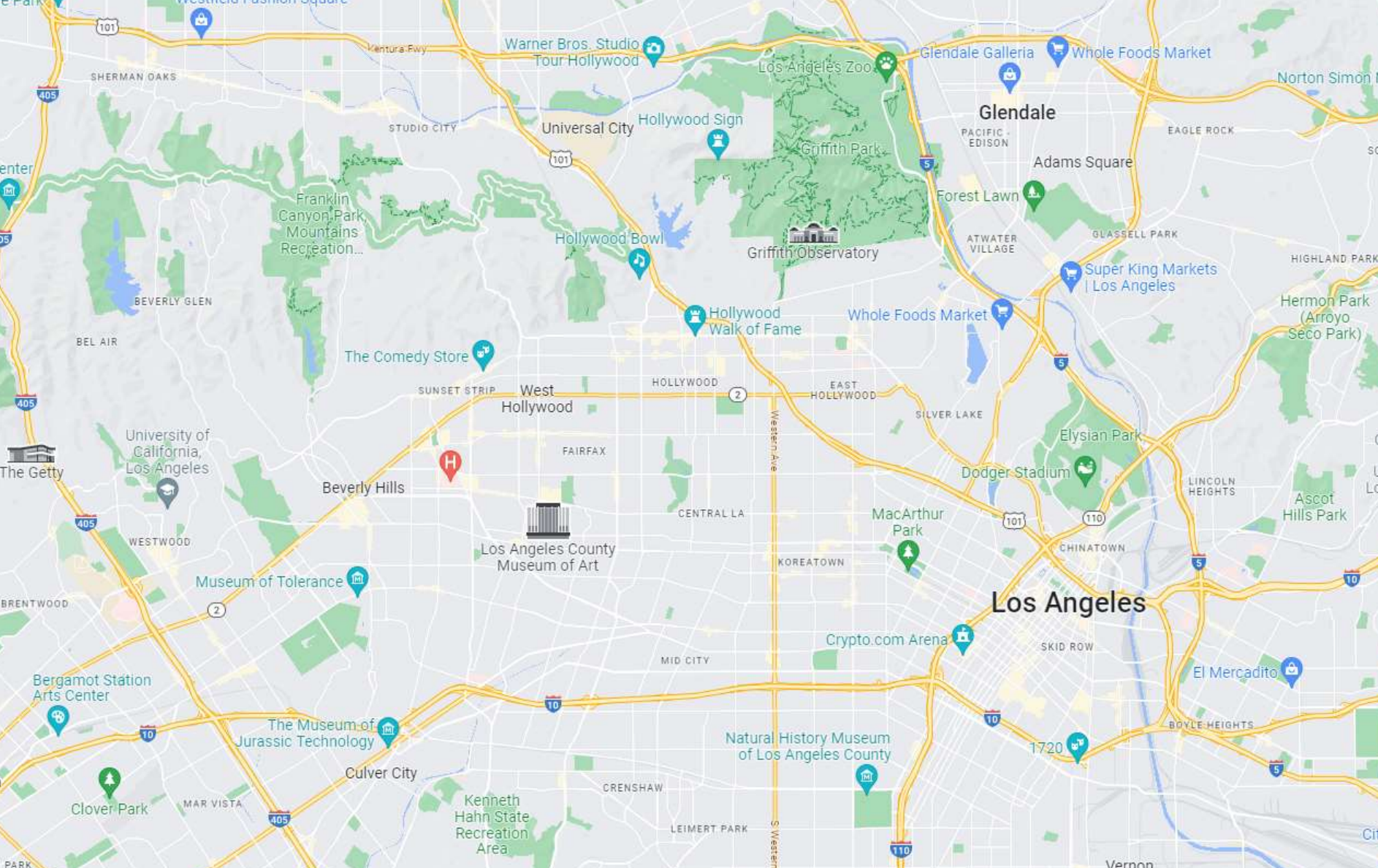} 
\label{fig7_c} 
}

\subfloat[Average traffic flow speed of T-drive dataset.]{ 
\centering 
\includegraphics[width=0.34\textwidth]{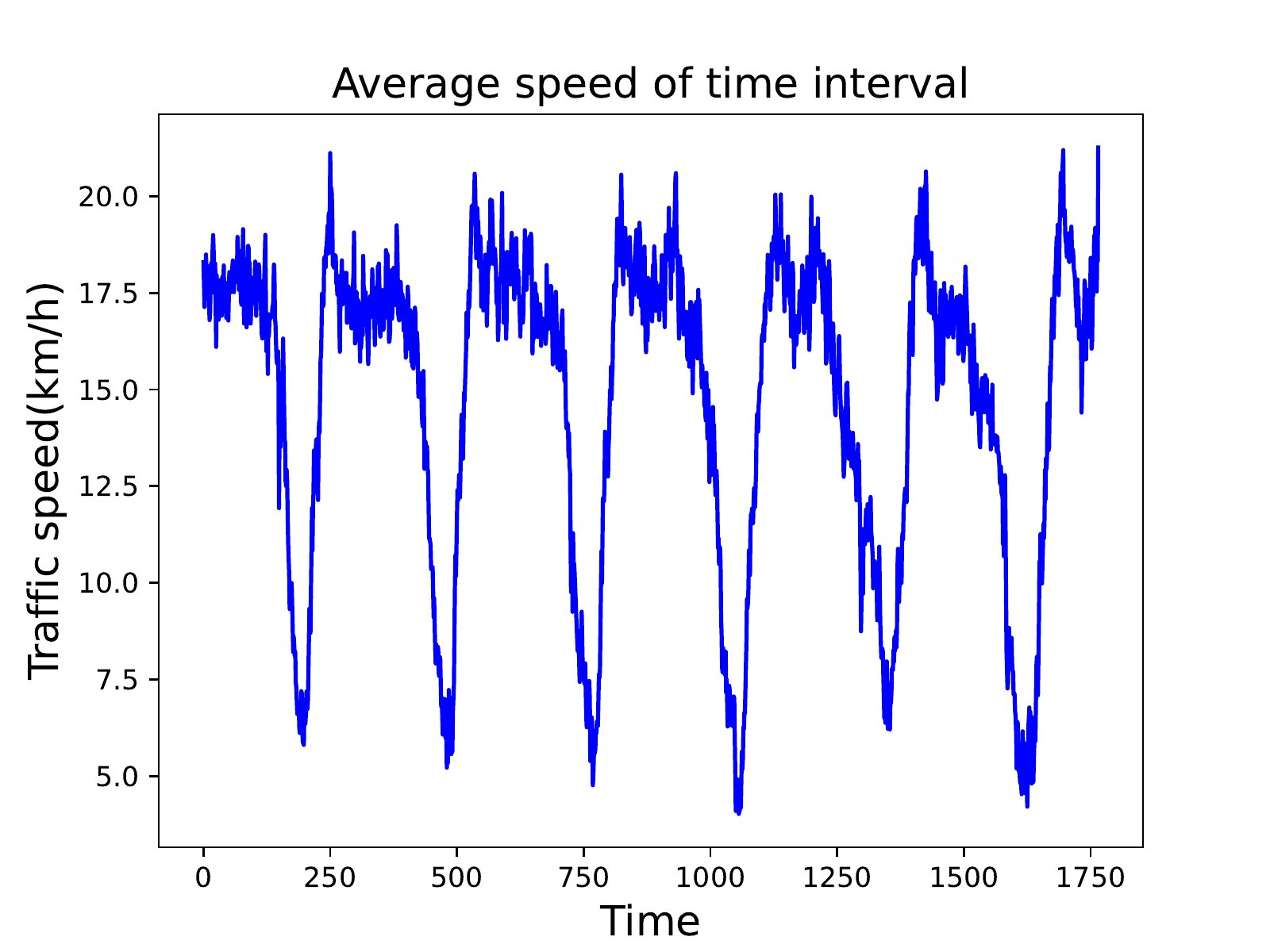} 
\label{fig7_d}
} 
\subfloat[Average traffic flow number of PeMS08 testset.]{
\centering \includegraphics[width=0.34\textwidth]{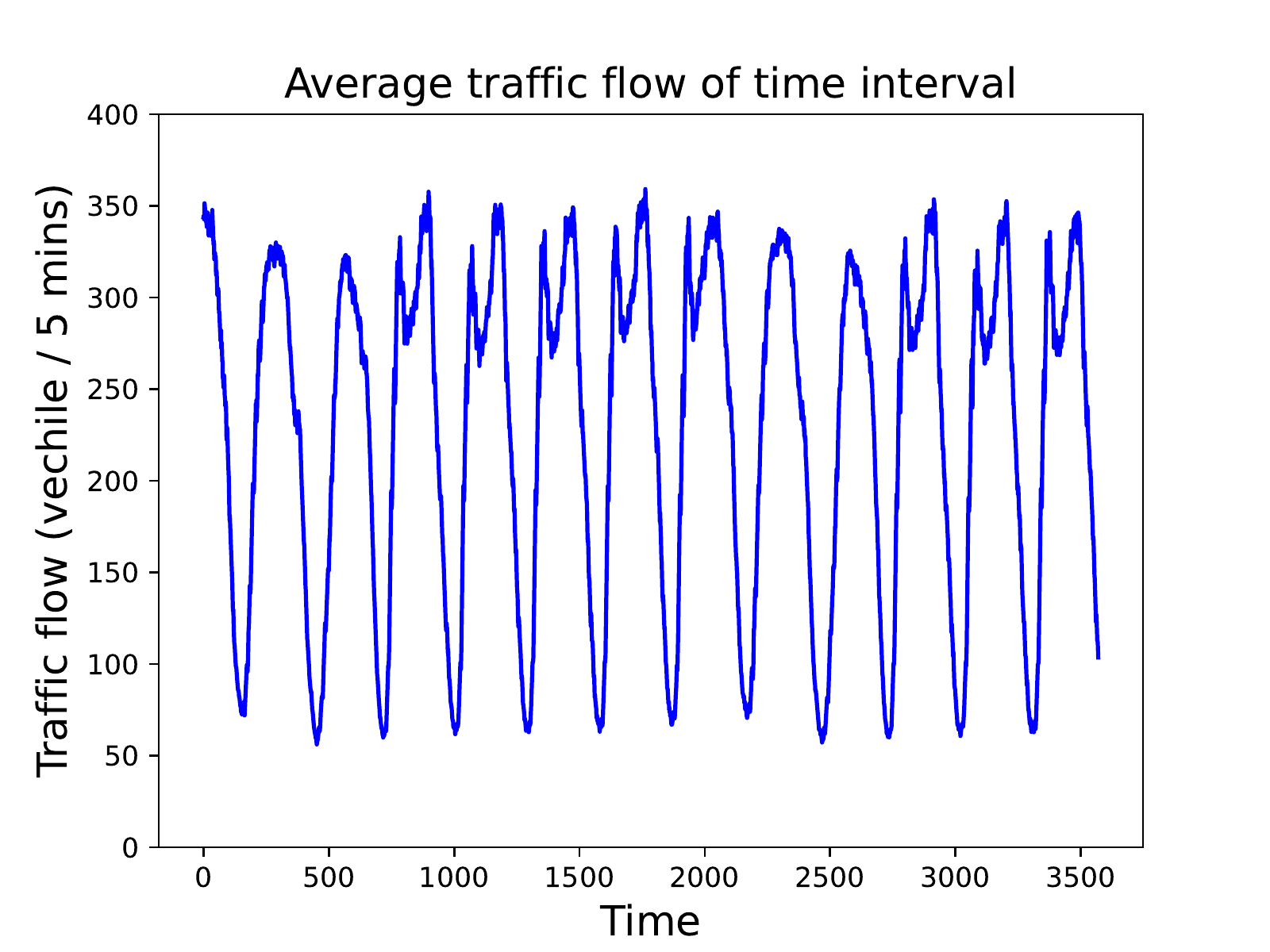}
\label{fig7_e} 
} 
\subfloat[Average traffic flow speed of Los-loop dataset.]{ 	
\centering 	 
\includegraphics[width=0.34\textwidth]{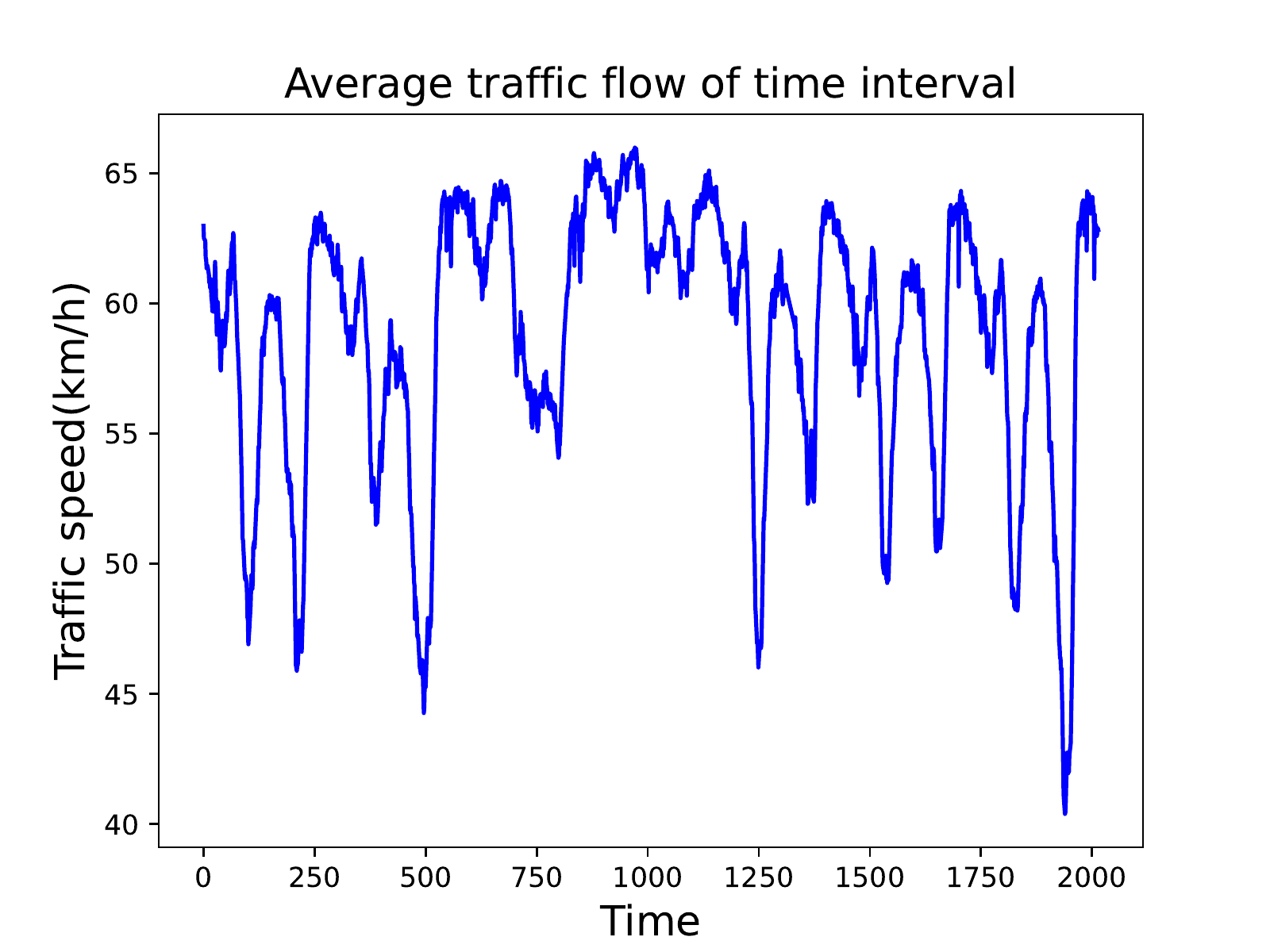} 
\label{fig7_f} 
}

\caption{Traffic segment distribution and average traffic flow curve of transportation datasets.}  
\label{fig5}
\end{figure*}

PeMS dataset is collected by Caltrans Performance Measurement System (PeMS) at a sampling rate of every 30 seconds. PeMS08 contain 170 road segment detectors in California highways and aggregates them into 5 minutes. PeMS08 was collected from July to August 2016. It contains three features: Traffic flow number, Road occupancy, and Traffic flow speed. All three features are used to predict the only traffic flow number in the future. The road segment is shown in Fig. \ref{fig7_b}. The average traffic flow of PeMS08 is shown in Fig. \ref{fig7_e}. The T-Drive dataset is generated by the GPS trajectories of 10357 taxis from Feb. 2 to Feb. 8, 2008, in Beijing, which is processed every 5 minutes to traffic flow speed. We used only the historical traffic flow speed feature to predict the traffic flow speed in the future. We partition Beijing city into 8 × 8 grids. The grids and average traffic flow speed are shown in Fig. \ref{fig7_a} and \ref{fig7_d}. Los-loop is collected from the highway of Los Angeles County in real-time by loop detectors. There are 207 sensors and traffic speeds from Mar. 1 to Mar. 7, 2012, which are aggregated every 5 minutes. The missing values are filled by the linear interpolation method. The road segment and average traffic flow speed are shown in Fig. \ref{fig7_c} and \ref{fig7_f}.

\subsection{Visualization of the prediction result} 
\begin{figure*} 
\centering 
\subfloat[Prediction of T-drive test dataset.]{ 
\centering 	
\includegraphics[width=0.34\textwidth]{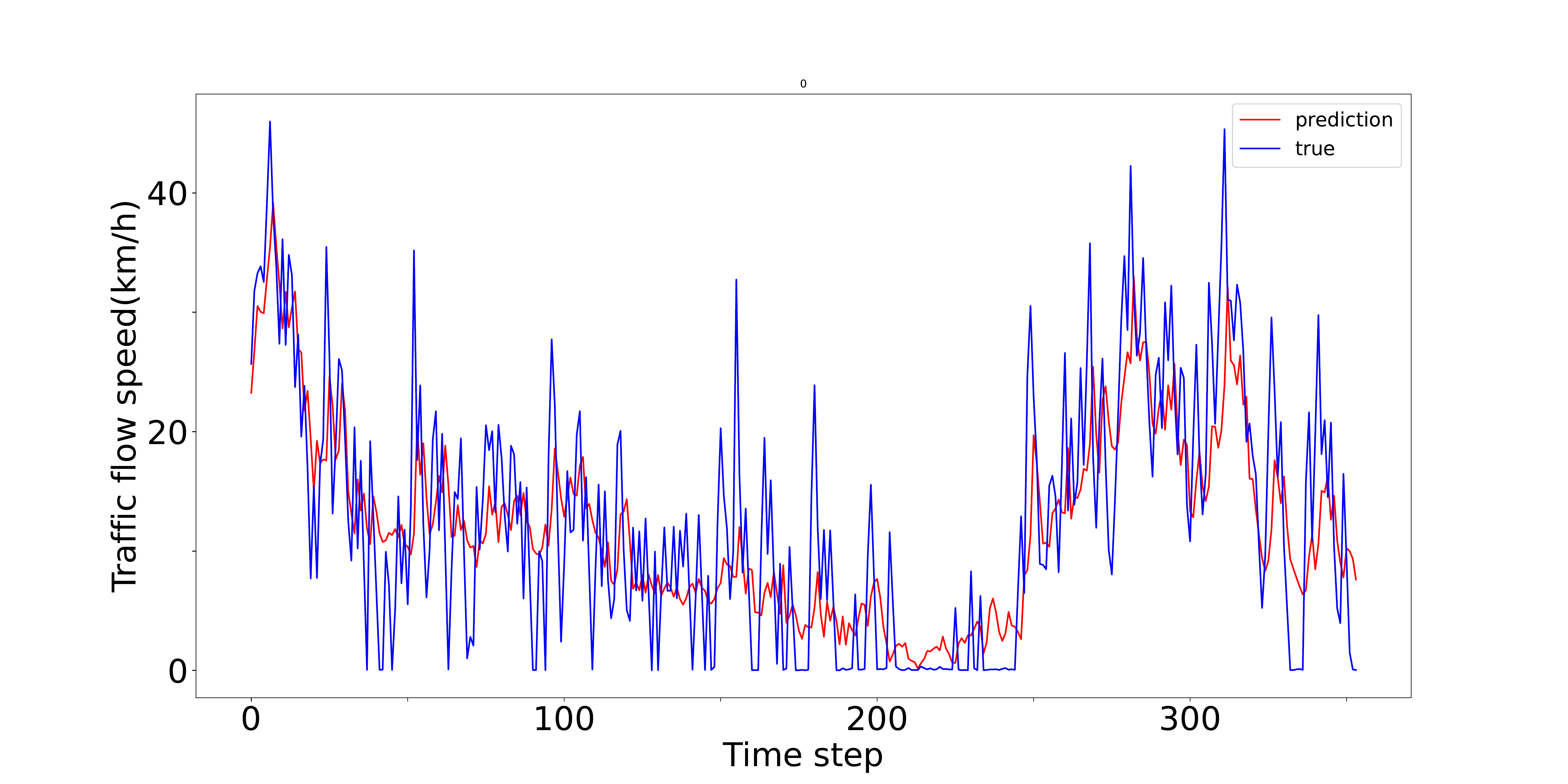} 
\label{fig8_a} 
} 
\subfloat[Part of prediciton of PeMS08 test dataset.]{ 
\centering
\includegraphics[width=0.34\textwidth]{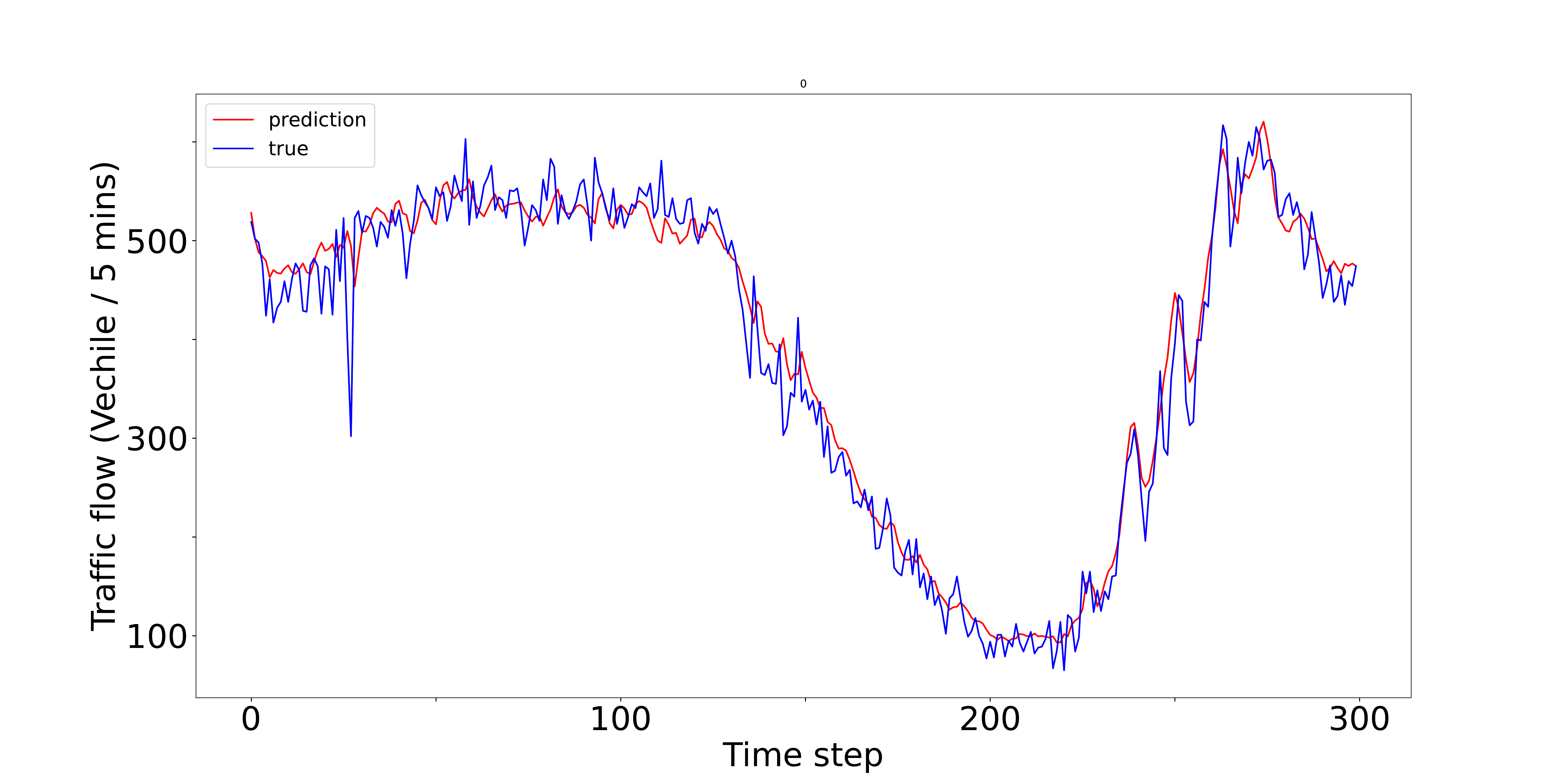} 
\label{fig8_b}
} 
\subfloat[Prediciton of Los-loop test dataset.]{ 
\centering
\includegraphics[width=0.34\textwidth] {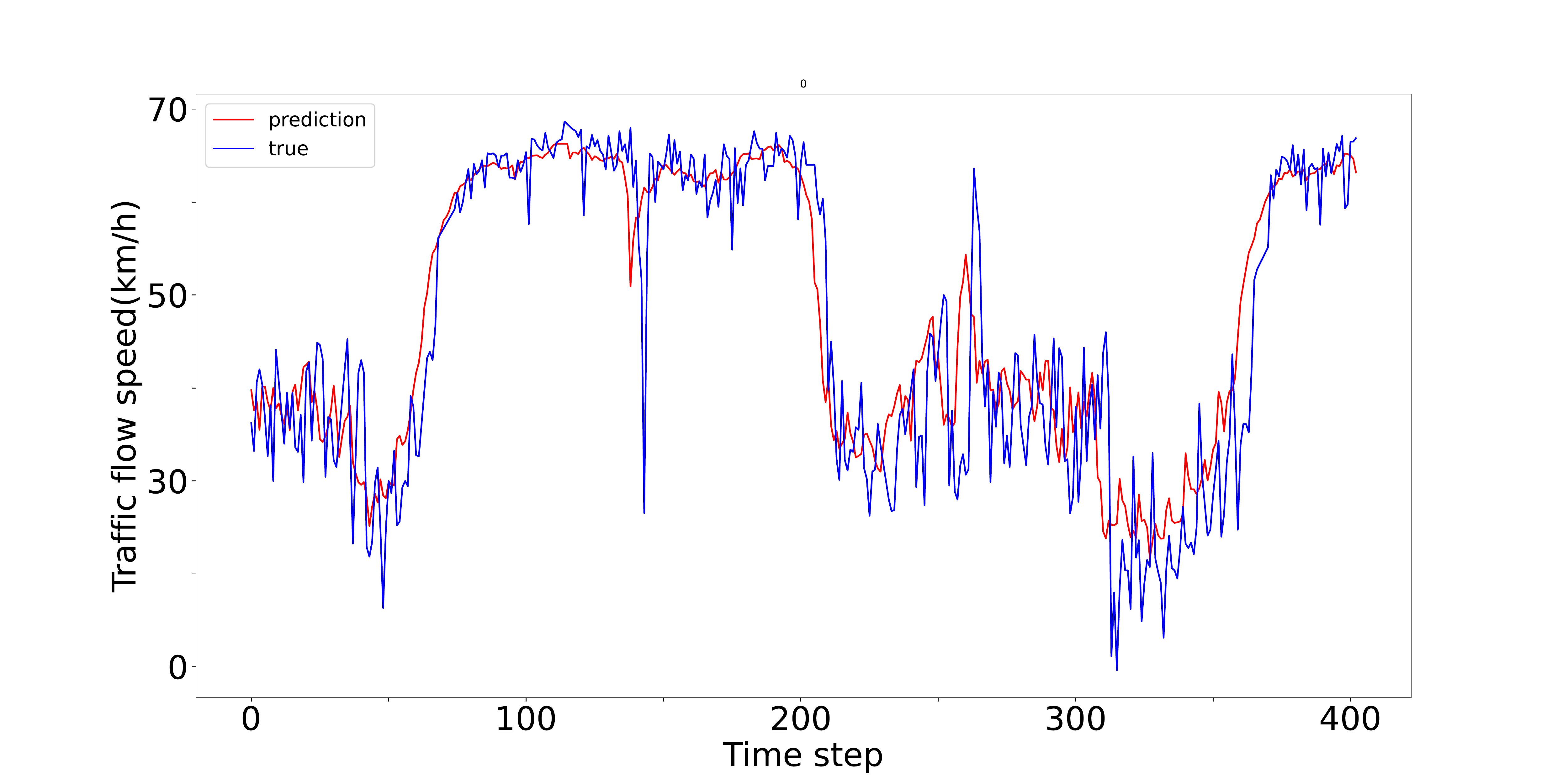} 
\label{fig8_c} 
}  

\subfloat[Half day prediction of T-drive test dataset.]{
\centering 	
\includegraphics[width=0.34\textwidth]{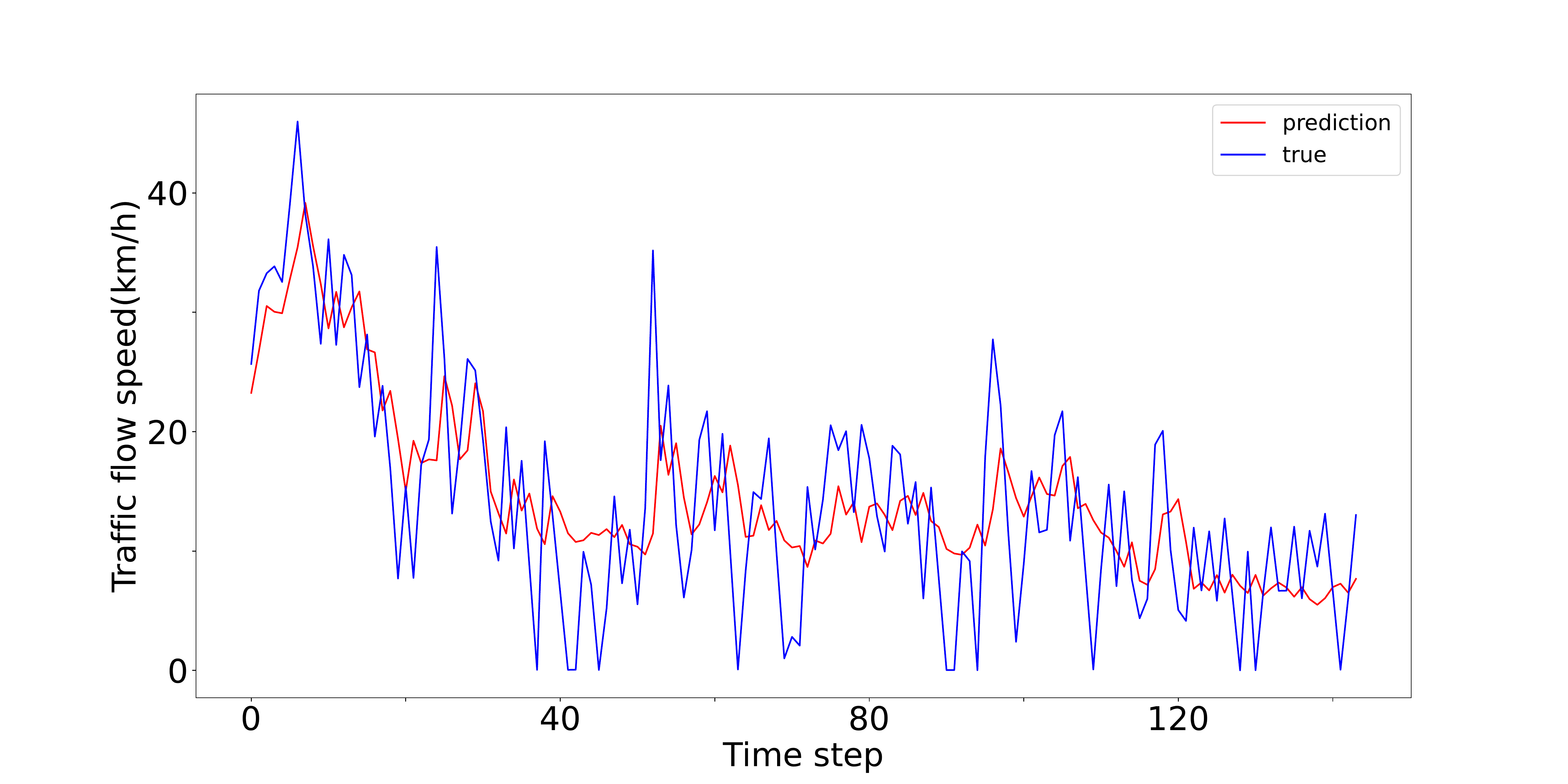} 
\label{fig8_d}
}
\subfloat[Half day prediction of PeMS08 test dataset.]{ 	
\centering 
\includegraphics[width=0.34\textwidth]{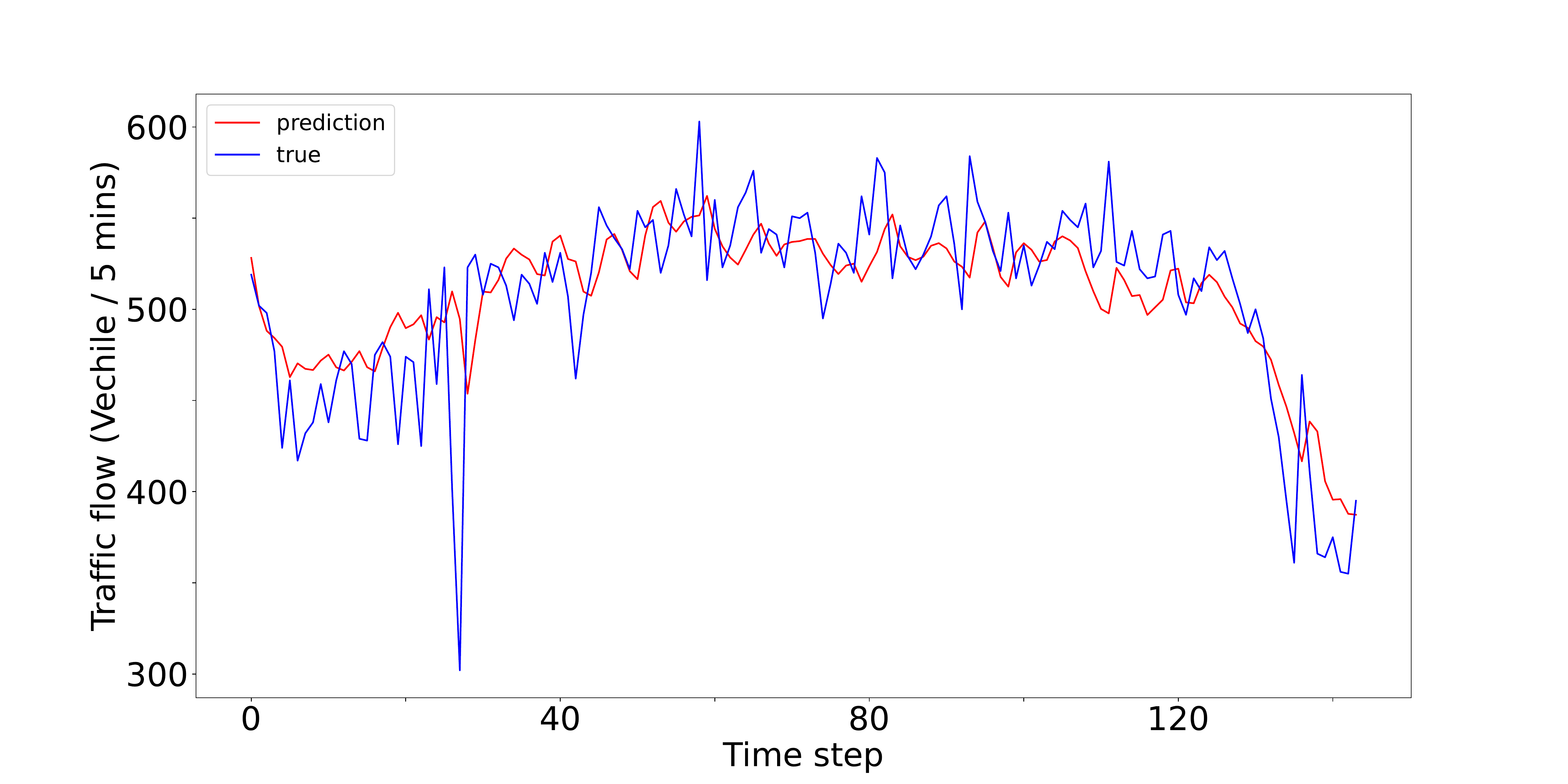} 
\label{fig8_e} 
}
\subfloat[Half day prediction of Los-loop test dataset.]{ 
\centering 
\includegraphics[width=0.34\textwidth] {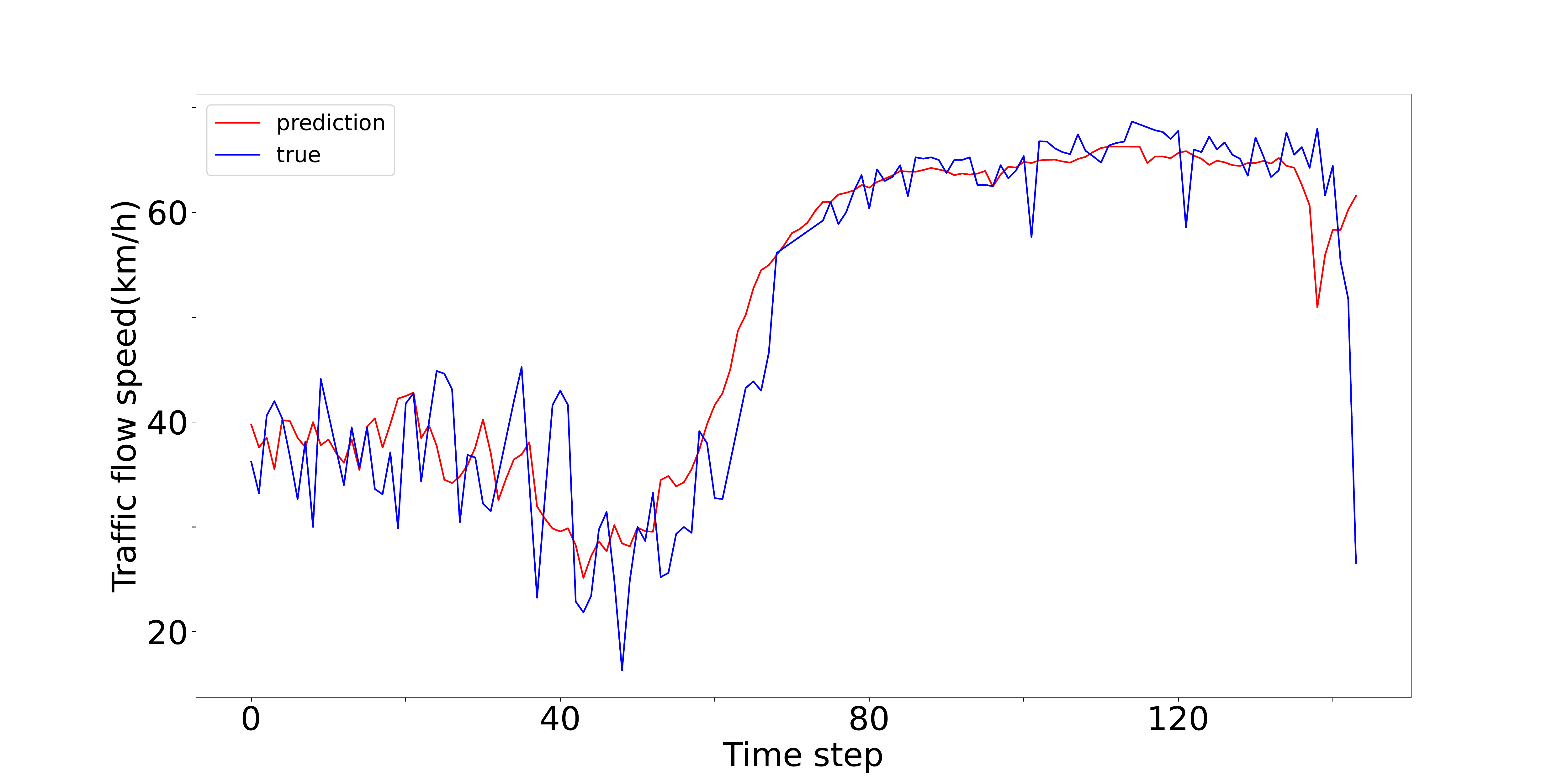} 
\label{fig8_f} 
} 
\caption{visualization results for prediction and ground truth of transportation datasets.} \label{fig8}
\end{figure*}

In order to evaluate the prediction results of our model more intuitively, we plot the predicted results and the ground truth values in Fig. \ref{fig8}. Fig. \ref{fig8_a} is the total predicted value of grid \#4 in the 354 testset time series of T-Drive. We magnify half-day prediction results in 144-time series, which is shown in Fig. \ref{fig8_d}. Fig. \ref{fig8_b} is part of the predicted value of road \#4 in the 300-time series of test time series of PeMS08 dataset. We magnify half-day prediction results in the 144-time series, which are shown in Fig. \ref{fig8_e}. Similarly, Fig. \ref{fig8_c} and Fig. \ref{fig8_f} are the total predicted value of road \#4 in the 403-time series and half-day prediction results of the testset time series of the Los-loop dataset.

\subsection{Healthcare dataset of FMRI}
\begin{figure*}
\centering
\subfloat[Visualization curve of FMRI-13 dataset.]{ 	
\centering	
\includegraphics[width=0.47\textwidth,height=6cm]{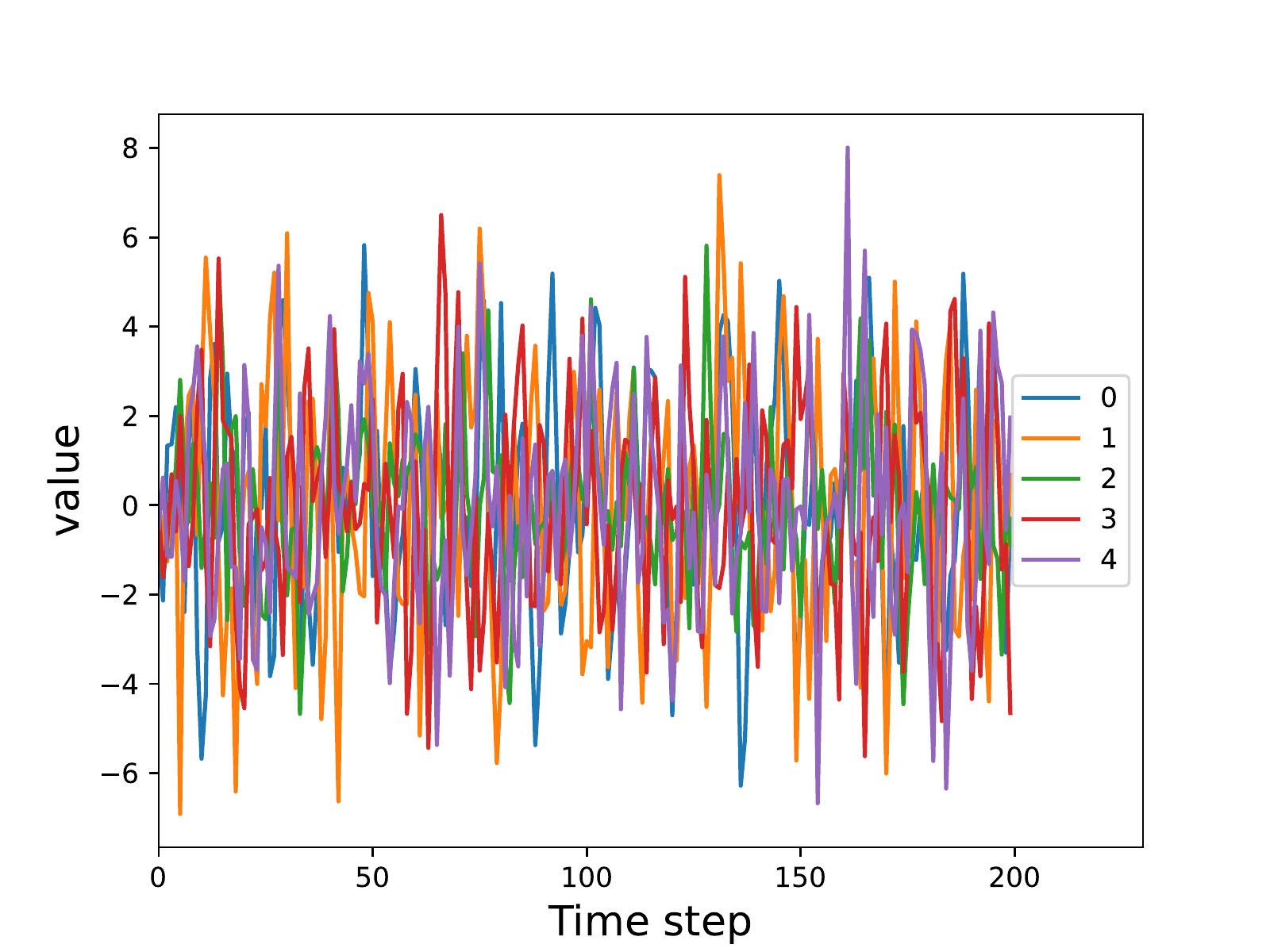} 
\label{fig9_a}
}
\subfloat[Ground truth causal graph of FMRI-13.]{ 
\centering	
\includegraphics[width=0.47\textwidth,height=6cm]{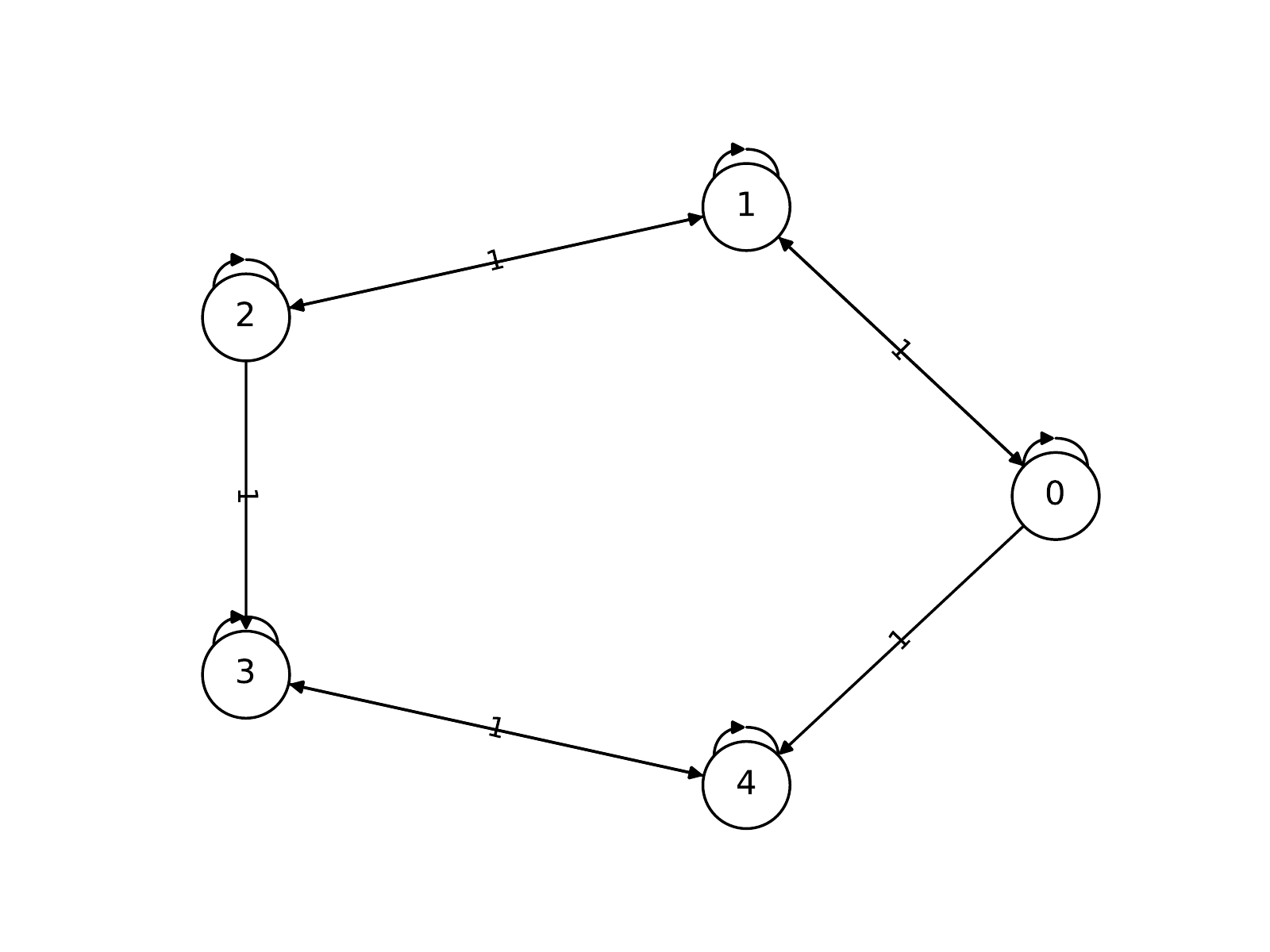} 	\label{fig9_b}
} 

\subfloat[Visualization curve of FMRI-3 dataset.]{ 
\centering	
\includegraphics[width=0.47\textwidth,height=6cm]{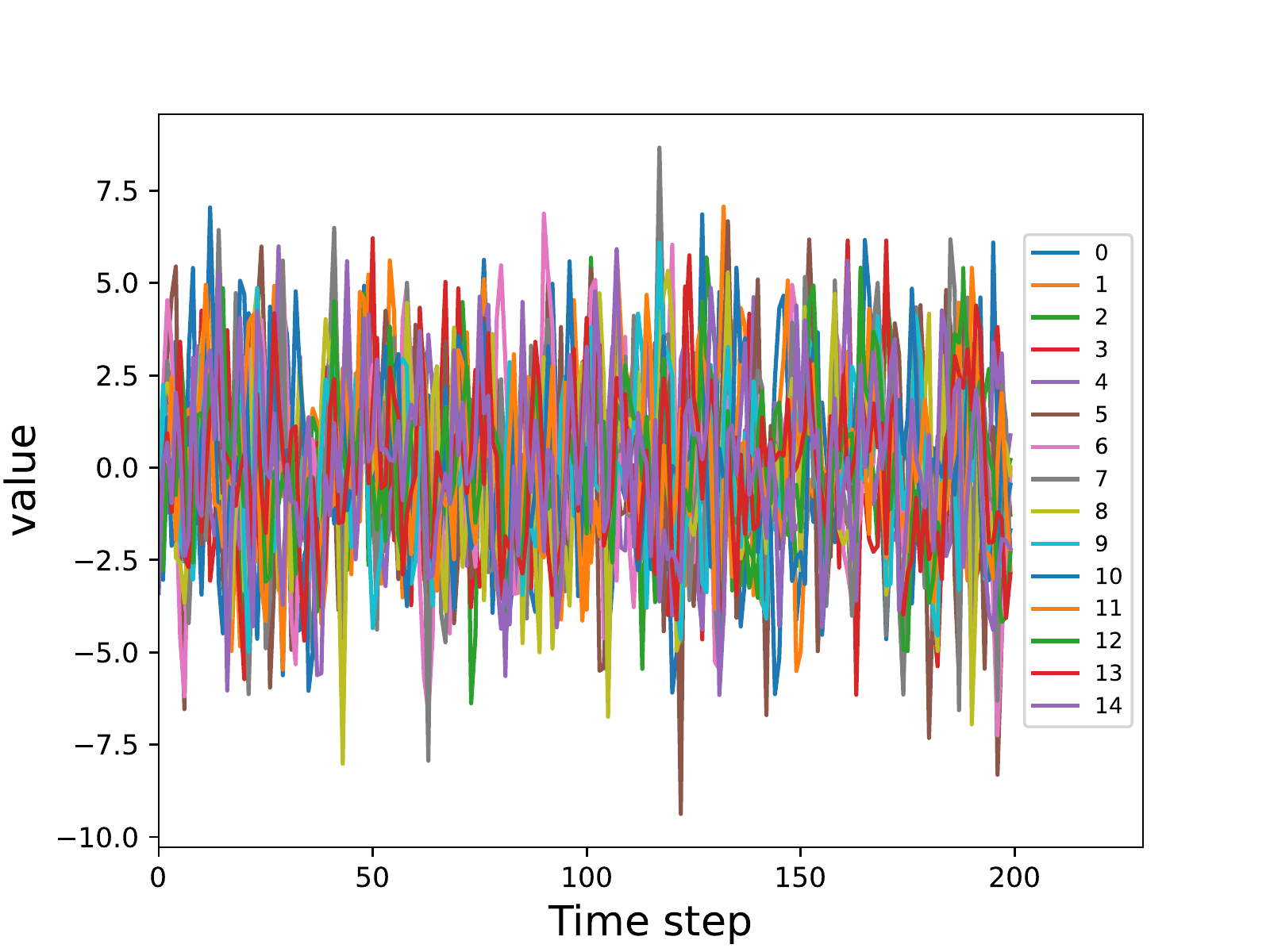}
\label{fig9_c} 
}
\subfloat[Ground truth causal graph of FMRI-3.]{ 
\centering
\includegraphics[width=0.47\textwidth,height=6cm]{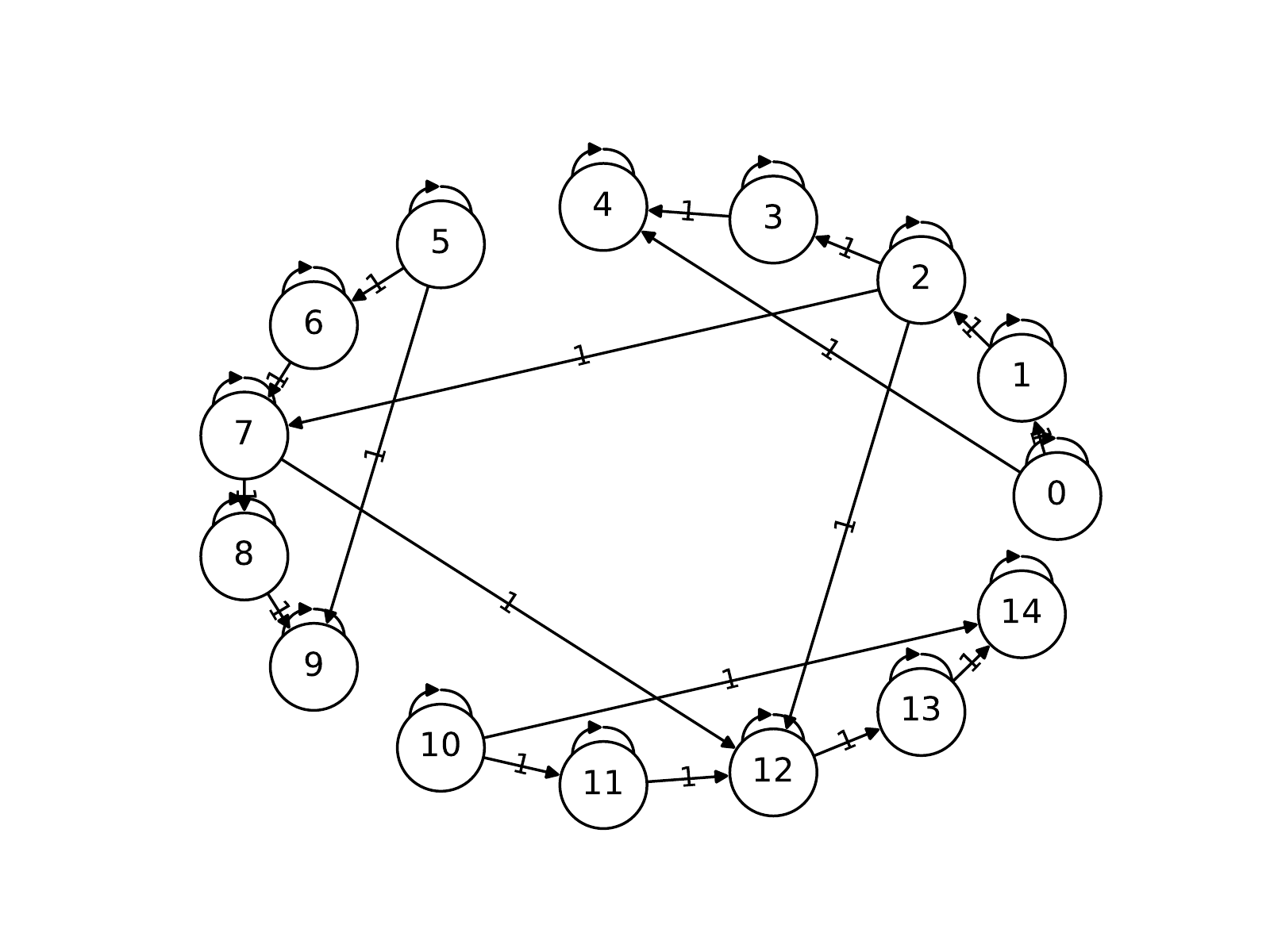} 	\label{fig9_d}
}  

\subfloat[Visualization curve of FMRI-4 dataset.]{ 
\centering	
\includegraphics[width=0.47\textwidth,height=6cm]{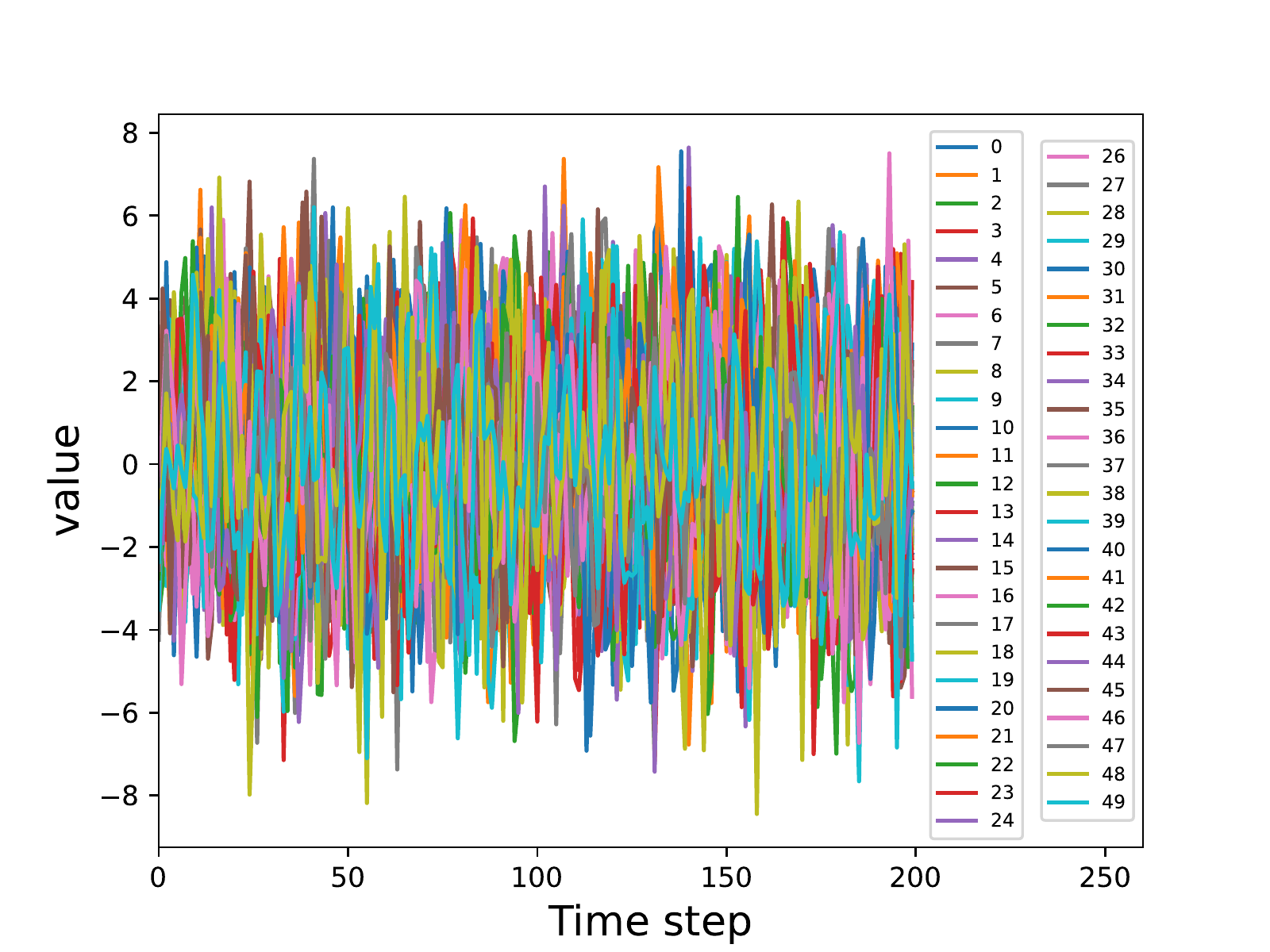} 
\label{fig9_e}
} 
\subfloat[Ground truth causal graph of FMRI-4.]{ 
\centering	
\includegraphics[width=0.47\textwidth,height=6cm]{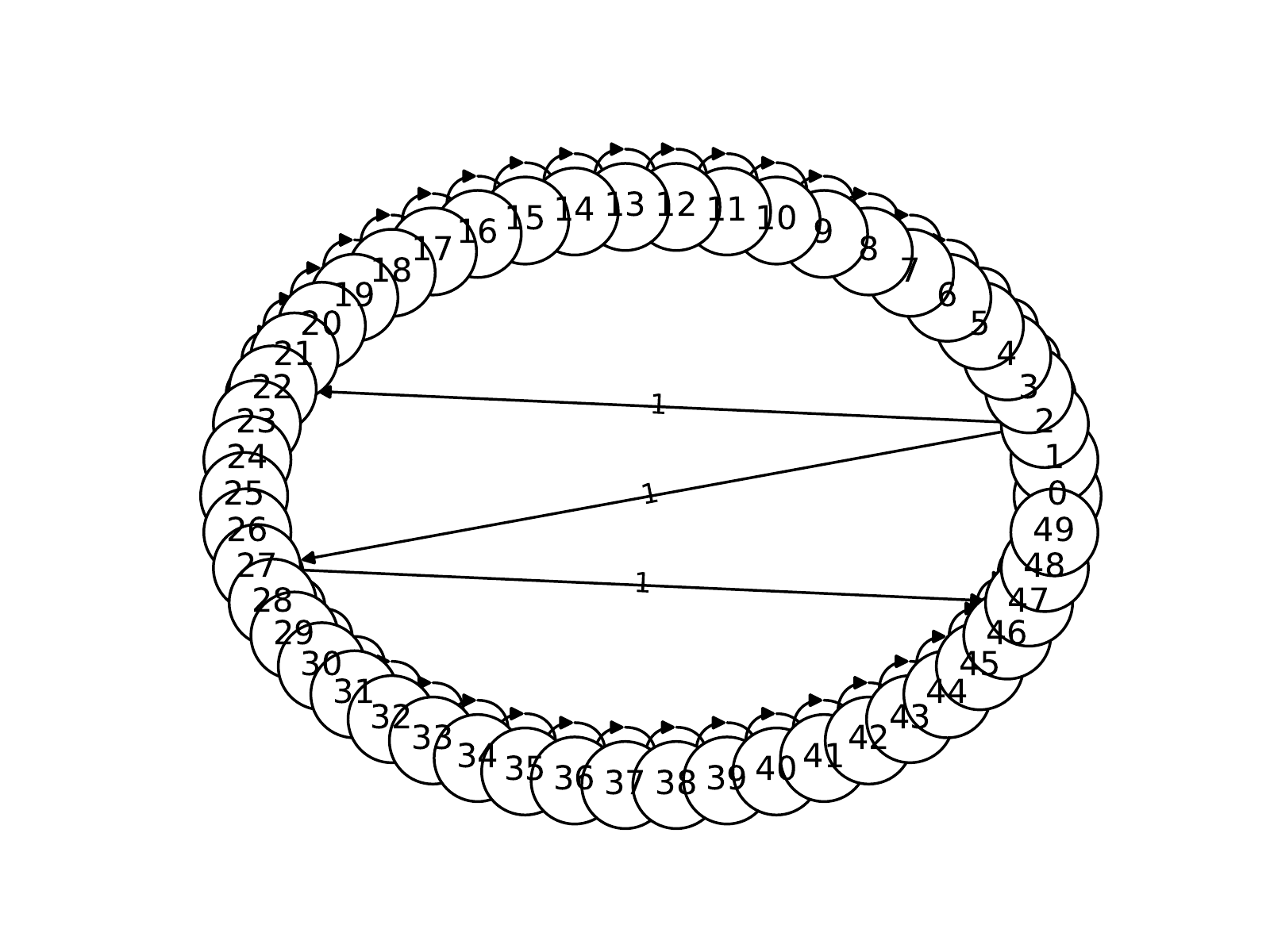} 	\label{fig9_f}
} 
\caption{Visualization curve and ground truth causal graph of FMRI dataset.}  
\label{fig9} 
\end{figure*}

The original Functional Magnetic Resonance Imaging (FMRI) dataset is collected by the FMRIB Analysis Group at Oxford University and preprocessed by Nauta. The whole data contain 28 different underlying brain networks. The dataset contains  different nodes, self-causation, and time series lengths. We choose FMRI-3, FMRI-4, and FMRI-13 for our experiments since the causal graph structure is the most important factor in our paper. The visualization curve of the three sub-datasets and ground truth causal graphs are shown in Fig. \ref{fig9}

\subsection{ROC curve of FMRI}
\begin{figure*} 
\centering 
\subfloat[FMRI-13 ROC curve of DVGNN model]{ 
\centering	 
\includegraphics[width=0.4\textwidth]{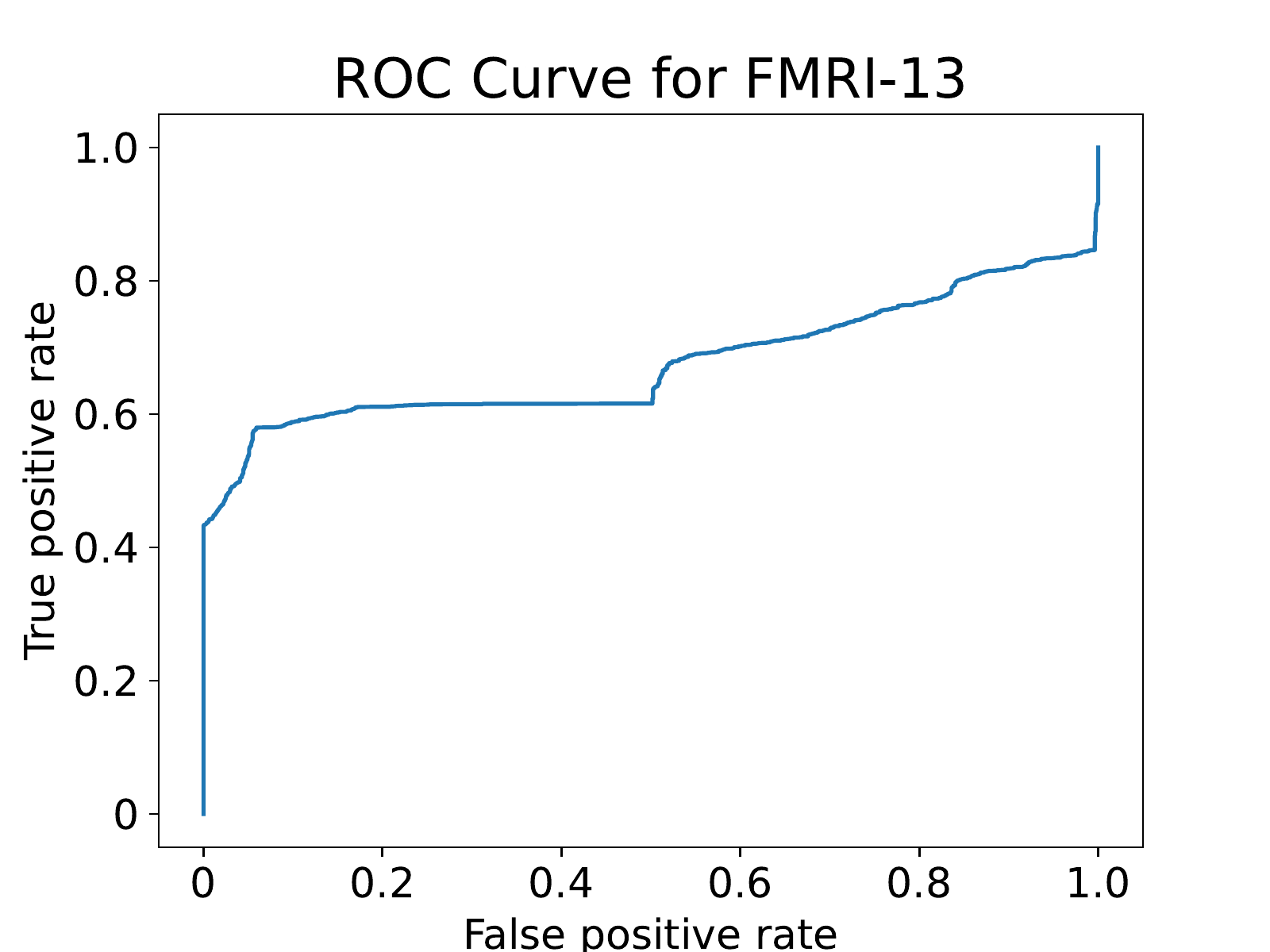}
\label{fig10_a}
}  

\subfloat[FMRI-3 ROC curve of DVGNN model]{ 
\centering	
\includegraphics[width=0.4\textwidth]{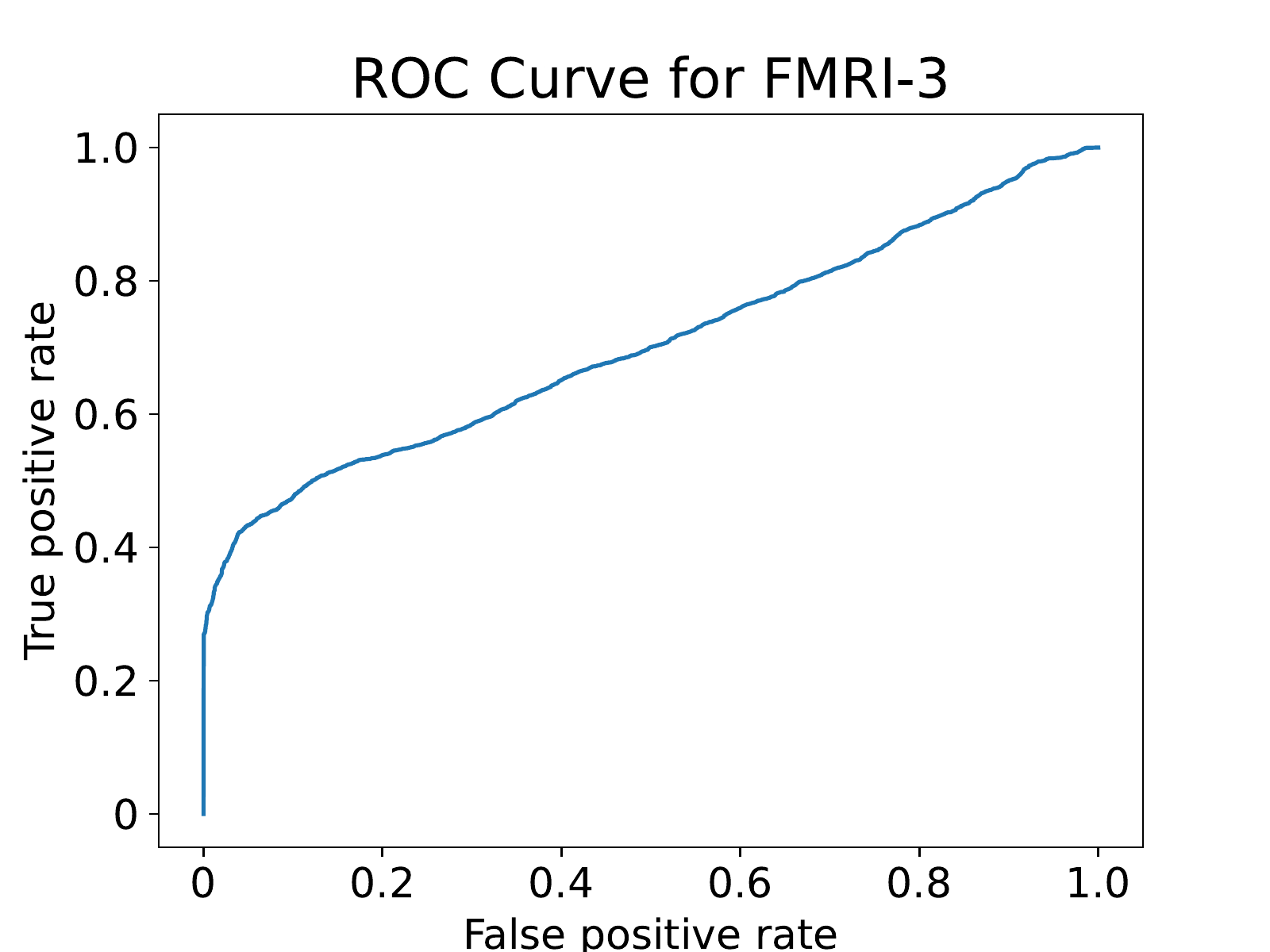} 
\label{fig10_b}
} 

\subfloat[FMRI-4 ROC curve of DVGNN model.]{
\centering	
\includegraphics[width=0.4\textwidth]{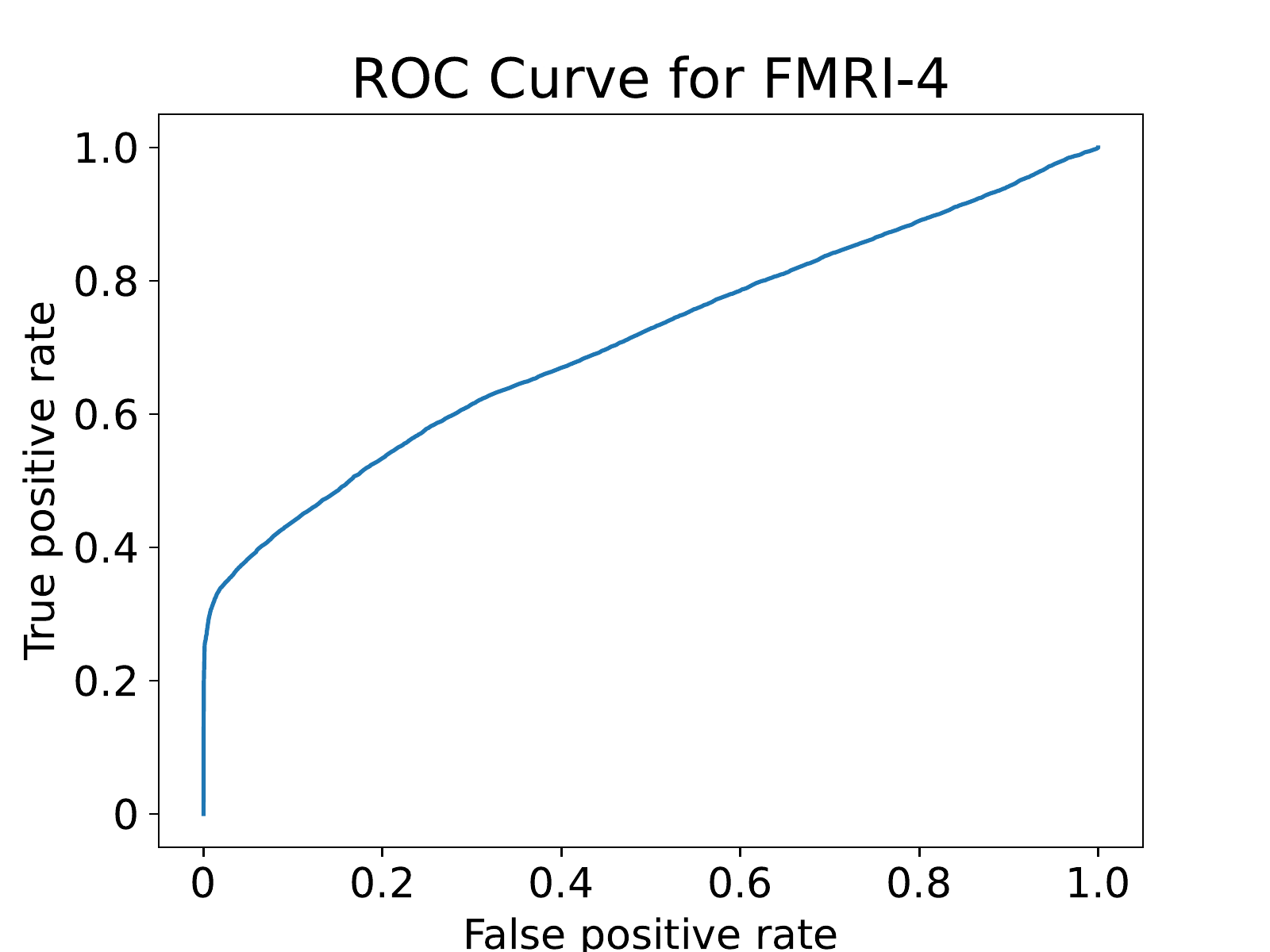} 
\label{fig10_c}
} 
\caption{FMRI ROC curve of DVGNN model.}
\label{fig10} 
\end{figure*}

In order to study the performance of the generated causal graph in more detail, we plot the ROC curve of FMRI, which is shown in Fig. \ref{fig10}. From Fig. \ref{fig10}, high accuracy and performance can be demonstrated from the ROC curve.

\end{document}